\documentclass{article}

    \PassOptionsToPackage{authoryear,round}{natbib}

\usepackage[preprint]{neurips_2026}

\usepackage[utf8]{inputenc} %
\usepackage[T1]{fontenc}    %
\usepackage{url}            %
\usepackage{booktabs}       %
\usepackage{amsfonts}       %
\usepackage{nicefrac}       %
\usepackage{xcolor}         %
\usepackage{amsmath,amsthm,amsfonts,amssymb,dsfont}
\usepackage{multirow}
\usepackage{xspace}
\usepackage{float}
\usepackage{adjustbox}
\usepackage{enumitem}
\usepackage{wrapfig}
\usepackage{subcaption}

\newcommand{\efficientparagraph}[1]{\noindent\textbf{#1.}\hspace{0.5em}}

\usepackage[colorlinks=true]{hyperref}
\hypersetup{
    citecolor=blue!50!black,
    linkcolor=blue!50!black,
    urlcolor=blue!50!black
}

\usepackage{titletoc}
\newcommand{\cb}{\textsc{ContinuousBench}\xspace}
\newcommand{\news}{\textsc{News}\xspace}
\newcommand{\geminon}{\textsc{Geminon}\xspace}

\newcommand{\eg}{{e.g.},\xspace}

\usepackage[most]{tcolorbox}
\usepackage{listings}
\definecolor{promptbg}{RGB}{240,240,240}
\definecolor{promptframe}{RGB}{155,155,155}
\definecolor{prompttitle}{RGB}{170,170,170}

\definecolor{promptbg}{HTML}{F7F9FE}
\definecolor{promptframe}{HTML}{B8C3D9}
\definecolor{prompttitle}{HTML}{6B74C9}
\lstdefinestyle{promptstyle}{
  basicstyle=\ttfamily\scriptsize,
  breaklines=true,
  breakatwhitespace=false,
  breakautoindent=false,
  breakindent=0pt,
  columns=fullflexible,
  keepspaces=true,
  showstringspaces=false,
  upquote=true
}

\newtcblisting{promptbox}[1]{
  enhanced,
  breakable,
  colback=promptbg,
  colframe=promptframe,
  colbacktitle=prompttitle,
  coltitle=white,
  title={#1},
  boxrule=0.5pt,
  arc=1.5mm,
  left=1mm,
  right=1mm,
  top=1mm,
  bottom=1mm,
  title filled,
  listing only,
  listing options={style=promptstyle}
}

\definecolor{jsonbg}{RGB}{247,247,247}
\definecolor{jsonframe}{RGB}{180,180,180}
\definecolor{jsontitle}{RGB}{110,110,110}

\lstdefinestyle{jsonstyle}{
  basicstyle=\ttfamily\scriptsize,
  breaklines=true,
  breakatwhitespace=false,
  breakautoindent=false,
  breakindent=0pt,
  columns=fullflexible,
  keepspaces=true,
  showstringspaces=false,
  upquote=true
}

\newtcolorbox{minicard}[1]{
  enhanced,
  colback=jsonbg, colframe=jsonframe,
  colbacktitle=jsontitle, coltitle=white,
  fonttitle=\bfseries\footnotesize,
  title={#1},
  boxrule=0.5pt, arc=1.5mm,
  left=1.5mm, right=1.5mm, top=0.8mm, bottom=0.8mm,
  fontupper=\ttfamily\scriptsize,
  width=\linewidth
}

\newtcblisting{jsonblock}[1]{
  enhanced,
  colback=jsonbg,
  colframe=jsonframe,
  colbacktitle=jsontitle,
  coltitle=white,
  fonttitle=\bfseries,
  title={#1},
  boxrule=0.5pt,
  arc=1.5mm,
  left=1mm,
  right=1mm,
  top=1mm,
  bottom=1mm,
  listing only,
  listing options={style=jsonstyle}
}

\usepackage{tikz}
\usetikzlibrary{positioning, arrows.meta}

\definecolor{flowgray}{RGB}{110,110,110}

\tikzset{
  card/.style={
    rectangle, rounded corners=3pt,
    draw=jsonframe, line width=0.5pt,
    fill=jsonbg,
    inner sep=7pt,
    align=left,
    anchor=north,
    minimum height=3.6cm,
    font=\ttfamily\scriptsize,
  },
  cardtitle/.style={
    font=\bfseries\small,
    text=jsontitle,
    anchor=south,
    yshift=4pt,
  },
  flow/.style={
    -{Stealth[length=2.5mm, width=2.5mm]},
    line width=0.8pt, draw=flowgray,
    shorten <=2pt, shorten >=2pt,
  },
  flowlbl/.style={
    font=\itshape\scriptsize, color=flowgray, align=center,
  },
}

\title{\cb: \\Can Differentially Private Synthetic Text Improve Capabilities?}

\author{
  Peihan Liu\thanks{Project leads. Correspondence to \texttt{peihanliu@cs.columbia.edu} and \texttt{alexbie@google.com}.}~~\thanks{Work done as a Student Researcher at Google.} \\
  Columbia University\\
  \And
  Lucas Rosenblatt\textsuperscript{$\dagger$}\\
  NYU \\
  \And
  Weiwei Kong \\
  Google Research \\
  \And
  Natalia Ponomareva \\
  Google Research\\
  \And
  Gautam Kamath \\
  University of Waterloo \\ \& Vector Institute\\
  \And
  Rachel Cummings \\
  Columbia University \\
  \And
  Roxana Geambasu \\
  Columbia University\\
  \And
  Yu Gan \\
  Google \\
  \And
  Lillian Tsai \\
  Google \\
  \And
  Alex Bie\textsuperscript{*} \\
  Google Research \\
}

\begin{document}

\maketitle
\begin{abstract}
Differentially private (DP) text synthesis promises to unlock sensitive corpora for model training, but it remains unclear whether DP synthetic data transmits genuinely new \emph{knowledge and capabilities} present \emph{only} in those corpora. This is because existing evaluations rely on tasks that are nearly solvable without training, so strong benchmark performance does not establish that DP synthesis can substitute original data access. Thus, we introduce \cb{}, a continuously and automatically-regenerated benchmark that measures \emph{capability gain} from DP synthetic text. Each quarter, a new release pairs a never-before-seen training corpus with a derived QA set, constructed to be: (1) unsolvable sans-corpus; and (2) learnable under DP, as the tested knowledge is supported by hundreds of independent records. Researchers produce DP synthetic data from the training corpus and run our standardized training and evaluation harness on their synthetic data to measure gains. We instantiate two tracks: \geminon, a procedurally-generated dataset about fictional creatures; and \news, a stream of newly crawled public news articles. Although standard benchmarks are nearly saturated, on \cb{} we find that non-private synthesis transfers substantial knowledge from the original corpus, while state-of-the-art DP synthesis methods generally fail to do so, even at $\varepsilon=100$. We make \cb available at \url{https://hf.co/ContinuousBench} to measure and accelerate progress on DP synthetic data.

\end{abstract}

\maketitle

\section{Introduction}\label{sec:introduction}
As publicly available text becomes increasingly depleted as a training source for model improvement, much of the remaining high-value data resides in sensitive or proprietary corpora that cannot be directly shared or trained on (\eg clinical notes, user data). Differentially private (DP) text synthesis offers a compelling alternative in which we can learn from such corpora through a privacy-preserving synthetic release. The key question, however, is not whether DP synthesis can generate stylistically similar text, but rather, whether DP synthetic text can preserve the \emph{capability gains} that access to the original restricted corpus would have provided. That is,
\begin{quote}
    \emph{Does sensitive data unlocked via differentially private synthesis improve the frontier of model capabilities?}
\end{quote}

Yet this substitutability question is not answered by existing literature, largely because we lack a rigorous
evaluation framework that tests whether a given DP synthesis method is capable of preserving the high-value facts and skills learnable from the sensitive corpus. In particular, to claim that DP synthesis can replace access to sensitive data in generality, an evaluation must rule out the possibility of downstream improvements coming from: (1) elicitation of knowledge already
present in the base model from public pretraining; (2) the simplicity of the evaluated task, that is, if task success requires only superfical distributional matching rather than learning high-value facts and skills; and (3) teacher-student distillation
effects when a stronger generator produces data for a
weaker downstream model. We expand on these three confounders in more detail.

\begin{figure}[t]
    \centering
    \includegraphics[width=\linewidth]{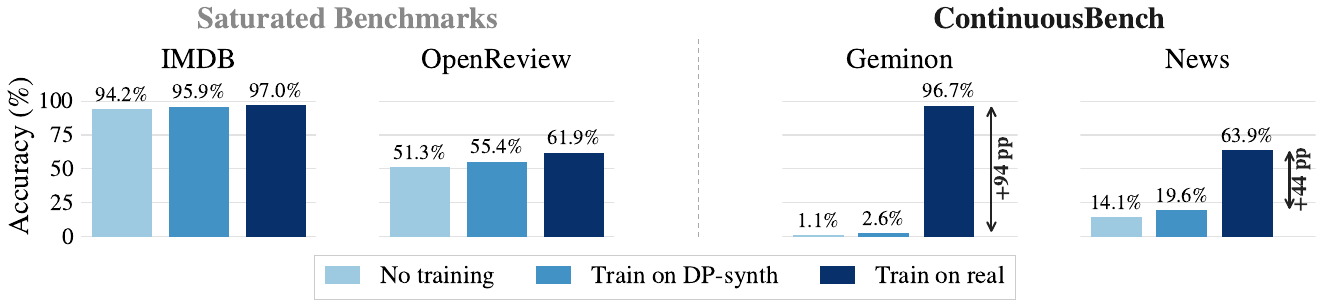}
    \caption{\textbf{A saturated benchmark leaves little room to distinguish methods.} We compare downstream accuracy of \textsc{Gemma 3 4B} after 3 increasingly informative training regimes: no training, training on DP synthetic data ($\varepsilon=10$), and training on the real corpus. Standard benchmarks are near-saturated, but \cb{} reveals that DP synthesis has significant headroom.
    }
    \label{fig:teaser}
\end{figure}

\emph{First}, we seek to address the risk that evaluated knowledge and capabilities may already be present in the pretrained base model \citep{pmlr-v235-tramer24a, arora2022focus}. In this case, strong downstream performance does not imply that DP synthesis transferred knowledge from the corpus, since we cannot rule out possible elicitation of knowledge already learned during pretraining. Indeed, this is the case for existing evaluations in the literature. Standard benchmark settings based on classification-style utility \citep{yue2023synthetic,kurakin2023harnessing, pmlr-v235-xie24g,tan2025synthesizing} are often nearly saturated for modern language models, leaving little room to distinguish methods (see Figure~\ref{fig:teaser}). 
Intuitively, access to increasingly informative training data should yield monotonically increasing utility: 
\[
\textit{No training $<$ Train on DP synthetic $<$ Train on real corpus}.
\]

Yet many widely used evaluation settings do not provide enough resolution for this comparison. 
IMDB~\citep{IMDB}, for example, is a binary sentiment-classification task over movie reviews and has been widely used in evaluations of DP text synthesis~\citep{kurakin2023harnessing,amin-etal-2024-private,zou2025contrastive,banayeeanzade2026epsvec}. It illustrates the problem: sentiment classification is a generic skill rather than a corpus-specific capability, and modern pretrained models already solve it. Therefore strong performance on such an evaluation does not imply a DP synthesis method can transfer new capabilities.

\emph{Second}, the evaluated task may not measure the kind of capability that makes sensitive data valuable. Metrics commonly used in the DP synthetic text literature, such as MAUVE \citep{pillutla2021mauvemeasuringgapneural}, next-token prediction accuracy, or multi-class classification accuracy, can improve through style matching without requiring recovery of specific facts or skills from the sensitive corpus. 
Such metrics are useful diagnostics, but they are poorly suited as measures of desired corpus-specific capability transfer. 
They conflate general linguistic competence with genuine information transfer, and may therefore overstate the practical utility of a DP synthetic corpus. Evaluating DP synthetic text requires tasks that directly test whether the model acquired the high-value knowledge or capabilities that motivated using the sensitive data in the first place.

\emph{Third}, synthetic data is often generated by a stronger model and consumed by a weaker one \citep{yue2023synthetic,kurakin2023harnessing,pmlr-v235-xie24g,amin-etal-2024-private}. This makes it difficult to separate gains due to the private corpus from gains due to ordinary distillation from a more capable teacher. Furthermore, such an evaluation does not answer whether sensitive data via unlocked via DP synthesis can improve the best model available today.

\begin{figure}[t]
    \centering
    \vspace{-0.5em}
    \includegraphics[width=0.95\linewidth]{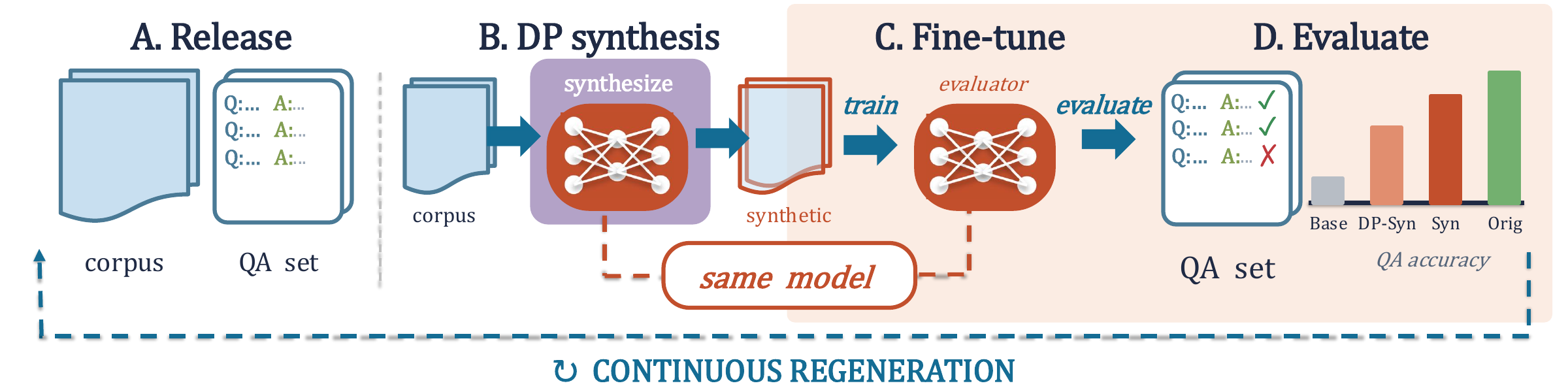}
    \caption{\textbf{Workflow of \cb.} Each release (\textbf{A}) pairs a freshly generated training corpus with a derived QA set. The participant runs their DP synthesis method (\textbf{B}) on the corpus to obtain synthetic data; our standardized harness (shaded) then fine-tunes (\textbf{C}) the same checkpoint used for synthesis on this data and scores it on the held-out QA test set (\textbf{D}). 
    Releases are regenerated periodically (bottom loop), keeping the benchmark contamination-free.
    }
    \label{fig:workflow}
\end{figure}

\efficientparagraph{Contributions} We introduce \cb (illustrated in Figure \ref{fig:workflow}), a benchmark designed specifically to remove the three aforementioned issues with existing evaluations. Quarterly, automatic releases will couple a never-before-seen training corpus with a derived question-answer (QA) set whose answers are supported by hundreds of independent records in the corpus, making them realistically learnable under DP constraints. In particular, we make the following contributions:
\begin{enumerate}%
    \item[•] \textbf{Continuously regenerated, access-dependent benchmark.} \cb enforces \emph{access dependency}: the tasks are unsolvable without access to the released training corpus. We implement this by testing the model on either fictional or post-cutoff knowledge. To guarantee access dependency for future model releases, \cb uses an automated curation pipeline allowing for periodic regeneration of new versions with minimal human intervention.

    \item[•] \textbf{Grounded factual QA as the evaluation task.} We focus on short-answer question answering tasks that cannot be solved with surface-level distribution matching. Questions are constructed so that the answer appears across hundreds of independent records, making them learnable under DP. Short-answer factual QA tasks, like MMLU \citep{hendryckstest2021}, TriviaQA~\citep{joshi2017triviaqalargescaledistantly}, and Natural Questions \citep{kwiatkowski-etal-2019-natural} allow unambiguous automatic evaluation and have served as reliable proxies for tracking progress in model capabilities. %

    \item[•]  \textbf{Frontier constraint to eliminate distillation confounds.} To rule out distillation from a stronger teacher, we enforce that identical base models (e.g., Gemma 3 1B PT) are used for generating the DP synthetic data as well as for measuring downstream task improvement. This ensures that any performance improvements arise strictly from the information captured from the sensitive data.

    \item[•]  \textbf{An improved understanding of the gaps between existing methods.} While existing evaluations suggest that most methods perform similarly, on \cb{}, we find that DP synthesis falls far short of non-private synthesis (Figure \ref{fig:teaser}), even at $\varepsilon=100$, which suggests a core open problem in DP text synthesis. We also study Private Evolution (PE) \citep{pmlr-v235-xie24g}, a popular training-free approach, and find that it significantly underperforms DP training -- training on PE-generated data does not improve over random guessing.

    \item[•]  \textbf{Standardized, open-source evaluation harness.} We release the corpus, QA sets, training recipes, and prompting and scoring scripts as an easy-to-use benchmarking package. This provides a reliable way to compare methods, and will help us measure and accelerate progress on DP synthetic data.
\end{enumerate}

\section{Background and problem formulation}
\label{sec:prelims}
\textbf{DP text synthesis.}
Differential privacy bounds how much any individual's data can affect the output of an algorithm. The privacy parameters $(\varepsilon, \delta)$ controls the strength of the guarantee, with smaller $(\varepsilon, \delta)$ corresponding to stronger privacy. In this work, we call any DP algorithm that releases a dataset a \emph{DP synthesis mechanism}. A common approach for text is to fine-tune a pretrained generator with DP-SGD~\citep{abadi2016dpsgd}, and then sample a synthetic corpus from the resulting model~\citep{yue2023synthetic,kurakin2023harnessing}. Compared to standard SGD, DP-SGD clips each per-example gradient, aggregates the clipped gradients, and adds calibrated noise before each update. See Section~5 of~\cite{ponomareva2025dpfy} for a broad overview of DP synthetic text. 

Table \ref{tab:related_dp_text_compact} compares prior work on DP text synthesis evaluation, in terms of the tasks used for evaluation, whether the evaluation is performed on fresh data relative to the base model's knowledge, and whether the generator and evaluator models were matched in scale. See Appendix \ref{app:related_work_detailed} for a more detailed discussion of each work.
\begin{table}[h]
    \centering
    \small
    \caption{
        DP text synthesis evaluation in existing work. \textbf{NTP} = next-token prediction; \textbf{Cls} = classification; \textbf{Sim} = distributional similarity. \textbf{Fresh} indicates whether the evaluated knowledge is constructed to be unavailable to the base model. \textbf{Matched} indicates that the generator and downstream evaluator are of comparable scale.
    }
    \label{tab:related_dp_text_compact}
    \begin{tabular}{llcc}
        \toprule
        \textbf{Study}
        & \textbf{Task}
        & \textbf{Fresh?}
        & \textbf{Matched?} \\
        \midrule
        \citet{yu2023training}
        & NTP
        & as of 2023\textsuperscript{1}
        & No \\
        \citet{kurakin2023harnessing}
        & Cls / Sim
        & No\textsuperscript{2}
        & No \\
        \citet{pmlr-v235-xie24g}
        & NTP / Cls / Sim
        & as of 2023\textsuperscript{1}
        & No \\
        \citet{amin-etal-2024-private}
        & Cls
        & No\textsuperscript{2}
        & No \\
        \citet{tan2025synthesizing}
        & NTP / Cls / Sim
        & No\textsuperscript{2}
        & No \\
        \citet{hu2025actgarl}
        & NTP / Cls / Sim
        & No
        & No \\
        \citet{sun2025synbench}
        & Cls / Sim
        & No
        & No \\
        \midrule
        \cb{}
        & QA
        & Yes + Regenerable
        & Yes \\
        \bottomrule
    \end{tabular}
\end{table}

\footnotetext[1]{
\citet{yu2023training} collected PubMed abstracts published between August 1--7, 2023, and \citet{pmlr-v235-xie24g} collected reviews for ICLR 2023 submissions from OpenReview.}

\footnotetext[2]{
They mitigate data contamination by retraining the base model from scratch on corpora curated to exclude the downstream task data, which are not publicly available. 
}

\textbf{Capability gain from training data.}
We study whether DP synthetic text can preserve the \emph{capability gain} that would have been obtained by training on the original corpus. In the ideal evaluation, one would re-run pretraining with and without the candidate dataset and measure the difference in downstream capability, which is prohibitively expensive. We therefore use \emph{continual pretraining} as a practical proxy: starting from a fixed pretrained checkpoint, we train on the candidate corpus and then evaluate the resulting model with a fixed set of eval prompts. In this setting, a dataset is useful if additional training on it improves the model's eval accuracy. \cb instantiates the source dataset to be synthesized, as well as the downstream training and evaluation process to facilitate the search for better DP synthesis mechanisms.

\efficientparagraph{Population-level knowledge, not singleton memorization}
A natural concern is that grounded QA rewards the type of memorization DP is designed to suppress. We agree that a DP benchmark should not reward recovery of facts that appear only once in a corpus.
It is equally clear that no reasonable parameterization of DP should erase a model's ability to recall information attested across hundreds or thousands of independent records.\footnote{For example: a model trained on DP synthetic clinical notes~\citep{wu2025term2note,hu2025actgarl}, should, at minimum, be able to recall basic details about conditions described across hundreds of independent records in order to be useful.} Information at this level of redundancy is population-level knowledge rather than individual-specific memorization, and a synthesis method that cannot recover it has not meaningfully preserved the corpus.

\cb therefore measures utility as a function of frequency rather than collapsing it to a binary, and uses short-answer QA as a minimal, automatically verifiable proxy for corpus-specific capability transfer, in the spirit of factual-recall benchmarks such as MMLU~\citep{hendryckstest2021,hendrycks2021ethics}, TriviaQA~\citep{joshi2017triviaqalargescaledistantly}, and Natural Questions~\citep{kwiatkowski-etal-2019-natural}. These benchmarks are routinely used as proxies for general model capability, on the premise that a model which cannot reliably retrieve well-attested facts is unlikely to be reasoning competently over them. We expand on this in Appendix~\ref{app:memorization}.

\section{\cb}\label{sec:benchmark}

\efficientparagraph{Benchmark overview} \cb is a benchmark and evaluation harness for (differentially private) text synthesis. 
Each \cb release is a versioned evaluation package containing: (i) a training corpus, (ii) a grounded short-answer QA set derived from that corpus, and (iii) a standardized harness for downstream training, prompting, decoding, normalization, and scoring. 
Section~\ref{subsec:cb_tracks} describes the concrete dataset tracks instantiated in this work.

\efficientparagraph{Benchmark Usage} Given a release, the participant selects a privacy regime and a target base model checkpoint, and then applies their DP synthesis method to the release corpus to obtain a synthetic training set. The synthetic training set is handed over to the \cb harness, which conducts continued pretraining on \emph{same base checkpoint} on that synthetic data using our fixed recipe. The resulting model is then evaluated on the release QA set. Figure \ref{fig:workflow} illustrates this workflow. For every configuration, we require three reference runs: \textbf{No training}, which evaluates the checkpoint without corpus training; \textbf{Train on real}, which fine-tunes on the original corpus; and \textbf{Syn}, which fine-tunes on synthetic data generated without privacy constraints ($\varepsilon=\infty$).

\paragraph{Benchmark design principles.} Each \cb release is produced by an automated pipeline that constructs a corpus and derives a matched QA set from it. We plan to refresh and release \cb quarterly. Each release is designed to satisfy three properties: 
\begin{enumerate}%
    \item \textbf{Freshness and access dependency.}
    QA items test knowledge unavailable to the base checkpoint but recoverable from the release corpus. We enforce freshness through post-cutoff collection or procedural regeneration, and validate access dependency by checking that the base checkpoint performs near chance while training on the original corpus yields a substantial gain.

    \item \textbf{Verifiable grounded QA.}
    Each QA item is linked to supporting corpus records and has a normalized short answer supporting automatic verifiablility.

    \item \textbf{DP learnability through many independent supporting records.}
    DP suppresses rare knowledge present in only a single record. Hence, for each eval QA, we track the number of independent records that contain the information needed to answer it, and evaluate on questions supported by hundreds of independent records as learnable population-level knowledge. %
\end{enumerate}

\subsection{Dataset tracks}
\label{subsec:cb_tracks}
\cb provides two complementary dataset tracks. \geminon is a fully controlled fictional domain that enables clean analyses of DP learnability. \news is a continuously collected real-world corpus, derived from Common-Crawl-News \citep{ccnews} that captures the messiness of natural text and time-evolving facts. Together, these tracks support both mechanism-level analysis and end-to-end evaluation.

\efficientparagraph{Track I: \geminon}
\geminon 
is a fully fictional, Pok\'emon-inspired world designed to make freshness procedural and unambiguous. Each release regenerates a new set of fictional entities together with attributes such as types, stats, and evolutions, along with a corpus describing them across multiple in-world formats. These formats are chosen to distribute each atomic fact across multiple independent records while avoiding trivial one-to-one paraphrases, enabling explicit control over the number of records containing a fact and textual diversity. From this corpus, we construct QA sets %
for both repeated facts and singleton facts. Since the domain is fully fictional, \geminon provides a synthetic setting for isolating how non-private and DP synthesis methods preserve fresh factual knowledge, without confounds from pretraining contamination or real-world extraction noise.

\begin{figure*}[ht]
\centering
\begin{minipage}[t]{0.35\textwidth}
\begin{minicard}{Index entry}
name~~~: Boreling\\
class.~: Frost Geminon\\
type~~~: ice\\
ability: Berserk\\
stats~~: HP 69, Atk 60, Def 63, SpA 67, SpD 68, Spe 40, BST 367\\
size~~~: 12 m, 52 lbs\\
evol.~~: Boreling, Borelash, Borastat\\
move~~~: Powder Snow
\end{minicard}
\end{minipage}%
\hfill
\begin{minipage}[t]{0.33\textwidth}
\begin{minicard}{Training article}
text: Boreling is an ice-type Geminon, known as the Frost Geminon, the first form of a three-stage evolution line that includes Borelash and Borastat. Physically, it is 12\ldots''\\
type: wiki\\
tag~: name, class., type, ability, all  stats, evol., size, move
\end{minicard}
\end{minipage}%
\hfill
\begin{minipage}[t]{0.3\textwidth}
\begin{minicard}{QAs}
question: What is the classification of Boreling?\\
answer~~: Frost Geminon\\
supports: 7410, 8929,\ldots\\[4pt]
question: What are the types of Boreling?\\
answer~~: Ice\\
supports: 3, 1515,\ldots\\[4pt]
\dots
\end{minicard}
\end{minipage}
\caption{\textsc{Geminon} example records. A procedurally generated \emph{index entry} (left), one of the articles (middle), and example QA pairs (right). \texttt{support} contains the article ids that support this answer.}
\label{fig:small-geminon-examples}
\end{figure*}

\efficientparagraph{Track II: \news}
\news 
evaluates DP synthesis in a realistic, noisy setting derived from contemporary news. 
For each release, we collect articles from a fixed post-cutoff window, clean and segment them into timestamped records, and deduplicate near-identical content while preserving natural event-level redundancy across outlets. Unlike \geminon, \news reflects real-world language, topical breadth, extraction errors, uneven event coverage, and occasional cross-source inconsistencies, making it a stress test for DP synthesis beyond controlled settings.

\begin{figure*}[ht]
\centering
\begin{minipage}[t]{0.48\textwidth}
\begin{minicard}{Source article}
url~~: www.hindustantimes.com/cricket/pak\ldots\\
date~: 2025-09-15\\
title: Pakistan knocks on ICC's door, demands removal of match referee\ldots\\
text~: Pakistan knocks on ICC's door, demands immediate removal of match referee for staying silent on India's no handshake. The PCB has escalated India's no-handshake controversy\ldots
\end{minicard}
\end{minipage}%
\hfill
\begin{minipage}[t]{0.48\textwidth}
\begin{minicard}{QA pair}
question~~: Who was the match referee for the India vs.\ Pakistan Asia Cup match on Sep 14, 2025?\\
answer~~~~: Andy Pycroft\\
\#supports~: 648\\
closedbook: Javagal Srinath \textcolor{red}{$\times$} \\
openbook~~: ``Not mentioned'' \textcolor{red}{$\times$} (art.\ 126152);\\
~~~~~~~~~~~~~~~``Andy Pycroft'' \textcolor{teal}{$\checkmark$} (art.\ 747116); \ldots
\end{minicard}
\end{minipage}
\caption{\textsc{News} example records. A CC-News source article (left) and one of the derived QA pairs (right). Each QA includes a closed-book baseline (Gemini answers without context) and open-book answers from retrieved candidate articles; article whose corresponding \texttt{closedbook} answer is correct is counted towards support}
\label{fig:news-examples}
\end{figure*}

\efficientparagraph{Estimating the number of records supporting a question}
The two tracks estimate support count differently. In \geminon, support count is controlled by construction or counted from recorded attribute mentions in generated articles. In \news, support count is estimated by verification. First, we cluster articles to identify events and generate QAs for each event. For each QA, we retrieve candidate supporting articles, and designate the article as supporting only if its counterfactual inclusion in a model's context causes it to start producing the correct answer. Since the procedure only tests retrieved candidate articles, the resulting support count estimate is a conservative lower bound.

Each dataset is released in \textbf{Large}, \textbf{Medium}, and \textbf{Small} variants, with train--validation--test splits. The corresponding QA sets are released with validation and test splits. 
In \geminon, each QA split contains two subsets: a \emph{high-repetition} subset, where each target fact appears in approximately 200 records, and a \emph{singleton} subset, where each target fact appears only once. Both subsets are derived from the same release corpus; the singleton subset is used only as a privacy sanity check and is not intended to be recoverable under DP. For \news, the canonical evaluation set keeps all questions with support count at least 200; we also report results on the full QA set as well as subsets with support count at least 400, 600, and 800 in Appendix \ref{app:experimental_details}. Full curation details, qualitative examples, corpus statistics, and QA statistics are provided in Appendices~\ref{app:geminon_details} and \ref{app:news_details} for the two tracks.

\subsection{Evaluation harness and metrics}
\label{subsec:cb_metrics}

Given a candidate corpus, either real or (DP) synthetic, \cb trains a \emph{fixed evaluator} initialized from the same base checkpoint as used for generation, using a prescribed continual-pretraining recipe. The resulting model is then evaluated on the target QA set. The harness fixes the downstream training recipe, checkpoint selection strategy, few-shot prompts for evaluation, sampling, and answer scoring. Thus, performance differences are intended to reflect the utility of the candidate corpus rather than downstream evaluation choices. Fixed downstream training enables cleaner, more compute-efficient, and reproducible evaluation. It also reflects the reality of data contributions to model training: pretraining is performed on a mixture of datasets and hyperparameters cannot be specifically for each individual component.

Final results are reported on the held-out QA test split using the track-specific primary metric: exact-match accuracy for \geminon, whose answers have canonical surface forms, and LLM-match accuracy for \news, whose answers may admit paraphrases or other free-form variation. The evaluator checkpoint is selected based on the best ``contains'' accuracy on the QA validation split, where a prediction is scored as correct when the normalized ground-truth answer is contained in the normalized model prediction. All QA metrics use the same deterministic response parsing and normalization before scoring. Additional details are presented in  Appendix~\ref{app:eval-details}. We release the full evaluation harness, including training recipes, prompts, sampling configurations, normalization rules, and scoring scripts at \url{https://github.com/plau666/ContinuousBenchEval}.

\section{Experimental results}\label{sec:experiment}
We organize our experiments around five questions: \textbf{(i)} Does the benchmark satisfy its design desiderata, namely freshness and DP learnability? \textbf{(ii)} How large is the capability gap between non-private and DP synthesis? \textbf{(iii)} How do training-free synthesis methods such as Private Evolution perform? 
\textbf{(iv)} Can direct DP fine-tuning close this gap? 
\textbf{(v)} How does the number of supporting records affect knowledge transferability?

\efficientparagraph{Datasets and training details} 
We evaluate on \textsc{Geminon-Small} and \textsc{News-Small}, each containing approximately 200K training examples. We use pretrained checkpoints from the \textsc{Gemma 3} family for both the generator and the evaluator. For synthesis, we train both public and DP generators on the training split of the original corpus using LoRA with rank 128. For DP generator training, we employ state-of-the-art, optimized DP-SGD configurations employing truncated poisson sampling \citep{chua2024scalable}, PLD accounting \citep{ganesh2025tighter} with $\delta = |\texttt{dataset}|^{-1.1}$, and normalized clipping \citep{de2022unlocking} with aggressive clipping norm \texttt{1e-3}. We follow recommendations from existing literature regarding the necessity of larger batch sizes and compute budgets than standard training \citep{anil2022largescale, li2022large, de2022unlocking, sander2023tan}. Specifically our DP runs use 16x larger batch sizes and $\approx$ 5.3x more accelerator hours and FLOPs than standard training. We use the highly-performant JAX Privacy \citep{mckenna2026jax} library for training.

For downstream evaluation, we train evaluator models on one of three data sources: the original corpus, a non-private synthetic corpus \textbf{Syn} ($\varepsilon=\infty$), or a DP synthetic corpus \textbf{DP-Syn} with $\varepsilon \in \{100,10\}$. We conduct both full-parameter and LoRA evaluator training. All generators and evaluators are trained with the standard causal language modeling objective in continual-pretraining style. Unless otherwise stated, we report exact-match accuracy for \geminon and LLM-match accuracy for \news using full-parameter trained evaluators. By default, QA results are evaluated on \geminon high-repetition QA test split, whose facts have approximately 200 repetitions, and the \news QA test subset with repetition at least 200. 
Full details on evaluator scoring, evaluator training, and generator training can be found  in Appendix~\ref{app:eval-details}.

\subsection{Validation of benchmark}\label{subsec:benchmark_validation}
We first validate that the benchmark satisfies two basic desiderata: (i) the evaluated facts are not already accessible to the base models, and (ii) the facts are learnable from the released corpus under our standardized training recipes. As shown in Table~\ref{tab:validation_main}, the base checkpoints achieve near-chance QA accuracy without access to the corpus, especially on \geminon. This indicates that the evaluated knowledge is not readily available from previous training stages. In contrast, training on the real corpus yields large gains for both datasets, confirming that the target knowledge is present in the corpus and can be acquired through training. These results validate the benchmark as a testbed for measuring whether synthetic data preserves learnable, corpus-specific knowledge.
\begin{table}[ht]%
    \centering
    \caption{
        Models obtain high \cb{} QA accuracy (\%) if and only if they train on the corpus. %
    }
    \label{tab:validation_main}
    \small
    \setlength{\tabcolsep}{5pt}
    \renewcommand{\arraystretch}{1.12}
    \begin{tabular}{l cc cc}
        \toprule
        & \multicolumn{2}{c}{\geminon (\%)} 
        & \multicolumn{2}{c}{\news (\%)} \\
        \cmidrule(lr){2-3} \cmidrule(lr){4-5}
        \textbf{Model}
        & \textbf{No training} & \textbf{Train on real}
        & \textbf{No training} & \textbf{Train on real} \\
        \midrule
        1B & 1.0 & 88.2 & 9.6  & 51.3 \\
        4B & 1.1 & 96.4 & 14.1 & 70.4 \\
        \bottomrule
    \end{tabular}
    \vspace{-1.5em}
\end{table}

\news results stratified by support count can be found in Table~\ref{tab:news_validation_by_support}, Appendix \ref{app:experimental_details}. Accuracy increases with support count and larger evaluator models learn repeated facts more reliably. Full exact-match, contains, and LLM-match results for all validation configurations are in Table~\ref{tab:validation_all_metrics}.

\subsection{The capability gap under differential privacy}\label{subsec:capability_gap}
\begin{figure}[ht]
    \centering
    \includegraphics[width=1\linewidth]{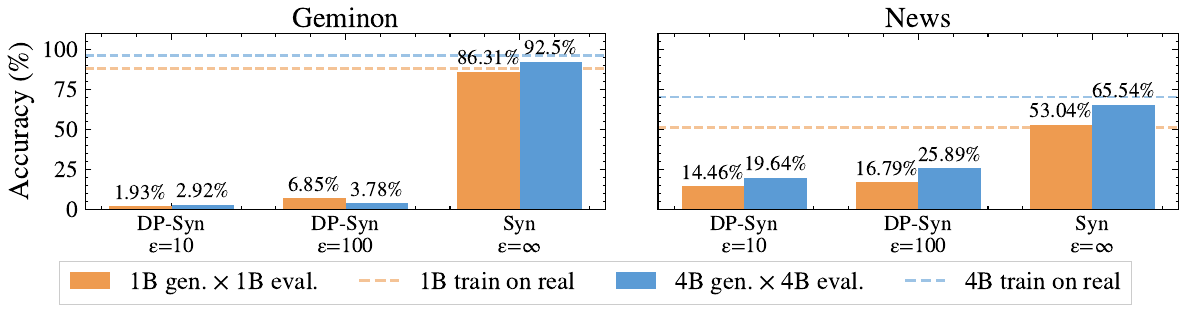}
    \caption{
        Capability gap between non-private and DP synthesis, measured by downstream QA accuracy. Bars show evaluators trained on DP synthetic data (\textbf{DP-Syn}) at $\varepsilon \in \{10, 100\}$ and on non-private synthetic data (\textbf{Syn}, $\varepsilon=\infty$), for matched generator-evaluator sizes (1B$\times$1B in orange, 4B$\times$4B in blue). Dashed horizontal lines mark the train-on-real-corpus reference.}

    \label{fig:dpsyn_acc_gap}
\end{figure}

We next address the second question: how much utility is lost when moving from non-private synthesis to DP synthesis? Figure~\ref{fig:dpsyn_acc_gap} shows a large capability gap, and that DP synthetic data fails to transfer corpus knowledge. Non-private synthesis (\textbf{Syn}, $\varepsilon=\infty$) transfers substantial factual knowledge from the original corpus: with matched 4B generator and 4B evaluator, downstream accuracy reaches 92.5\% on \geminon and 65.54\% on \news. The corresponding 1B setting also achieves strong performance, reaching 86.31\% on \geminon and 53.04\% on \news. Thus, synthetic data can serve as an effective medium for transferring corpus-specific knowledge when generated without privacy constraints.

However, when the synthetic corpus is generated by a DP fine-tuned generator (\textbf{DP-Syn}), this transfer sharply reduces this transfer. At $\varepsilon=100$, accuracy drops to 13.7\% on \geminon and 20.55\% on \news for the 4B setting, and to 6.85\% and 13.7\% for the 1B setting. At $\varepsilon=10$, performance falls further, reaching only 3.86\% on \geminon and 5.79\% on \news for the 4B setting. 

This capability gap is consistent across downstream evaluator training regimes (LoRA vs.\ full fine-tuning), generator--evaluator size configs (\{1B, 4B\} x \{1B, 4B\}), and datasets. This suggests that the main bottleneck lies in the DP-constrained synthetic corpus,  more specifically, in the DP fine-tuned synthesizer. Metrics for all training and generator--evaluator configs are in Tables~\ref{tab:all_acc_geminon} and~\ref{tab:all_acc_news}. 

One hypothesis for this gap relates to gradient clipping in DP-SGD. Rare tokens, such as names, abilities, or numeric stats, tend to have large gradient norms because they are infrequent and thus ``unexpected'' under the model's prior.  Per-example clipping disproportionately affects these high-norm gradients, causing the generator to underweight factual content relative to stylistic patterns. 

\begin{figure}[ht]
    \centering
    \includegraphics[width=1\linewidth]{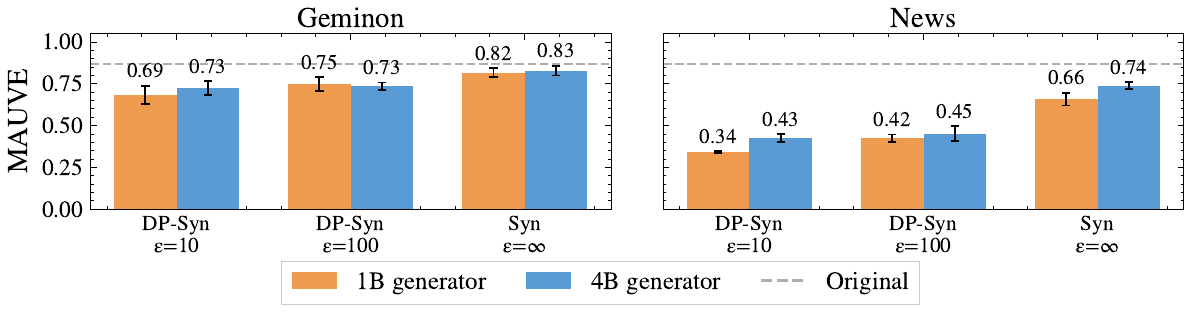}
    \caption{
    MAUVE scores for synthetic corpora, from either a 1B or 4B generator, against original corpora. The dashed gray line is the MAUVE score between disjoint splits of the original corpora.
}
    \label{fig:dpsyn_mauve}
\end{figure}

We further examine this distinction using MAUVE \citep{pillutla2021mauvemeasuringgapneural}, which measures distributional similarity between generated and original corpora. We use Gecko 110M embedding models \citep{lee2024geckoversatiletextembeddings} with 1K examples and report mean scores and standard deviation over 5 runs.
Figure~\ref{fig:dpsyn_mauve} shows that MAUVE decreases much more mildly than QA accuracy under DP synthesis. On \geminon, for example, the 4B generator achieves MAUVE 0.83 without privacy and 0.73 at $\varepsilon=10$, even though QA accuracy drops from 92.5\% to 3.86\%. The same pattern appears on \news. These results indicate that DP generators can preserve the \emph{distributional form} of the corpus---style, topic distribution, and document structure---while failing to preserve the \emph{specific factual content} required for grounded QA. Additional results are in Table \ref{tab:full_mauve}.

Together, these results show that distributional metrics alone are insufficient for evaluating DP text synthesis, for certain type of tasks. A method can obtain moderate MAUVE while transmitting little of the corpus-specific knowledge that makes the private data useful.

\subsection{Private evolution fails to transfer corpus knowledge}\label{subsec:private_evolution}

We next address our third question by evaluating the performance of Private Evolution for this task. Private Evolution (PE) is a training-free DP synthesis method that has achieved strong results on existing benchmarks \citep{pmlr-v235-xie24g}, and has been the subject of significant follow-up work \citep{zhang2025pcevolve, gonzalez2026private, hou2024pretext}. 
We test it on \cb{}. We follow the recipe of~\citep{pmlr-v235-xie24g}, using 10 PE iterations with 7 paraphrases per example at $\varepsilon=10$. We use \textsc{Gemma 3 1B PT} as the generator. (In Appendix \ref{subsec:pe_details} we also evaluate \textsc{Gemma 3 1B IT} to test whether stronger instruction following improves PE performance; it does not). For each configuration, we choose the sampling temperature with the lowest FID\footnote{FID is the main distributional similarity metric reported by PE \citep{pmlr-v235-xie24g}; lower values indicate that generated samples are closer to the target corpus in the embedding space used by the metric.}  and generate a 20K-example PE corpus.\footnote{Due to computational reasons, since there is a $7\times 10$ sampling overhead for each additional example. Although \textsc{Geminon-Small} and \textsc{News-Small} each contain 200K training examples, this 20K-example scale is comparable to prior Aug-PE evaluations: Xie et al.~\citep{pmlr-v235-xie24g} generated at most 100K examples for Yelp~\citep{yelp_dataset}, whose source corpus contains 1.7M examples, and 5K examples for PubMed~\citep{yu2023training}, whose source corpus contains 75K examples.} We then train downstream 1B evaluators on the resulting PE corpora with the same standardized evaluation pipeline. 

\begin{wraptable}{r}{0.5\textwidth}
    \vspace{-1em}
    \centering
    \caption{
        Private Evolution results at $\varepsilon=10$.
    }
    \label{tab:main_pe_eps10}
    \small
    \setlength{\tabcolsep}{5pt}
    \renewcommand{\arraystretch}{1.08}
    \resizebox{0.9\linewidth}{!}{%
    \begin{tabular}{l ccc ccc}
        \toprule
        & \multicolumn{2}{c}{\geminon} 
        & \multicolumn{2}{c}{\news} \\
        \cmidrule(lr){2-3} \cmidrule(lr){4-5}
        \textbf{Method}
        & \textbf{FID}
        & \textbf{QA (\%)}        & \textbf{FID}
        & \textbf{QA (\%)} \\
        \midrule
        1B No training & - & 1.0%
            & - & 9.6 \\
        \midrule
        1B PE $(\varepsilon=10)$
            & 33.54 & 0.2 
            & 27.44 & \textit{9.2} \\
        \bottomrule
    \end{tabular}%
    }
\end{wraptable}

Table \ref{tab:main_pe_eps10} shows that evaluators trained on PE data perform worse than untrained. Although PE produces corpora with reasonable FID scores, e.g., 27.4 for \news and 33.5 for \geminon, downstream QA accuracy remains near zero (0.2\%) on \geminon and never exceeds the base checkpoint on \news.\footnote{We also observe that, when training evaluators on PE-generated \geminon data, cross-entropy on the original \geminon validation corpus consistently increases while training loss decreases. On \news,  accuracy decreases over training; see Figure~\ref{fig:pe_news_acc_dynamics}. This suggests that the failure is not primarily explained by the smaller number of synthetic examples.}
As noted by~\cite{pmlr-v235-xie24g}, PE relies on knowledge already present in the language model to generate high-quality candidate texts without domain-specific fine-tuning. In our setting, however, the target facts are fresh: the base model does not already know the relevant \geminon entities and attributes or post-cutoff \news events. DP selection can therefore choose candidates that are stylistically close to the private corpus, but it cannot reliably recover facts that are absent from the generator's proposal distribution.

\subsection{Direct DP fine-tuning does not close the gap}\label{subsec:direct_dpft}

We next ask where the DP synthesis gap comes from. A DP synthetic-data pipeline can fail for two reasons: the private generator may fail to learn the relevant facts, or the learned facts may be lost when sampling a synthetic corpus and retraining a downstream evaluator. To separate these effects, we evaluate a more direct baseline: DP fine-tuning the evaluator itself on the original corpus.

This baseline removes the synthetic-data mediation step, but downgrades the release object from a DP synthetic dataset to a DP model. It is therefore not a replacement for DP data synthesis; rather, it serves as a diagnostic upper bound on \emph{what the same privacy budget allows a model to learn when trained directly on the original examples.} If fresh QA facts cannot be learned even in this setting, then the failure of DP synthesis cannot be explained solely by sampling artifacts or imperfect transmission through synthetic data.

\begin{figure}[ht]
    \centering
        \includegraphics[width=1.0\linewidth]{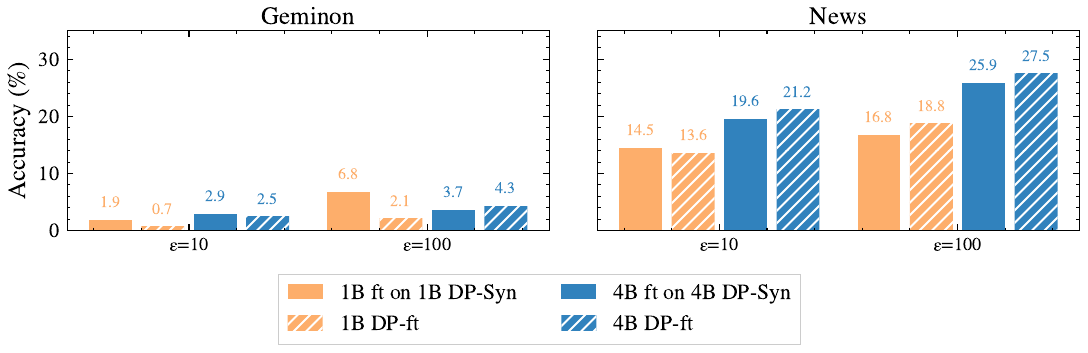}
        \caption{%
Evaluator training on DP-synthetic data vs. direct DP fine-tuning on the original corpus.   
        }
    \label{fig:direct_dpsgd}
\end{figure}

Figure~\ref{fig:direct_dpsgd} and Table~\ref{tab:direct_dpsgd} show that direct DP fine-tuning performs in the same range as the DP-synthesis pipeline, and remains far below non-private synthesis. This indicates that synthetic-data mediation is not the only source of degradation: even when differentially privately training directly on the original corpus, it struggles to encode the fresh factual signal needed for QA. The gap is especially pronounced on \geminon, where non-private synthesis achieves high accuracy, but direct DP fine-tuning remains near the DP-synthesis baseline at both privacy budgets. 

Overall, direct DP fine-tuning slightly narrows the gap but does not close it. This suggests that improving DP synthetic text for factual knowledge transfer will require progress at both stages: better private learning that retains rare factual content, and better synthesis procedures that faithfully transmit whatever factual signal the private model has learned.

\subsection{The number of supporting records affects knowledge transferability}

A crucial feature of \cb is that each test QA is paired with its \emph{support count} -- the number of independent training records that contain the information required to answer it correctly. We use this to ensure that our tasks are solvable under DP; but furthermore, it allows us to investigate the performance profile of a given synthesis method on questions of varying support counts.

For \geminon, we evaluate on the singleton set, which consists of questions whose answers are, by design, present in only a single corpus record. Hence, DP should strongly suppress their learnability. Consistent with this expectation, downstream evaluators perform poorly on the sensitive split, as shown in Table~\ref{tab:sensitive_qas}. Full results on \geminon singleton set are reported in Table~\ref{tab:full_sensitive_qas_by_1b}.

\begin{table}[h]
    \centering
    \caption{
        Singleton set accuracy $(\%)$ on \geminon for facts that appear in only one record in the corpus.}
    \label{tab:sensitive_qas}
    \small
    \setlength{\tabcolsep}{5pt}
    \renewcommand{\arraystretch}{1.12}
    \begin{tabular}{l c c c c c}
        \toprule
        Model
        & No Training
        & Train on real
        & \textbf{Syn}
        & \textbf{DP-Syn}$@\varepsilon=100$
        & \textbf{DP-Syn}$@\varepsilon=10$ \\
        \midrule
        1B
        & 0.6
        & 3.0
        & 1.2
        & 0.8
        & 0.6 \\
        4B
        & 0.9
        & 5.0
        & 2.1
        & 0.8
        & 0.8 \\
        \bottomrule
    \end{tabular}
\end{table}

\begin{figure}[ht]
  \centering
  \includegraphics[width=\linewidth]{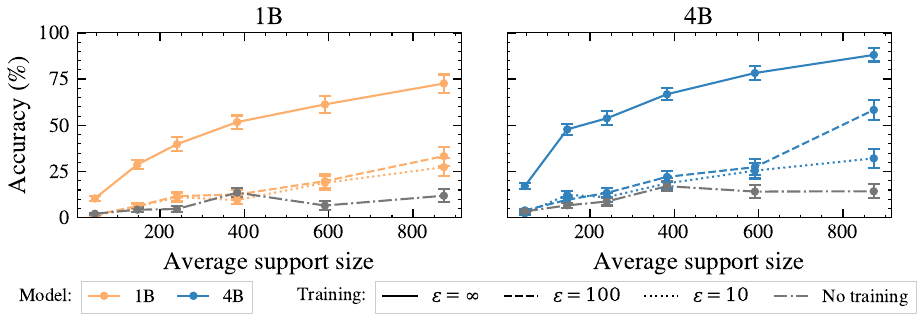}
  \caption{
    Accuracy (\%) on \news as a function of QA support count. QAs are bucketed by support count; we plot the average support count and accuracy inside each bucket. Error bars show standard deviation. 
  }
  \label{fig:news_support_vs_accuracy}
\end{figure}

For \news, the number of records supporting a question varies naturally across our evaluation set. Figure \ref{fig:news_support_vs_accuracy} shows the effect of support count on accuracy, via bucketing questions by support count and plotting the average average accuracy inside the bucket. We see that the gap between DP synthetic and ``No training'' emerges around $k\approx 600$. Full results for \news accuracy by support count are reported in Tables~\ref{tab:news_validation_by_support}, ~\ref{tab:news_syn_by_support_1b_generator}, ~\ref{tab:news_syn_by_support_4b_generator} in Appendix \ref{app:experimental_details}.

The training dynamics support the same conclusion. For \geminon, high-repetition facts are learned rapidly from the real corpus and non-private synthetic data, while DP-synthetic training remains near the floor; the singleton split remains low throughout training. For \news, higher-support buckets improve earlier and reach higher final accuracy across training regimes, but DP-synthetic training remains substantially below real-corpus and non-private synthetic training. We provide a more detailed empirical analysis of the learning dynamics in Appendix~\ref{app:learning_dynamics}.

\section{Related work}

\label{sec:related_work}
\textbf{DP synthetic text.} A growing literature studies DP text synthesis, including API-only private evolution~\citep{pmlr-v235-xie24g}, private prediction and inference-time methods~\citep{amin-etal-2024-private}, clustering-guided fine-tuning~\citep{tan2025synthesizing}, and RL-based controllable generation~\citep{hu2025actgarl}, among others~\citep{yu2023training,ramesh-etal-2024-evaluating,kurakin2023harnessing,ponomareva2025dpfy}. Across these works, evaluation typically relies on classification accuracy or next-token prediction on a small set of standard benchmarks, often with a mismatch between the generator and the downstream evaluator. We identify three recurring issues that limit existing evaluations: (i)~evaluated knowledge may already reside in the base model from pretraining, (ii)~the downstream tasks are too simple to measure corpus-specific capability transfer, and (iii)~teacher-student gaps make it difficult to attribute gains to the private corpus rather than to distillation. \cb is designed to address all three; we provide a detailed positioning relative to each prior work in Appendix~\ref{app:related_work_detailed}.

\textbf{DP synthesis benchmarks.} A small number of benchmarks target the evaluation of (DP) synthetic data generation. Struct-Bench~\citep{wang2025structbench} and \citet{chen2025benchmarking} focus on structured and tabular data; DPImageBench~\citep{gong2025dpimagebenchunifiedbenchmarkdifferentially} standardizes evaluation for DP synthetic images by fixing downstream tasks and comparing generation methods. In the text domain, \citet{sun2025evaluating} benchmark DP text generation using similarity metrics and downstream classification utility. These efforts are valuable, but none addresses the three evaluation challenges motivating our work: ensuring freshness of the evaluated knowledge, using tasks that measure \textit{capability transfer} rather than surface similarity, and controlling for teacher-student mismatch.

\textbf{Contamination-resistant and data-centric benchmarks.} AntiLeakBench~\citep{wu2025antileakbench} and FreshQA~\citep{vu-etal-2024-freshllms} construct continually updated QA sets to reduce contamination, but they are not packaged as matched train-test releases for measuring learning from a released corpus. StreamingQA~\citep{pmlr-v162-liska22a} provides a timestamped corpus and QA set, but its news stream largely predates modern LLM training cutoffs. TOFU~\citep{maini2024tofu} is synthetically constructed and conceptually adjacent to our setting, but lacks the knowledge repetition needed to distinguish DP-learnable population facts from singleton facts and is not continuously regenerated. None of these benchmarks were designed to evaluate (DP) synthesis methods.
Our benchmark design draws on data-centric evaluation principles. DataComp-LM argues that when the object of study is the data itself, downstream training, compute, and evaluation should be held fixed so that performance differences are attributable to the data~\citep{li2024datacomp}. WildBench emphasizes explicit benchmark scope and dataset criteria~\citep{lin2024wildbench}, while BetterBench distills benchmark construction into concrete best practices~\citep{reuel2024betterbench}. \cb adopts this philosophy for DP synthetic text, combining fresh corpus-specific capability transfer, grounded QA, and frontier-matched evaluation in a single continuously regenerated package.

\section{Conclusion}
We introduce \cb{}, a continuously regenerated benchmark and evaluation harness for measuring whether DP synthetic text preserves the capability gains that access to a sensitive corpus would have provided. By coupling a fresh training corpus with grounded QA, enforcing matched generators and evaluators, while ensuring DP learnability, \cb{} resolves issues present in previous, largely saturated evaluations.

Our experiments reveal a stark gap. Non-private synthesis transfers substantial corpus-specific knowledge, while state-of-the-art DP methods for text synthesis do not, even at $\varepsilon=100$, despite producing corpora that look similar to the original. Closing this gap is, in our view, the central open problem for DP text synthesis: a method that cannot recover information attested across hundreds of independent records has not meaningfully preserved a corpus. Since thoughtfully designed and implemented benchmarks drive progress in machine learning, we hope that \cb{} can serve as a reproducible yardstick for measuring and accelerating progress in DP synthetic data.

\begin{ack}
Thanks to Arun Ganesh for helpful pointers on how to set parameters for truncated poisson sampling. We thank Thomas Steinke for helpful discussions. We also thank the developers of JAX privacy, a library we make heavy use of in this work.
G.K. was supported by a Canada CIFAR AI Chair, an NSERC Discovery Grant, and an Ontario Early Researcher Award. R.C., P.L., and R.G. supported in part by the NSF Center for Smart Streetscapes (CS3) under NSF Cooperative Agreement No. EEC-2133516.
\end{ack}

\bibliographystyle{plainnat}
\bibliography{ref}

@InProceedings{pmlr-v235-tramer24a,
  title = 	 {Position: Considerations for Differentially Private Learning with Large-Scale Public Pretraining},
  author = 	 {Tram\`{e}r, Florian and Kamath, Gautam and Carlini, Nicholas},
  booktitle = 	 {Proceedings of the 41st International Conference on Machine Learning},
  pages = 	 {48453--48467},
  year = 	 {2024},
  volume = 	 {235},
  series = 	 {Proceedings of Machine Learning Research},
  publisher =    {PMLR},
  url = 	 {https://proceedings.mlr.press/v235/tramer24a.html}
}

@misc{lee2024geckoversatiletextembeddings,
      title={Gecko: Versatile Text Embeddings Distilled from Large Language Models}, 
      author={Jinhyuk Lee and Zhuyun Dai and Xiaoqi Ren and Blair Chen and Daniel Cer and Jeremy R. Cole and Kai Hui and Michael Boratko and Rajvi Kapadia and Wen Ding and Yi Luan and Sai Meher Karthik Duddu and Gustavo Hernandez Abrego and Weiqiang Shi and Nithi Gupta and Aditya Kusupati and Prateek Jain and Siddhartha Reddy Jonnalagadda and Ming-Wei Chang and Iftekhar Naim},
      year={2024},
      eprint={2403.20327},
      archivePrefix={arXiv},
      primaryClass={cs.CL},
      url={https://arxiv.org/abs/2403.20327}, 
}

@misc{pillutla2021mauvemeasuringgapneural,
      title={MAUVE: Measuring the Gap Between Neural Text and Human Text using Divergence Frontiers}, 
      author={Krishna Pillutla and Swabha Swayamdipta and Rowan Zellers and John Thickstun and Sean Welleck and Yejin Choi and Zaid Harchaoui},
      year={2021},
      eprint={2102.01454},
      archivePrefix={arXiv},
      primaryClass={cs.CL},
      url={https://arxiv.org/abs/2102.01454}, 
}

@inproceedings{yu2023training,
  title={Training Private and Efficient Language Models with Synthetic Data from LLMs},
  author={Yu, Da and Backurs, Arturs and Gopi, Sivakanth and Inan, Huseyin and Kulkarni, Janardhan and Lin, Zinan and Xie, Chulin and Zhang, Huishuai and Zhang, Wanrong},
  booktitle={NeurIPS 2023 Workshop on Socially Responsible Language Modelling Research (SoLaR)},
  year={2023},
  url={https://openreview.net/forum?id=FKwtKzglFb}
}

@InProceedings{pmlr-v235-xie24g,
  title = 	 {Differentially Private Synthetic Data via Foundation Model {API}s 2: Text},
  author =       {Xie, Chulin and Lin, Zinan and Backurs, Arturs and Gopi, Sivakanth and Yu, Da and Inan, Huseyin A and Nori, Harsha and Jiang, Haotian and Zhang, Huishuai and Lee, Yin Tat and Li, Bo and Yekhanin, Sergey},
  booktitle = 	 {Proceedings of the 41st International Conference on Machine Learning},
  pages = 	 {54531--54560},
  year = 	 {2024},
  editor = 	 {Salakhutdinov, Ruslan and Kolter, Zico and Heller, Katherine and Weller, Adrian and Oliver, Nuria and Scarlett, Jonathan and Berkenkamp, Felix},
  volume = 	 {235},
  series = 	 {Proceedings of Machine Learning Research},
  month = 	 {21--27 Jul},
  publisher =    {PMLR},
  url = 	 {https://proceedings.mlr.press/v235/xie24g.html},
}

@article{reuel2024betterbench,
  title={{BetterBench}: Assessing ai benchmarks, uncovering issues, and establishing best practices},
  author={Reuel, Anka and Hardy, Amelia and Smith, Chandler and Lamparth, Max and Hardy, Malcolm and Kochenderfer, Mykel J},
  journal={Advances in Neural Information Processing Systems},
  volume={37},
  pages={21763--21813},
  year={2024}
}

@article{lin2024wildbench,
  title={{WildBench}: Benchmarking llms with challenging tasks from real users in the wild},
  author={Lin, Bill Yuchen and Deng, Yuntian and Chandu, Khyathi and Brahman, Faeze and Ravichander, Abhilasha and Pyatkin, Valentina and Dziri, Nouha and Bras, Ronan Le and Choi, Yejin},
  journal={arXiv preprint arXiv:2406.04770},
  year={2024}
}

@article{li2024datacomp,
  title={{DataComp-LM}: In search of the next generation of training sets for language models},
  author={Li, Jeffrey and Fang, Alex and Smyrnis, Georgios and Ivgi, Maor and Jordan, Matt and Gadre, Samir and Bansal, Hritik and Guha, Etash and Keh, Sedrick and Arora, Kushal and others},
  journal={Advances in Neural Information Processing Systems},
  volume={37},
  pages={14200--14282},
  year={2024}
}

@inproceedings{amin-etal-2024-private,
    title = "Private prediction for large-scale synthetic text generation",
    author = "Amin, Kareem and Bie, Alex and Kong, Weiwei and Kurakin, Alexey and Ponomareva, Natalia and Syed, Umar and Terzis, Andreas and Vassilvitskii, Sergei",
    booktitle = "Findings of the Association for Computational Linguistics: EMNLP 2024",
    year = "2024",
    address = "Miami, Florida, USA",
    publisher = "Association for Computational Linguistics",
    url = "https://aclanthology.org/2024.findings-emnlp.425",
    pages = "7244--7262",
}

@InProceedings{tan2025synthesizing,
  title = 	 {Synthesizing Privacy-Preserving Text Data via Finetuning without Finetuning Billion-Scale LLMs},
  author = 	 {Tan, Bowen and Xu, Zheng and Xing, Eric and Hu, Zhiting and Wu, Shanshan},
  booktitle = 	 {Proceedings of the 42nd International Conference on Machine Learning},
  year = 	 {2025},
  series = 	 {Proceedings of Machine Learning Research},
  publisher =    {PMLR},
  url = 	 {https://arxiv.org/abs/2503.12347}
}

@inproceedings{wang2025structbench,
  title={{Struct-Bench}: A Benchmark for Differentially Private Structured Text Generation},
  author={Wang, Shuaiqi and Raunak, Vikas and Backurs, Arturs and Reis, Victor and Zhou, Pei and Chen, Sihao and Yang, Longqi and Lin, Zinan and Yekhanin, Sergey and Fanti, Giulia},
  booktitle={Advances in Neural Information Processing Systems (NeurIPS) 2025 Datasets and Benchmarks Track},
  year={2025},
  url={https://openreview.net/forum?id=59vXWteYuh}
}

@misc{hu2025actgarl,
      title={{ACTG-ARL}: Differentially Private Conditional Text Generation with RL-Boosted Control}, 
      author={Yuzheng Hu and Ryan McKenna and Da Yu and Shanshan Wu and Han Zhao and Zheng Xu and Peter Kairouz},
      year={2025},
      eprint={2510.18232},
      archivePrefix={arXiv},
      primaryClass={cs.LG},
      doi={10.48550/arXiv.2510.18232},
      url={https://arxiv.org/abs/2510.18232}
}

@misc{ponomareva2025dpfy,
      title={{How to DP-fy Your Data}: A Practical Guide to Generating Synthetic Data With Differential Privacy}, 
      author={Natalia Ponomareva and Zheng Xu and H. Brendan McMahan and Peter Kairouz and Lucas Rosenblatt and Vincent Cohen-Addad and Cristóbal Guzmán and Ryan McKenna and Galen Andrew and Alex Bie and Da Yu and Alex Kurakin and Morteza Zadimoghaddam and Sergei Vassilvitskii and Andreas Terzis},
      year={2025},
      eprint={2512.03238},
      archivePrefix={arXiv},
      primaryClass={cs.CR},
      doi={10.48550/arXiv.2512.03238},
      url={https://arxiv.org/abs/2512.03238}
}

@misc{kurakin2023harnessing,
      title={Harnessing large-language models to generate private synthetic text}, 
      author={Alexey Kurakin and Natalia Ponomareva and Umar Syed and Liam MacDermed and Andreas Terzis},
      year={2023},
      eprint={2306.01684},
      archivePrefix={arXiv},
      primaryClass={cs.LG},
      doi={10.48550/arXiv.2306.01684},
      url={https://arxiv.org/abs/2306.01684}
}

@misc{maca11_all_pokemon_dataset,
  author       = {maca11},
  title        = {All Pokemon Dataset},
  year         = {2021},
  publisher    = {Kaggle},
  howpublished = {\url{https://www.kaggle.com/datasets/maca11/all-pokemon-dataset}},
}

@misc{amancio_pokemon_moves_2024,
  author       = {Thiago Amancio},
  title        = {Full Pokemons and Moves Datasets},
  year         = {2024},
  publisher    = {Kaggle},
  howpublished = {\url{https://www.kaggle.com/datasets/thiagoamancio/full-pokemons-and-moves-datasets}},
}

@inproceedings{wu2025antileakbench,
  title={{AntiLeakBench}: Preventing Data Contamination by Automatically Constructing Benchmarks with Updated Real-World Knowledge},
  author={Wu, Xiaobao and Pan, Liangming and Xie, Yuxi and Zhou, Ruiwen and Zhao, Shuai and Ma, Yubo and Du, Mingzhe and Mao, Rui and Luu, Anh Tuan and Wang, William Yang},
  booktitle={Proceedings of the 63rd Annual Meeting of the Association for Computational Linguistics (ACL)},
  year={2025},
  url={https://arxiv.org/abs/2412.13670}
}

@inproceedings{vu-etal-2024-freshllms,
    title = "{F}resh{LLM}s: Refreshing Large Language Models with Search Engine Augmentation",
    author = "Vu, Tu  and
      Iyyer, Mohit  and
      Wang, Xuezhi  and
      Constant, Noah  and
      Wei, Jerry  and
      Wei, Jason  and
      Tar, Chris  and
      Sung, Yun-Hsuan  and
      Zhou, Denny  and
      Le, Quoc  and
      Luong, Thang",
    booktitle = "Findings of the Association for Computational Linguistics: ACL 2024",
    year = "2024",
    address = "Bangkok, Thailand",
    publisher = "Association for Computational Linguistics",
    url = "https://aclanthology.org/2024.findings-acl.813",
    doi = "10.18653/v1/2024.findings-acl.813",
    pages = "13697--13720",
}

@inproceedings{maini2024tofu,
  title={{TOFU}: A Task of Fictitious Unlearning for {LLM}s},
  author={Maini, Pratyush and Feng, Zhili and Schwarzschild, Avi and Lipton, Zachary C. and Kolter, J. Zico},
  booktitle={Conference on Language Modeling (COLM)},
  year={2024},
  url={https://openreview.net/forum?id=B41hNBoWLo}
}

@InProceedings{pmlr-v162-liska22a,
  title = 	 {{S}treaming{QA}: A Benchmark for Adaptation to New Knowledge over Time in Question Answering Models},
  author =       {Liska, Adam and Kocisky, Tomas and Gribovskaya, Elena and Terzi, Tayfun and Sezener, Eren and Agrawal, Devang and De Masson D'Autume, Cyprien and Scholtes, Tim and Zaheer, Manzil and Young, Susannah and Gilsenan-Mcmahon, Ellen and Austin, Sophia and Blunsom, Phil and Lazaridou, Angeliki},
  booktitle = 	 {Proceedings of the 39th International Conference on Machine Learning},
  pages = 	 {13604--13622},
  year = 	 {2022},
  volume = 	 {162},
  series = 	 {Proceedings of Machine Learning Research},
  publisher =    {PMLR},
  url = 	 {https://proceedings.mlr.press/v162/liska22a.html}
}

@inproceedings{sun2025evaluating,
  title={Evaluating Differentially Private Generation of Domain-Specific Text},
  author={Sun, Yidan and Schlegel, Viktor and Nandakumar, Srinivasan and Zahid, Iqra and Wu, Yuping and Del-Pinto, Warren and Nenadic, Goran and Lam, Siew-Kei and Zhang, Jie and Bharath, Anil A.},
  booktitle={Proceedings of the 34th ACM International Conference on Information and Knowledge Management (CIKM '25)},
  pages={5273--5278},
  year={2025},
  publisher={Association for Computing Machinery},
  address={New York, NY, USA},
  doi={10.1145/3627673.3680074},
  url={https://dl.acm.org/doi/10.1145/3627673.3680074}
}

@inproceedings{ramesh-etal-2024-evaluating,
    title = "Evaluating Differentially Private Synthetic Data Generation in High-Stakes Domains",
    author = "Ramesh, Krithika  and
      Gandhi, Nupoor  and
      Madaan, Pulkit  and
      Bauer, Lisa  and
      Peris, Charith  and
      Field, Anjalie",
    booktitle = "Findings of the Association for Computational Linguistics: EMNLP 2024",
    month = nov,
    year = "2024",
    address = "Miami, Florida, USA",
    publisher = "Association for Computational Linguistics",
    url = "https://aclanthology.org/2024.findings-emnlp.894",
    doi = "10.18653/v1/2024.findings-emnlp.894",
    pages = "15254--15269",
}

@article{ponomareva2023how,
  title={How to {DP}-fy {ML}: A Practical Guide to Machine Learning with Differential Privacy},
  author={Ponomareva, Natalia and Hazimeh, Hussein and Kurakin, Alex and Xu, Zheng and Denison, Carson and McMahan, H. Brendan and Vassilvitskii, Sergei and Chien, Steve and Thakurta, Abhradeep},
  journal={Journal of Artificial Intelligence Research},
  volume={77},
  pages={1113--1201},
  year={2023},
  doi={10.1613/jair.1.14649},
  url={https://doi.org/10.1613/jair.1.14649}
}

@inproceedings{yu2022differentially,
  title={Differentially Private Fine-tuning of Language Models},
  author={Yu, Da and Naik, Saurabh and Backurs, Arturs and Gopi, Sivakanth and Inan, Huseyin A and Kamath, Gautam and Kulkarni, Janardhan and Lee, Yin Tat and Manoel, Andre and Wutschitz, Lukas and Yekhanin, Sergey and Zhang, Huishuai},
  booktitle={International Conference on Learning Representations (ICLR)},
  year={2022},
  url={https://openreview.net/forum?id=Q42f0dfjECO}
}

@InProceedings{pmlr-v267-mckenna25a,
  title = 	 {Scaling Laws for Differentially Private Language Models},
  author = 	 {Mckenna, Ryan and Huang, Yangsibo and Sinha, Amer and Balle, Borja and Charles, Zachary and Choquette-Choo, Christopher A. and Ghazi, Badih and Kaissis, Georgios and Kumar, Ravi and Liu, Ruibo and Yu, Da and Zhang, Chiyuan},
  booktitle = 	 {Proceedings of the 42nd International Conference on Machine Learning},
  pages = 	 {43375--43398},
  year = 	 {2025},
  volume = 	 {267},
  series = 	 {Proceedings of Machine Learning Research},
  publisher =    {PMLR},
  url = 	 {https://proceedings.mlr.press/v267/mckenna25a.html}
}

@misc{comanici2025gemini25pushingfrontier,
      title={Gemini 2.5: Pushing the Frontier with Advanced Reasoning, Multimodality, Long Context, and Next Generation Agentic Capabilities}, 
      author={{Gemini Team}},
      year={2025},
      eprint={2507.06261},
      archivePrefix={arXiv},
      primaryClass={cs.CL},
      url={https://arxiv.org/abs/2507.06261}, 
}

@misc{embeddinggemma,
      title={{EmbeddingGemma}: Powerful and Lightweight Text Representations}, 
      author={EmbeddingGemma Team},
      year={2025},
      eprint={2509.20354},
      archivePrefix={arXiv},
      primaryClass={cs.CL},
      url={https://arxiv.org/abs/2509.20354}, 
}

@article{chen2025benchmarking,
author = {Chen, Kai and Li, Xiaochen and Gong, Chen and Mckenna, Ryan and Wang, Tianhao},
title = {Benchmarking Differentially Private Tabular Data Synthesis: [Experiments \& Analysis]},
year = {2025},
issue_date = {December 2025},
publisher = {Association for Computing Machinery},
address = {New York, NY, USA},
volume = {3},
number = {6},
url = {https://doi.org/10.1145/3769764},
doi = {10.1145/3769764},
journal = {Proc. ACM Manag. Data},
month = dec,
articleno = {299},
numpages = {25}
}

@misc{gong2025dpimagebenchunifiedbenchmarkdifferentially,
      title={{DPImageBench}: A Unified Benchmark for Differentially Private Image Synthesis}, 
      author={Chen Gong and Kecen Li and Zinan Lin and Tianhao Wang},
      year={2025},
      eprint={2503.14681},
      archivePrefix={arXiv},
      primaryClass={cs.CR},
      url={https://arxiv.org/abs/2503.14681}, 
}

@article{mckenna2026jax,
  title={{JAX-Privacy}: A library for differentially private machine learning},
  author={McKenna, Ryan and Andrew, Galen and Balle, Borja and Doroshenko, Vadym and Ganesh, Arun and Kong, Weiwei and Kurakin, Alex and McMahan, Brendan and Pravilov, Mikhail},
  journal={arXiv preprint arXiv:2602.17861},
  year={2026}
}

@inproceedings{cheng2024instruction,
  title={Instruction pre-training: Language models are supervised multitask learners},
  author={Cheng, Daixuan and Gu, Yuxian and Huang, Shaohan and Bi, Junyu and Huang, Minlie and Wei, Furu},
  booktitle={Proceedings of the 2024 Conference on Empirical Methods in Natural Language Processing},
  pages={2529--2550},
  year={2024}
}

@article{ovadia2025knowledge,
  title={Knowledge-instruct: Effective continual pre-training from limited data using instructions},
  author={Ovadia, Oded and Brief, Meni and Lemberg, Rachel and Sheetrit, Eitam},
  journal={arXiv preprint arXiv:2504.05571},
  year={2025}
}

@InProceedings{IMDB,
  author    = {Maas, Andrew L.  and  Daly, Raymond E.  and  Pham, Peter T.  and  Huang, Dan  and  Ng, Andrew Y.  and  Potts, Christopher},
  title     = {Learning Word Vectors for Sentiment Analysis},
  booktitle = {Proceedings of the 49th Annual Meeting of the Association for Computational Linguistics: Human Language Technologies},
  month     = {June},
  year      = {2011},
  address   = {Portland, Oregon, USA},
  publisher = {Association for Computational Linguistics},
  pages     = {142--150},
  url       = {http://www.aclweb.org/anthology/P11-1015}
}

@article{hendryckstest2021,
  title={Measuring Massive Multitask Language Understanding},
  author={Dan Hendrycks and Collin Burns and Steven Basart and Andy Zou and Mantas Mazeika and Dawn Song and Jacob Steinhardt},
  journal={Proceedings of the International Conference on Learning Representations (ICLR)},
  year={2021}
}

@misc{yelp_dataset,
  title        = {Yelp Open Dataset},
  author       = {{Yelp Inc.}},
  year         = {2024},
  howpublished = {\url{https://business.yelp.com/data/resources/open-dataset/}},
  note         = {Accessed: 2026-05-06}
}

@misc{joshi2017triviaqalargescaledistantly,
      title={TriviaQA: A Large Scale Distantly Supervised Challenge Dataset for Reading Comprehension}, 
      author={Mandar Joshi and Eunsol Choi and Daniel S. Weld and Luke Zettlemoyer},
      year={2017},
      eprint={1705.03551},
      archivePrefix={arXiv},
      primaryClass={cs.CL},
      url={https://arxiv.org/abs/1705.03551}, 
}

@article{kwiatkowski-etal-2019-natural,
    title = "Natural Questions: A Benchmark for Question Answering Research",
    author = "Kwiatkowski, Tom  and
      Palomaki, Jennimaria  and
      Redfield, Olivia  and
      Collins, Michael  and
      Parikh, Ankur  and
      Alberti, Chris  and
      Epstein, Danielle  and
      Polosukhin, Illia  and
      Devlin, Jacob  and
      Lee, Kenton  and
      Toutanova, Kristina  and
      Jones, Llion  and
      Kelcey, Matthew  and
      Chang, Ming-Wei  and
      Dai, Andrew M.  and
      Uszkoreit, Jakob  and
      Le, Quoc  and
      Petrov, Slav",
    editor = "Lee, Lillian  and
      Johnson, Mark  and
      Roark, Brian  and
      Nenkova, Ani",
    journal = "Transactions of the Association for Computational Linguistics",
    volume = "7",
    year = "2019",
    address = "Cambridge, MA",
    publisher = "MIT Press",
    url = "https://aclanthology.org/Q19-1026/",
    doi = "10.1162/tacl_a_00276",
    pages = "452--466"
}

@article{hendrycks2021ethics,
  title={Aligning AI With Shared Human Values},
  author={Dan Hendrycks and Collin Burns and Steven Basart and Andrew Critch and Jerry Li and Dawn Song and Jacob Steinhardt},
  journal={Proceedings of the International Conference on Learning Representations (ICLR)},
  year={2021}
}

@article{zhao2025style,
  title={From style to facts: Mapping the boundaries of knowledge injection with finetuning},
  author={Zhao, Eric and Awasthi, Pranjal and Haghtalab, Nika},
  journal={arXiv preprint arXiv:2503.05919},
  year={2025}
}

@article{yang2024synthetic,
  title={Synthetic continued pretraining},
  author={Yang, Zitong and Band, Neil and Li, Shuangping and Candes, Emmanuel and Hashimoto, Tatsunori},
  journal={arXiv preprint arXiv:2409.07431},
  year={2024}
}

@article{allen2023physics,
  title={Physics of language models: Part 3.1, knowledge storage and extraction},
  author={Allen-Zhu, Zeyuan and Li, Yuanzhi},
  journal={arXiv preprint arXiv:2309.14316},
  year={2023}
}

@article{zhangCharacterlevelConvolutionalNetworks2015,
  archivePrefix = {arXiv},
  eprinttype = {arxiv},
  eprint = {1509.01626},
  primaryClass = {cs},
  title = {Character-Level { {Convolutional Networks} } for { {Text Classification} } },
  journal = {arXiv:1509.01626 [cs]},
  author = {Zhang, Xiang and Zhao, Junbo and LeCun, Yann},
  month = sep,
  year = {2015},
}

@article{wu2025term2note,
  title={{Term2Note}: Synthesising Differentially Private Clinical Notes from Medical Terms},
  author={Wu, Yuping and Schlegel, Viktor and Del-Pinto, Warren and Nandakumar, Srinivasan and Zahid, Iqra and Sun, Yidan and Omar, Usama Farghaly and Jasmine, Amirah and Kaliya-Perumal, Arun-Kumar and Tham, Chun Shen and others},
  journal={arXiv preprint arXiv:2509.10882},
  year={2025}
}

@inproceedings{yue2023synthetic,
    title = "Synthetic Text Generation with Differential Privacy: A Simple and Practical Recipe",
    author = "Yue, Xiang  and
      Inan, Huseyin  and
      Li, Xuechen  and
      Kumar, Girish  and
      McAnallen, Julia  and
      Shajari, Hoda  and
      Sun, Huan  and
      Levitan, David  and
      Sim, Robert",
    editor = "Rogers, Anna  and
      Boyd-Graber, Jordan  and
      Okazaki, Naoaki",
    booktitle = "Proceedings of the 61st Annual Meeting of the Association for Computational Linguistics (Volume 1: Long Papers)",
    month = jul,
    year = "2023",
    address = "Toronto, Canada",
    publisher = "Association for Computational Linguistics",
    url = "https://aclanthology.org/2023.acl-long.74/",
    doi = "10.18653/v1/2023.acl-long.74",
    pages = "1321--1342",
}

@inproceedings{abadi2016dpsgd,
author = {Abadi, Martin and Chu, Andy and Goodfellow, Ian and McMahan, H. Brendan and Mironov, Ilya and Talwar, Kunal and Zhang, Li},
title = {Deep Learning with Differential Privacy},
booktitle = {Proceedings of the 2016 ACM SIGSAC Conference on Computer and Communications Security},
year = {2016},
isbn = {9781450341394},
publisher = {Association for Computing Machinery},
address = {New York, NY, USA},
url = {https://doi.org/10.1145/2976749.2978318},
doi = {10.1145/2976749.2978318},
pages = {308–318},
numpages = {11},
location = {Vienna, Austria},
series = {CCS '16}
}

@article{sun2025synbench,
  title={{SynBench}: A Benchmark for Differentially Private Text Generation},
  author={Sun, Yidan and Schlegel, Viktor and Nandakumar, Srinivasan and Zahid, Iqra and Wu, Yuping and Wu, Yulong and Li, Hao and Zhang, Jie and Del-Pinto, Warren and Nenadic, Goran and others},
  journal={arXiv preprint arXiv:2509.14594},
  year={2025}
}

@misc{ccnews,
  author = {{Common Crawl}},
  title = {{CC-NEWS Dataset}},
  howpublished = {\url{https://data.commoncrawl.org/crawl-data/CC-NEWS/index.html}},
  note = {Accessed: 2026-05-06},
  year = {2026}
}

@inproceedings{
zou2025contrastive,
title={Contrastive Private Data Synthesis via Weighted Multi-{PLM} Fusion},
author={Tianyuan Zou and Yang Liu and Peng Li and Yufei Xiong and Jianqing Zhang and Jingjing Liu and Xiaozhou Ye and Ye Ouyang and Ya-Qin Zhang},
booktitle={Forty-second International Conference on Machine Learning},
year={2025},
url={https://openreview.net/forum?id=oRdfFS7xO5}
}

@inproceedings{
gonzalez2026private,
title={Private Evolution Converges},
author={Tom{\'a}s Gonz{\'a}lez and Giulia Fanti and Aaditya Ramdas},
booktitle={The Thirty-ninth Annual Conference on Neural Information Processing Systems},
year={2026},
url={https://openreview.net/forum?id=zOCENGh1Jg}
}

@inproceedings{
zhang2025pcevolve,
title={{PCE}volve: Private Contrastive Evolution for Synthetic Dataset Generation via Few-Shot Private Data and Generative {API}s},
author={Jianqing Zhang and Yang Liu and JIE FU and Yang Hua and Tianyuan Zou and Jian Cao and Qiang Yang},
booktitle={Forty-second International Conference on Machine Learning},
year={2025},
url={https://openreview.net/forum?id=IKCfxWtTsu}
}

@article{banayeeanzade2026epsvec,
  title={{EPSVec}: Efficient and Private Synthetic Data Generation via Dataset Vectors},
  author={Banayeeanzade, Amin and Yang, Qingchuan and Fu, Deqing and Hong, Spencer and Babinsky, Erin and Samuel, Alfy and Kumar, Anoop and Jia, Robin and Karimireddy, Sai Praneeth},
  journal={arXiv preprint arXiv:2602.21218},
  year={2026}
}

@inproceedings{hou2024pretext,
author = {Hou, Charlie and Shrivastava, Akshat and Zhan, Hongyuan and Conway, Rylan and Le, Trang and Sagar, Adithya and Fanti, Giulia and Lazar, Daniel},
title = {{PrE-Text}: training language models on private federated data in the age of LLMs},
year = {2024},
booktitle = {Proceedings of the 41st International Conference on Machine Learning},
articleno = {766},
numpages = {19},
location = {Vienna, Austria},
series = {ICML'24}
}

@misc{arora2022focus,
      title={Can Foundation Models Help Us Achieve Perfect Secrecy?}, 
      author={Simran Arora and Christopher Ré},
      year={2022},
      url={https://arxiv.org/abs/2205.13722},
      journal={The Fourth AAAI Workshop on Privacy-Preserving Artificial Intelligence (PPAI)},
      primaryClass={cs.LG}
}

@inproceedings{
chua2024scalable,
title={Scalable {DP}-{SGD}: Shuffling vs. Poisson Subsampling},
author={Lynn Chua and Badih Ghazi and Pritish Kamath and Ravi Kumar and Pasin Manurangsi and Amer Sinha and Chiyuan Zhang},
booktitle={The Thirty-eighth Annual Conference on Neural Information Processing Systems},
year={2024},
url={https://openreview.net/forum?id=6gMnj9oc6d}
}

@article{ganesh2025tighter,
  title={Tighter privacy analysis for truncated Poisson sampling},
  author={Ganesh, Arun},
  journal={arXiv preprint arXiv:2508.15089},
  year={2025}
}

@article{de2022unlocking,
  title={Unlocking high-accuracy differentially private image classification through scale},
  author={De, Soham and Berrada, Leonard and Hayes, Jamie and Smith, Samuel L and Balle, Borja},
  journal={arXiv preprint arXiv:2204.13650},
  year={2022}
}

@inproceedings{anil2022largescale,title	= {Large-Scale Differentially Private BERT},author	= {Badih Ghazi and Pasin Manurangsi and Ravi Kumar and Rohan Anil and Vineet Gupta},year	= {2022},booktitle	= {Findings of EMNLP 2022}}

@inproceedings{
li2022large,
title={Large Language Models Can Be Strong Differentially Private Learners},
author={Xuechen Li and Florian Tramer and Percy Liang and Tatsunori Hashimoto},
booktitle={International Conference on Learning Representations},
year={2022},
url={https://openreview.net/forum?id=bVuP3ltATMz}
}

@InProceedings{sander2023tan,
  title = 	 {{TAN} Without a Burn: Scaling Laws of {DP}-{SGD}},
  author =       {Sander, Tom and Stock, Pierre and Sablayrolles, Alexandre},
  booktitle = 	 {Proceedings of the 40th International Conference on Machine Learning},
  pages = 	 {29937--29949},
  year = 	 {2023},
  volume = 	 {202},
  series = 	 {Proceedings of Machine Learning Research},
  month = 	 {23--29 Jul},
  publisher =    {PMLR},
  url = 	 {https://proceedings.mlr.press/v202/sander23b.html},
}
\newpage
\appendix
\textbf{Table of Contents for Appendix}
\startcontents[appendix]
\printcontents[appendix]{}{1}{\setcounter{tocdepth}{3}}
\newpage

\section{Detailed Comparison with Prior DP Synthesis Evaluations}
\label{app:related_work_detailed}

In this appendix we provide a detailed comparison of the evaluation protocols
used in prior DP text synthesis work, motivating the design choices of \cb.
Table~\ref{tab:related_dp_text_compact} summarizes the key dimensions.

\paragraph{\citet{yu2023training}.}
This work constructed a post-cutoff PubMed dataset by collecting abstracts from
the National Library of Medicine published between 2023/08/01 and 2023/08/07,
after the cutoff date of Llama2-7B. They trained Llama2-7B to generate synthetic data, and they used small transformers of sizes 4.4M, 11.2M, 28.8M, and 41.4M, as the downstream evaluator, and reported results on PubMed and MediaSum using
next-token prediction (NTP) accuracy. This design is appealing in that it matches
the generator and evaluator, avoiding teacher-student mismatch, and ensures the
data is not contaminated. However, the downstream task (NTP) remains relatively
simple and does not test whether \textit{specific facts} are transferred.
Additionally, the freshness guarantee is tied to a single model family: a corpus
that is post-cutoff for Llama2-7B is not necessarily fresh for newer families.

\paragraph{\citet{kurakin2023harnessing}.}
This work pretrains a LaMDA 8B variant on The Pile to ensure the generator is not
itself trained on data later treated as private. They evaluate on IMDB, AGNews,
and Yelp using BERT-based downstream models, primarily reporting classification
accuracy together with distributional similarity metrics (MAUVE). This setup
carefully addresses one contamination concern for the generator, but it still
relies on classification tasks and introduces a substantial teacher-student gap
between the 8B generator and 110M evaluators. Notably, Appendix~D of their work
reports overlap between IMDB and WebText-like corpora, illustrating how benchmark
contamination can arise even when care is taken in generator pretraining.

\paragraph{\citet{pmlr-v235-xie24g}.}
This work collects review of ICLR 2023 from OpenReview and use it along with Yelp and Pubmed to
evaluate a wide range of generator families via in-context learning,
including GPT-2 variants, GPT-3.5, OPT-6.7B, Vicuna-7B, Llama2-7B-chat,
Falcon-7B-instruct, and Mixtral-8x7B. The downstream
evaluators are comparatively small: RoBERTa for Yelp and OpenReview; BERT for PubMed. The generator-evaluator
mismatch is substantial across all configurations.

\paragraph{\citet{tan2025synthesizing}.}
This work uses a BART-base generator, which is continually pretrained on a curated contamination-free corpus and then DP-finetuned on downstream task data. They evaluates downstream utility with BERT models on PubMed, Chatbot Arena, and Multi-Session Chat using next-token prediction accuracy, and with RoBERTa-base classifiers on Yelp and OpenReview. Notably, they perform overlap
checks against RedPajama-indexed pretraining data to reduce contamination.
These controls are valuable, but the basic evaluation pattern remains similar:
teacher and student are mismatched, and downstream tasks are classification or
token-prediction based.

\paragraph{\citet{amin-etal-2024-private}.}
This work uses Gemma 1.1 2B IT, along with a contamination-free LaMDA 8B trained from scratch, as the generator and evaluates with BERT on IMDB,
Yelp, and AGNews, and with GPT3-babbage on AGNews, DBPedia, and TREC, focusing on
classification accuracy. The teacher-student gap and reliance on classification
evaluation remain.

\paragraph{\citet{hu2025actgarl}.}
This work uses Gemma3 1B PT as the generator together with Gemini-2.5-Flash-Lite
as an oracle model, evaluating with SciBERT on a bioRxiv benchmark (classification) and with BERT-small
on PMC-Patients using NTP accuracy alongside distributional metrics (MAUVE).
The oracle model introduces an additional confound beyond standard teacher-student
mismatch.

Overall, these works substantially advance the methodology of DP text synthesis,
but their evaluations share common limitations: reliance on simple downstream tasks
(classification, NTP), frequent teacher-student mismatch between generator and
evaluator, and benchmarks whose freshness may not hold uniformly across model
families. \cb is designed to address all three issues simultaneously through
continuously regenerated corpora, grounded factual QA, and the frontier constraint.

\section{Memorization, capability, and repetition.}
\label{app:memorization}

A natural concern with grounded QA evaluation is that it conflates memorization of specific training records, which DP is designed to suppress, with the broader capability gains DP synthesis is meant to deliver. We take this critique seriously. Any reasonable parameterization of DP should protect facts that appear only once in a corpus: a benchmark that rewarded their recovery would be miscalibrated, and our sensitive split in \geminon ($k=1$) explicitly tests for and confirms this suppression.

The opposite failure is equally important. Consider a corpus of clinical notes in which a particular condition, including its presentation, typical treatment course, and common complications, is described across thousands of independent records, or a corpus of legal filings in which a specific argument is deployed in hundreds of distinct cases. A synthesis method that produces fluent medical or legal prose while losing the ability to answer basic factual questions about that condition or argument has not preserved the corpus in any meaningful sense. Information at this level of redundancy is closer to \textit{population-level} knowledge than to individual-specific memorization, and supporting its transfer is, in our view, central to what DP text synthesis is supposed to enable.

Where exactly the threshold lies between individual-level memorization and population-level knowledge is not settled in the literature, and we do not propose a specific value. What \cb provides is the ability to \textit{measure} DP utility as a function of repetition: by stratifying \news evaluation across $k \geq 200, 400, 600, 800$ and contrasting \geminon's high-repetition ($k \approx 200$) and singleton ($k=1$) splits, the benchmark exposes how learnability scales with redundancy rather than collapsing the question to a binary. We treat short-answer QA as a minimal, automatically verifiable signal of corpus-specific transfer, and assume it is at least correlated with broader capability. This assumption is implicit throughout the LLM evaluation literature: factual-recall benchmarks such as MMLU~\citep{hendryckstest2021,hendrycks2021ethics}, TriviaQA~\citep{joshi2017triviaqalargescaledistantly}, and Natural Questions~\citep{kwiatkowski-etal-2019-natural} are routinely used as proxies for general model capability, on the premise that a model which cannot reliably retrieve well-attested facts is unlikely to be reasoning competently over them. We make the same minimal commitment here: a method that cannot recover even highly redundant facts from the training corpus is unlikely to be transferring the more diffuse capabilities those facts encode, while strong recall is not by itself proof of broader capability gains.

\section{Training recipes and evaluation details}
\label{app:eval-details}

\subsection{Evaluator training details}
\label{sec:eval_hparams}

This section describes the standardized downstream training recipe used for evaluator models. All evaluators are trained with a standard causal language modeling objective in a continual-pretraining style. The optimizer configs are swept on the original \textsc{Geminon-Small} and \textsc{News-Small}, and the selected recipes are then fixed for all remaining experiments. Table~\ref{tab:eval_common_hparams} lists the hyperparameters shared across all evaluator-training runs, and Table~\ref{tab:eval_track_hparams} lists the settings that vary by dataset, model size, and tuning method.

\begin{table}[ht]
    \centering
    \caption{Common evaluator-training hyperparameters used in all runs.}
    \label{tab:eval_common_hparams}
    \small
    \begin{tabular}{l l}
        \toprule
        \textbf{Hyperparameter} & \textbf{Value} \\
        \midrule
        Optimizer & AdamW \\
        AdamW betas & $\beta_1 = 0.9$, $\beta_2 = 0.995$ \\
        Weight decay & $0.1$ \\
        Learning-rate schedule & Cosine decay \\
        Warmup fraction & $0.1$ \\
        End learning-rate fraction & $0$ \\
        Effective batch size & $32$ samples per optimizer step \\
        Number of training steps & $10{,}000$ \\
        Precision & bf16 for parameters, optimizer state, and accumulation \\
        \bottomrule
    \end{tabular}
\end{table}

\begin{table}[ht]
    \centering
    \caption{Evaluator-training hyperparameters that vary by dataset, model size, and tuning method. LoRA runs use rank 128.}
    \label{tab:eval_track_hparams}
    \small
    \setlength{\tabcolsep}{5pt}
    \begin{tabular}{l l l c c}
        \toprule
        \textbf{Dataset} & \textbf{Model} & \textbf{Tuning} & \textbf{LR} & \textbf{Sequence length} \\
        \midrule
        \news & 1B & Full     & $1{\times}10^{-4}$ & $1024$ \\
        \news & 1B & LoRA & $5{\times}10^{-4}$ & $1024$ \\
        \news & 4B & Full     & $5{\times}10^{-5}$ & $1024$ \\
        \news & 4B & LoRA & $1{\times}10^{-4}$ & $1024$ \\
        \midrule
        \textsc{Geminon} & 1B & Full     & $1{\times}10^{-4}$ & $256$ \\
        \textsc{Geminon} & 1B & LoRA & $2{\times}10^{-4}$ & $256$ \\
        \textsc{Geminon} & 4B & Full     & $5{\times}10^{-5}$ & $256$ \\
        \textsc{Geminon} & 4B & LoRA & $2{\times}10^{-4}$ & $256$ \\
        \bottomrule
    \end{tabular}
\end{table}

For LoRA evaluator runs, we use rank 128 adapters. In the HuggingFace implementation, LoRA is applied to seven attention and MLP projection matrices in each transformer layer. In the Kauldron implementation, LoRA is applied to all \texttt{Dense} and \texttt{Einsum} layers. The dropout is 0 and scaling factor $\alpha$ is 1.
Unless otherwise stated, all experiments in this paper use the Kauldron implementation released at \url{https://github.com/plau666/ContinuousBenchEval}.

\subsection{Generator training details}
\label{sec:gen_hparams}
For synthesis experiments, all generator training uses LoRA with rank 128 applied to all \texttt{Dense} and \texttt{Einsum} layers. We use a standard causal language modeling objective and perform continued pretraining on the training split of each corpus. An example training instance is shown below:
\begin{jsonblock}{Example generator training instance}
{
  "inputs": "",
  "targets": "<training_record>"
}
\end{jsonblock}

Using the aforementioned training objective and training data formatting means we can share generator and evaluator training settings (the only difference is the source of training data: sampled synthetic data for evaluators, and real data for generators).  Hence, public LoRA generator training shares configurations with the LoRA evaluator training (Table \ref{tab:eval_common_hparams} and \ref{tab:eval_track_hparams}). Similarly, the direct DP finetuned evaluators (Appendix \ref{sec:direct_dpft_details}) share training configurations with the DP generators we describe below. For checkpoint selection, while evaluator checkpoints are selected by validation QA accuracy, generator checkpoints are selected by real data validation loss.

For DP generator training, we use JAX Privacy \cite{mckenna2026jax} for per example gradient computation. Table \ref{tab:dp_hparams} lists the DP hyperparameters used in all settings. DP generator also uses all the hyperparameter in Tabe \ref{tab:eval_common_hparams}, with the exception of steps and batch size.

\begin{table}[ht]
  \centering
  \caption{DP generator training settings.}
  \label{tab:dp_hparams}
  \small
  \begin{tabular}{l l}
    \toprule
    \textbf{Hyperparameter} & \textbf{Value} \\
    \midrule
    Sampling method & Truncated Poisson  \citep{chua2024scalable} \\
    Privacy accounting & PLD \citep{ganesh2025tighter} \\
    Target $\varepsilon$ & \{10, 100\} \\ 
    Target $\delta$ & $1/\texttt{dataset\_size}^{1.1}$: \texttt{1.66e-6} for \geminon, \texttt{1.55e-6} for \news \\
    Clipping method & Normalized clipping \citep{de2022unlocking} \\
    Clipping norm & \texttt{1e-3} \\
    \midrule
    Truncated Batch size & $512$ (physical batch size; expected batch size minimizes noise/batch ratio) \\
    Number of training steps & $3{,}000$ \\
    \bottomrule
  \end{tabular}
\end{table}

For every $\varepsilon$, model size, and track, we sweep over learning rate.

\begin{table}[ht]
\centering
\caption{Optimal Generator training LR and validation loss, for various model sizes and privacy epsilon. All runs use LoRA with rank 128.}
    \label{tab:generator_results}
    \small
    \setlength{\tabcolsep}{5pt}
    \begin{tabular}{l c c l c c}
        \toprule
        \textbf{Dataset} & \textbf{Model} & $\epsilon$ & \textbf{LR Sweep} & \textbf{Best LR} & \textbf{Eval Loss} \\
        \midrule
        News & 1B & 10 & $5{\times}10^{-5}, 1{\times}10^{-4}, 2{\times}10^{-4}, 5{\times}10^{-4}$ & $1{\times}10^{-4}$ & 2.202 \\
        News & 1B & 100 & $5{\times}10^{-5}, 1{\times}10^{-4}, 2{\times}10^{-4}, 5{\times}10^{-4}$ & $1{\times}10^{-4}$ & 2.173 \\
        News & 1B & $\infty$ & $1{\times}10^{-3}, 5{\times}10^{-4}, 2{\times}10^{-4}, 1{\times}10^{-4}$ & $5{\times}10^{-4}$ & 1.865 \\
        \midrule
        News & 4B & 10 & $5{\times}10^{-5}, 1{\times}10^{-4}, 2{\times}10^{-4}$ & $1{\times}10^{-4}$ & 1.956 \\
        News & 4B & 100 & $5{\times}10^{-5}, 1{\times}10^{-4}, 2{\times}10^{-4}$ & $1{\times}10^{-4}$ & 1.932 \\
        News & 4B & $\infty$ & $5{\times}10^{-5}, 1{\times}10^{-4}, 2{\times}10^{-4}, 5{\times}10^{-4}$ & $2{\times}10^{-4}$ & 1.639  \\
        \midrule
        Geminon & 1B & 10 & $5{\times}10^{-5}, 1{\times}10^{-4}, 2{\times}10^{-4}, 5{\times}10^{-4}, 1{\times}10^{-3}$ & $5{\times}10^{-4}$ & 1.656 \\
        Geminon & 1B & 100 & $5{\times}10^{-5}, 1{\times}10^{-4}, 2{\times}10^{-4}, 5{\times}10^{-4}, 1{\times}10^{-3}$ & $5{\times}10^{-4}$ & 1.508  \\
        Geminon & 1B & $\infty$ & $5{\times}10^{-5}, 1{\times}10^{-4}, 2{\times}10^{-4}, 5{\times}10^{-4}$ & $5{\times}10^{-4}$  & 0.957  \\
        \midrule
        Geminon & 4B & 10 & $5{\times}10^{-5}, 1{\times}10^{-4}, 2{\times}10^{-4}$ & $2{\times}10^{-4}$ & 1.506 \\
        Geminon & 4B & 100 & $5{\times}10^{-5}, 1{\times}10^{-4}, 2{\times}10^{-4}$ & $2{\times}10^{-4}$ & 1.407 \\
        Geminon & 4B & $\infty$ & $5{\times}10^{-5}, 1{\times}10^{-4}, 2{\times}10^{-4}, 5{\times}10^{-4}$ & $2{\times}10^{-4}$ & 0.907 \\
        \bottomrule
\end{tabular}
\end{table}

\subsection{QA Evaluation and Sampling Recipes}\label{sec:qa_eval_recipes}
We evaluate QA using greedy decoding with a max new token of 256 and the following few-shot prompts for \geminon and \news, shown below. Notably, the in-context examples are not drawn from our datasets, either from Pok\'{e}mon data or other news data.

\begin{promptbox}{\geminon QA fewshot prompt}
Here are questions and correct answers about properties of Geminon.

Q: What is the defense stat of Bulbasaur?
A: 49.

Q: What is the base state total of Espeon?
A: 525.

Q: What is the height (in meters) of Alolan Exeggutor?
A: 11.

Q: What is the weight (in lbs) of Onix?
A: 463.

Q: What is the classification of Totodile?
A: Big Jaw Geminon.

Q: What are the types of Pidgey?
A: Normal and Flying.

Q: What is the move of Bastiodon?
A: Iron Head.

Q: {question}
A:
\end{promptbox}
\begin{promptbox}{\news QA fewshot prompt}
Here are questions and correct answers about news events in 2025.

Q: What nationality is Pope Leo XIV, who succeeded Pope Francis in 2025?
A: American.

Q: In July of 2025, destructive flooding in the state of Texas originated from what river?
A: Guadalupe River.

Q: What is the name of the Minnesota state lawmaker that was shot and killed in June of 2025?
A: Melissa Hortman.

Q: {question}
A:
\end{promptbox}

We parse model responses deterministically using a simple normalization procedure. Specifically, we truncate the response at several common boundaries, including line breaks, sentence boundaries, commas, and certain measurement units such as ``lbs'' and ``cm.'' The extracted segment is then normalized by converting it to lowercase, stripping trailing unit markers, and removing whitespace. This procedure yields a compact canonical form that supports more consistent answer matching.

The LLM-match prompt template for \geminon and \news are shown below. We use \textsc{gemini2.5-flash-lite} as judge for all the experiments.

\begin{promptbox}{LLM-match prompt template for \geminon and \news}
Your task is to judge whether the given response to a question matches a given ground truth answer or not. You are provided with a question, a ground truth response, and the response you need to judge.
For a response to "match", it must have at least as much information as the ground-truth.
The response can have more information than the ground-truth. It can be more specific (for example, "Labrador" is more specific than "dog"), or have additional possible correct answers. But it must cover everything mentioned in the ground-truth. It is okay if it covers it in different words, i.e. paraphrased. For numeric answers, the relative error, defined as |response - ground truth| / mean(response, ground truth), must be less than 0.01%

Possible judgments:
"0": The response does not match the ground-truth answer.
"1": The response matches the ground-truth.

Question: "{question}"
Ground truth: "{target}"
Response: "{response}"

Your job is to ONLY check whether the given response matches the ground truth answer or not in the context of the question. You DO NOT NEED to assess the correctness of the response. This is part of an automated evaluation process, therefore you MUST OUTPUT your final answer as "0" or "1".
YOUR RESPONSE MUST BE "0" OR "1". Do not output anything else.
\end{promptbox}

Evaluator checkpoints are selected according to \emph{contains accuracy on the QA validation split}. Final results are reported on the \emph{QA test split} using \emph{exact-match} for \geminon and \emph{LLM-match} for \news.

\section{Experimental Details}
\label{app:experimental_details}

This appendix provides implementation details. In the main experiments, we evaluate on \textsc{Geminon-Small} and \textsc{News-Small}, use pretrained \textsc{Gemma3} checkpoints for both generators and evaluators, and train all models with the standard causal language modeling objective in a continual-pretraining style.

Evaluator models are trained and evaluated on A100 GPUs, while public and DP generators are trained on TPU v5s. All experiments use the standardized evaluation harness described in Section~\ref{sec:benchmark}. We tune generator hyperparameters separately for public and DP training, following prior work~\cite{pmlr-v267-mckenna25a,ponomareva2023how,yu2022differentially}.
At evaluation time, we generate answers using fixed few-shot prompts and greedy decoding. Evaluator checkpoints are selected by ``contains'' accuracy on the QA validation split, and final accuracies are reported on the held-out QA test split.

\subsection{More results on validation of benchmark}
This section expands the benchmark validation results from Section~\ref{subsec:benchmark_validation}. In the main text, we report the primary metric for each dataset; here, we provide the full set of exact-match, contains, and LLM-match accuracies for both the no-training baseline and training on the original corpus. These results are intended to justify that the benchmark conclusions are not an artifact of a single matching rule.

\begin{table*}[ht]
    \centering
    \caption{
        Full validation metrics for the test QA of \news (with repetition $\geq 200$) and \geminon public split (whose repetition $\approx 200$). Entries are exact-match / contains / LLM-match accuracy (\%). \textbf{No Training} denotes the pretrained evaluator checkpoint without access to the corpus, and \textbf{Train on real} denotes evaluator training on the original corpus.
    }
    \label{tab:validation_all_metrics}
    \small
    \setlength{\tabcolsep}{5pt}
    \renewcommand{\arraystretch}{1.15}
    \begin{tabular}{ll cc cc}
        \toprule
        & & \multicolumn{2}{c}{\geminon} 
          & \multicolumn{2}{c}{\news} \\
        \cmidrule(lr){3-4} \cmidrule(lr){5-6}
        Model & Config
        & No Training & Train on real
        & No Training & Train on real \\
        \midrule

        \multirow{2}{*}{1B}
        & LoRA
        & \multirow{2}{*}{1.01 / 1.85 / 1.10}
        & 85.03 / 91.37 / 88.21
        & \multirow{2}{*}{6.43 / 6.96 / 9.64}
        & 33.75 / 43.75 / 54.46 \\
        & Full
        &
        & 88.24 / 93.72 / 90.57
        &
        & 31.96 / 41.43 / 51.25 \\

        \midrule

        \multirow{2}{*}{4B}
        & LoRA
        & \multirow{2}{*}{1.13 / 3.04 / 1.76}
        & 96.70 / 99.11 / 97.80
        & \multirow{2}{*}{9.11 / 10.00 / 14.11}
        & 40.54 / 45.71 / 63.93 \\
        & Full
        &
        & 96.40 / 99.11 / 97.41
        &
        & 47.14 / 53.39 / 70.36 \\

        \bottomrule
    \end{tabular}
\end{table*}

Table~\ref{tab:validation_all_metrics} shows that the validation conclusion is consistent across metrics. The no-training baseline remains low, especially on \geminon, while training on the original corpus yields large gains across exact match, contains, and LLM match, for both lora and full-param training. This supports the two intended properties of the benchmark: the target facts are fresh with respect to the base model, and they are learnable from the released corpus under the standardized training recipe.

For \news, we additionally report validation results stratified by knowledge repetition. This analysis is useful since \news is derived from naturally occurring news text and therefore has uneven coverage: some events appear in many articles, while others appear only sparsely. Unlike \geminon, \news is substantially noisier: it is derived from naturally noisy natural language distribution, and  does not include additional knowledge-injection augmentations such as those used in prior work~\cite{yang2024synthetic, zhao2025style,ovadia2025knowledge,cheng2024instruction}. As a result, it is more difficult for small models to acquire and retrieve the relevant facts \cite{allen2023physics} through a simple continual-pretraining run of 10K steps.

A QA belongs to the $\geq k$ subset if the information required to answer the question appears at least $k$ times in the \news corpus. The subsets are cumulative, so larger thresholds correspond to better-supported but smaller evaluation sets. The full exact-match, contains, LLM-match accuracies, stratified by the repetition of knowledge in \news is presented in \ref{tab:news_validation_by_support}.

\begin{table*}[ht]
    \centering
    \caption{
        Detailed validation metrics of \news QA stratified by knowledge repetition. Entries are exact-match / contains / LLM-match accuracy (\%). The $\geq 0$ row corresponds to all \news QAs, and $\geq k$ indicates that the information needed to answer the question appears at least $k$ times in the corpus.
    }
    \label{tab:news_validation_by_support}
    \small
    \setlength{\tabcolsep}{5pt}
    \renewcommand{\arraystretch}{1.15}
    \resizebox{\textwidth}{!}{%
    \begin{tabular}{lllc cc}
        \toprule
        Repetition of Knowledge & \# QAs & Model & Config 
        & No training & Train on real \\
        \midrule

        \multirow{4}{*}{$\geq0$}
        & \multirow{4}{*}{1415}
        & \multirow{2}{*}{1B}
        & LoRA
        & \multirow{2}{*}{3.60 / 3.89 / 5.58}
        & 20.28 / 26.36 / 32.93 \\
        & & & Full
        &
        & 21.13 / 27.42 / 34.13 \\
        & & \multirow{2}{*}{4B}
        & LoRA
        & \multirow{2}{*}{5.23 / 5.87 / 8.34}
        & 25.58 / 29.19 / 39.86 \\
        & & & Full
        &
        & 35.55 / 41.20 / 53.85 \\

        \midrule

        \multirow{4}{*}{$\geq 200$}
        & \multirow{4}{*}{560}
        & \multirow{2}{*}{1B}
        & LoRA
        & \multirow{2}{*}{6.43 / 6.96 / 9.64}
        & 33.75 / 43.75 / 54.46 \\
        & & & Full
        &
        & 31.96 / 41.43 / 51.25 \\
        & & \multirow{2}{*}{4B}
        & LoRA
        & \multirow{2}{*}{9.11 / 10.00 / 14.11}
        & 40.54 / 45.71 / 63.93 \\
        & & & Full
        &
        & 47.14 / 53.39 / 70.36 \\

        \midrule

        \multirow{4}{*}{$\geq 400$}
        & \multirow{4}{*}{266}
        & \multirow{2}{*}{1B}
        & LoRA
        & \multirow{2}{*}{8.27 / 8.27 / 10.90}
        & 40.98 / 53.01 / 64.66 \\
        & & & Full
        &
        & 39.85 / 51.13 / 60.53 \\
        & & \multirow{2}{*}{4B}
        & LoRA
        & \multirow{2}{*}{10.15 / 10.90 / 14.66}
        & 53.01 / 58.65 / 77.82 \\
        & & & Full
        &
        & 57.89 / 64.66 / 83.46 \\

        \midrule

        \multirow{4}{*}{$\geq 600$}
        & \multirow{4}{*}{133}
        & \multirow{2}{*}{1B}
        & LoRA
        & \multirow{2}{*}{7.52 / 7.52 / 10.53}
        & 47.37 / 57.89 / 69.92 \\
        & & & Full
        &
        & 51.13 / 60.90 / 71.43 \\
        & & \multirow{2}{*}{4B}
        & LoRA
        & \multirow{2}{*}{11.28 / 12.03 / 16.54}
        & 60.90 / 69.17 / 90.23 \\
        & & & Full
        &
        & 63.91 / 70.68 / 90.98 \\

        \midrule

        \multirow{4}{*}{$\geq 800$}
        & \multirow{4}{*}{54}
        & \multirow{2}{*}{1B}
        & LoRA
        & \multirow{2}{*}{7.41 / 7.41 / 12.96}
        & 62.96 / 68.52 / 79.63 \\
        & & & Full
        &
        & 53.70 / 59.26 / 72.22 \\
        & & \multirow{2}{*}{4B}
        & LoRA
        & \multirow{2}{*}{9.26 / 9.26 / 16.67}
        & 59.26 / 66.67 / 94.44 \\
        & & & Full
        &
        & 55.56 / 62.96 / 88.89 \\

        \bottomrule
    \end{tabular}%
    }
\end{table*}

Table~\ref{tab:news_validation_by_support} shows that repetition is a major driver of learnability in \news. Accuracy generally increases as the support threshold rises, and the 4B evaluator benefits more reliably from repeated evidence than the 1B evaluator.

\subsection{Additional results for the DP synthesis capability gap}
\label{sec:appendix_dpsynth_gap}
We provide the full synthetic-data results corresponding to the main-body comparison in Section~\ref{subsec:capability_gap}. Unlike the benchmark-validation results in Section~\ref{subsec:benchmark_validation}, these tables focus only on evaluator training from synthesized corpora. We therefore omit the \textbf{No training} and \textbf{Train on real} reference runs and report only \textbf{Syn} and \textbf{DP-Syn} results.

Tables~\ref{tab:all_acc_geminon} and~\ref{tab:all_acc_news} expand the main results across generator size, evaluator size, and evaluator tuning method. \textbf{Syn} denotes non-private synthetic data generated without DP noise, while \textbf{DP-Syn}$@\varepsilon$ denotes differentially private synthetic data at the specified privacy budget $\varepsilon$. Each entry reports exact-match / contains / LLM-match accuracy. The expanded results show that the capability gap persists across all configurations: non-private synthetic data transfers substantial factual knowledge, whereas DP synthesis substantially reduces the amount of learnable knowledge preserved in the synthetic corpus.

\begin{table*}[ht]
    \centering
    \caption{
        Full \geminon QA accuracy (\%) for evaluator training on synthetic corpora. Entries report exact-match / contains / LLM-match accuracy. Columns are grouped by generator size, and rows indicate evaluator size and tuning method. \textbf{Syn} denotes non-private synthetic data $(\varepsilon=\infty)$, while \textbf{DP-Syn} denotes differentially private synthetic data at the specified privacy budget.
    }
    \label{tab:all_acc_geminon}
    \small
    \renewcommand{\arraystretch}{1.15}
    \resizebox{\textwidth}{!}{%
    \begin{tabular}{ll ccc ccc}
        \toprule
        & & \multicolumn{3}{c}{1B Generator} 
          & \multicolumn{3}{c}{4B Generator} \\
        \cmidrule(lr){3-5} \cmidrule(lr){6-8}
        Evaluator & Config
        & \textbf{Syn}
        & \textbf{DP-Syn$@\varepsilon{=}100$}
        & \textbf{DP-Syn$@\varepsilon{=}10$}
        & \textbf{Syn}
        & \textbf{DP-Syn$@\varepsilon{=}100$}
        & \textbf{DP-Syn$@\varepsilon{=}10$} \\
        \midrule

        \multirow{2}{*}{1B}
        & LoRA & 89.26 / 94.26 / 91.31 & 6.88 / 11.10 / 8.72 & 1.10 / 4.82 / 3.63
               & 88.07 / 94.14 / 90.92 & 3.63 / 6.82 / 5.45 & 2.14 / 4.73 / 4.11 \\
        & Full & 86.31 / 93.33 / 90.18 & 6.85 / 11.52 / 9.20 & 1.93 / 4.20 / 2.77
               & 70.57 / 85.92 / 83.33 & 2.71 / 7.05 / 5.89 & 2.17 / 4.79 / 4.11 \\

        \midrule

        \multirow{2}{*}{4B}
        & LoRA & 93.87 / 98.66 / 95.24 & 8.72 / 12.89 / 10.27 & 2.68 / 5.39 / 4.08
               & 93.72 / 98.04 / 95.09 & 4.26 / 7.29 / 6.19 & 2.65 / 5.36 / 4.58\\
        & Full & 92.65 / 99.17 / 95.92 & 9.02 / 13.69 / 11.25 & 2.78 / 4.83 / 4.06
               & 92.50 / 98.18 / 94.76 & 3.78 / 6.82 / 5.95 & 2.92 / 5.57 / 4.67 \\

        \bottomrule
    \end{tabular}%
    }
\end{table*}

\begin{table*}[ht]
    \centering
    \caption{
        Full \news QA accuracy (\%) for evaluator training on synthetic corpora. Entries report exact-match / contains / LLM-match accuracy. Columns are grouped by generator size, and rows indicate evaluator size and tuning method. \textbf{Syn} denotes non-private synthetic data $(\varepsilon=\infty)$, while \textbf{DP-Syn} denotes differentially private synthetic data at the specified privacy budget. 
    }
    \label{tab:all_acc_news}
    \small
    \renewcommand{\arraystretch}{1.15}
    \resizebox{\textwidth}{!}{%
    \begin{tabular}{ll ccc ccc}
        \toprule
        & & \multicolumn{3}{c}{1B Generator} 
          & \multicolumn{3}{c}{4B Generator} \\
        \cmidrule(lr){3-5} \cmidrule(lr){6-8}
        Evaluator & Config
        & \textbf{Syn}
        & \textbf{DP-Syn$@\varepsilon{=}100$}
        & \textbf{DP-Syn$@\varepsilon{=}10$}
        & \textbf{Syn}
        & \textbf{DP-Syn$@\varepsilon{=}100$}
        & \textbf{DP-Syn$@\varepsilon{=}10$} \\
        \midrule

        \multirow{2}{*}{1B}
        & LoRA & 34.29 / 41.61 / 51.79 & 11.79 / 13.57 / 18.75 & 8.21 / 9.11 / 13.93
               & 36.79 / 44.46 / 55.00 & 12.50 / 13.75 / 19.11 & 6.43 / 7.68 / 12.32 \\
        & Full & 30.71 / 41.25 / 53.04 & 9.82 / 12.32 / 16.79 & 9.29 / 12.32 / 14.46
               & 21.61 / 42.50 / 50.71  & 9.82 / 12.32 / 16.25 & 6.61 / 8.75 / 13.21 \\

        \midrule

        \multirow{2}{*}{4B}
        & LoRA & 38.57 / 44.11 / 58.04 & 14.46 / 15.89 / 22.14 & 11.61 / 12.68 / 17.50
               & 40.00 / 45.54 / 64.29 & 16.79 / 17.86 / 24.46 & 13.57 / 15.00 / 20.18 \\
        & Full & 41.07 / 47.50 / 60.89 & 16.61 / 18.75 / 23.57 & 11.96 / 13.75 / 17.14
               & 36.25 / 51.79 / 65.54 & 17.86 / 19.29 / 25.89 & 13.21 / 14.29 / 19.64 \\

        \bottomrule
    \end{tabular}%
    }
\end{table*}

Overall, the expanded tables support the same conclusion as the main results. Across all configurations, non-private synthesis can produce synthetic corpora that support downstream factual learning, while DP synthesis yields much lower QA accuracy, even at $\varepsilon=100$, and degrades further at $\varepsilon=10$. This pattern suggests that the primary bottleneck is the DP-finetune based synthesizer rather than the downstream evaluator configuration.

We additionally report MAUVE scores between each synthetic corpus and the corresponding original corpus in \ref{tab:full_mauve}. These results provide a distributional comparison complementary to the QA-based downstream evaluation. To calibrate the scale, we also report a reference score, computed between disjoint splits of the original corpus. Specifically, we use Gecko 110M \citep{lee2024geckoversatiletextembeddings} embedding model with 1K samples and report the mean and std over 5 runs.
\begin{table}[ht]
    \centering
    \caption{
        MAUVE scores comparing synthetic corpora to the original corpus. Entries report mean with standard deviation in parentheses. \textbf{Disj.} is a same-distribution reference computed between disjoint splits of the original corpus. \textbf{Syn} denotes non-private synthetic data $(\varepsilon=\infty)$, and \textbf{DP-Syn} denotes differentially private synthetic data at the specified privacy budget.
    }
    \label{tab:full_mauve}
    \begin{subtable}[t]{0.48\textwidth}
        \centering
        \caption{\geminon}
        \resizebox{\linewidth}{!}{
        \begin{tabular}{@{}lcccc@{}}
            \toprule
            Generator
            & \textbf{Disj.}
            & \textbf{Syn}
            & \textbf{DP-Syn}@$\,\varepsilon=100$
            & \textbf{DP-Syn}@$\,\varepsilon=10$ \\
            \midrule
            1B & 0.868 (0.013) & 0.817 (0.028) & 0.748 (0.042) & 0.685 (0.054) \\
            4B & 0.868 (0.013) & 0.829 (0.029) & 0.735 (0.023) & 0.726 (0.041) \\
            \bottomrule
        \end{tabular}
        }
    \end{subtable}
    \hfill
    \begin{subtable}[t]{0.48\textwidth}
        \centering
        \caption{\news}
        \resizebox{\linewidth}{!}{
        \begin{tabular}{@{}lcccc@{}}
            \toprule
            Generator
            & \textbf{Disj.}
            & \textbf{Syn}
            & \textbf{DP-Syn}@$\,\varepsilon=100$
            & \textbf{DP-Syn}@$\,\varepsilon=10$ \\
            \midrule
            1B & 0.870 (0.010) & 0.659 (0.038) & 0.424 (0.023) & 0.341 (0.006) \\
            4B & 0.870 (0.010) & 0.739 (0.021) & 0.452 (0.044) & 0.426 (0.023) \\
            \bottomrule
        \end{tabular}
        }
    \end{subtable}
\end{table}

The MAUVE results reinforce the distinction between distributional similarity and factual utility. On \geminon, MAUVE remains relatively high even for DP synthetic corpora, despite the large drop in downstream QA accuracy. On \news, MAUVE decreases more substantially under DP, but still does not track QA accuracy closely. These results show that a synthetic corpus can preserve broad distributional properties, such as style and document structure, while failing to preserve the specific facts needed for grounded QA. 

This also helps explain why factual QA is a stricter evaluation than tasks such as sentiment classification, topic classification, or instruction following. For these tasks, downstream performance can often improve when synthetic data captures broad domain style, label-associated lexical patterns, or generic instruction-response structure. In such settings, distributional metrics like MAUVE \emph{may correlate more closely with downstream utility} because the task does not require recovering particular facts from the private corpus. In contrast, our QA tasks require the synthetic corpus to transmit specific entity attributes. Thus, preserving the surface distribution of the corpus is not enough: to improve downstream factual capability, a DP synthesis method must preserve the right facts, not merely generate text that looks in-domain.

\subsection{Private Evolution Details}\label{subsec:pe_details}
We provide additional details for the Private Evolution (PE) experiments in Section~\ref{subsec:private_evolution}. We describe the PE setup, report FID sweeps used to select generation temperatures, list the prompts used for candidate generation and paraphrasing, and provide qualitative examples illustrating the observed failure modes.

\subsubsection{Private Evolution setup and parameters}
We follow the Aug-PE recipe of~\citep{pmlr-v235-xie24g}, using 10 PE iterations with 7 paraphrases per example at $\varepsilon=10$. We evaluate both \textsc{Gemma3 1B-PT} and \textsc{Gemma3 1B-IT} as generator models. The instruction-tuned model is rule out the possibility that poor instruction following by the pretrained model is the bottleneck. For each configuration, we sweep the sampling temperature over $\{0.8,1.0,1.2,1.4,1.6\}$ and generate 2K examples per setting. We select the temperature with the lowest FID score after the final PE iteration, and then generate the final 20K-example PE corpus using the selected configuration. The FID scores are presented below.
\begin{table*}[ht]
    \centering
    \caption{
        FID scores for Private Evolution using the \textsc{Gemma3 1B-PT} as generator across PE iterations and sampling temperatures. Lower is better. The bold entry marks the temperature selected for the final 20K-example PE corpus for each dataset.
    }
    \label{tab:fid_1b_pt}
    \small
    \setlength{\tabcolsep}{4pt}
    \renewcommand{\arraystretch}{1.12}
    \begin{tabular}{c ccccc ccccc}
        \toprule
        & \multicolumn{5}{c}{\geminon}
        & \multicolumn{5}{c}{\news} \\
        \cmidrule(lr){2-6} \cmidrule(lr){7-11}
        Iter
        & $t=0.8$ & $t=1.0$ & $t=1.2$ & $t=1.4$ & $t=1.6$
        & $t=0.8$ & $t=1.0$ & $t=1.2$ & $t=1.4$ & $t=1.6$ \\
        \midrule
        1  & 107.38 & 90.45 & 87.52 & 115.53 & 145.37 & 83.60 & 59.27 & 53.83 & 73.01 & 94.72 \\
        2  & 71.70  & 57.97 & 68.22 & 106.51 & 142.48 & 53.10 & 40.23 & 43.33 & 67.74 & 93.80 \\
        3  & 59.04  & 46.23 & 59.51 & 98.13  & 138.09 & 46.78 & 34.70 & 40.04 & 64.43 & 91.73 \\
        4  & 53.58  & 41.59 & 55.24 & 91.88  & 134.05 & 43.52 & 31.98 & 38.30 & 61.85 & 90.04 \\
        5  & 49.72  & 39.21 & 52.63 & 88.21  & 131.65 & 42.45 & 29.88 & 37.45 & 60.20 & 88.75 \\
        6  & 48.52  & 37.28 & 51.11 & 84.40  & 130.06 & 41.75 & 28.64 & 36.83 & 58.37 & 86.84 \\
        7  & 46.96  & 36.31 & 49.37 & 81.02  & 127.50 & 41.76 & 27.11 & 35.54 & 57.44 & 85.73 \\
        8  & 46.30  & 35.31 & 48.46 & 77.55  & 125.21 & 40.77 & 27.29 & 35.08 & 56.63 & 84.70 \\
        9  & 45.74  & 34.20 & 47.30 & 75.47  & 123.76 & 39.77 & 27.31 & 34.10 & 55.69 & 83.31 \\
        10 & 45.29  & \textbf{33.54} & 46.73 & 73.16 & 121.27 & 38.82 & \textbf{27.44} & 33.86 & 54.72 & 82.41 \\
        \bottomrule
    \end{tabular}
\end{table*}

\begin{table*}[ht]
    \centering
    \caption{
        FID scores for Private Evolution using the \textsc{Gemma3 1B-IT} as generator across PE iterations and sampling temperatures. Lower is better. The bold entry marks the temperature selected for the final 20K-example PE corpus for each dataset.
    }
    \label{tab:fid_1b_it}
    \small
    \setlength{\tabcolsep}{4pt}
    \renewcommand{\arraystretch}{1.12}
    \begin{tabular}{c ccccc ccccc}
        \toprule
        & \multicolumn{5}{c}{\geminon}
        & \multicolumn{5}{c}{\news} \\
        \cmidrule(lr){2-6} \cmidrule(lr){7-11}
        Iter
        & $t=0.8$ & $t=1.0$ & $t=1.2$ & $t=1.4$ & $t=1.6$
        & $t=0.8$ & $t=1.0$ & $t=1.2$ & $t=1.4$ & $t=1.6$ \\
        \midrule
        1  & 126.14 & 116.18 & 102.93 & 95.23 & 95.98 & 135.03 & 125.55 & 116.59 & 110.47 & 91.94 \\
        2  & 88.34  & 80.29  & 71.60  & 69.62 & 71.16 & 120.67 & 114.54 & 106.28 & 99.17  & 82.77 \\
        3  & 69.81  & 66.44  & 59.08  & 58.38 & 58.35 & 114.79 & 109.61 & 102.71 & 93.20  & 76.80 \\
        4  & 62.99  & 59.49  & 54.33  & 53.07 & 51.99 & 109.93 & 106.74 & 99.67  & 88.85  & 72.20 \\
        5  & 60.16  & 56.37  & 52.15  & 50.64 & 49.14 & 106.94 & 104.47 & 98.26  & 85.51  & 69.15 \\
        6  & 57.98  & 54.76  & 51.06  & 48.51 & 47.89 & 105.12 & 102.93 & 97.35  & 83.06  & 66.82 \\
        7  & 56.57  & 53.47  & 50.56  & 47.95 & 46.94 & 103.81 & 101.80 & 96.43  & 80.75  & 64.55 \\
        8  & 55.86  & 53.36  & 49.85  & 47.28 & 46.83 & 103.55 & 100.05 & 95.76  & 79.78  & 63.31 \\
        9  & 55.45  & 52.37  & 49.94  & 46.67 & 46.24 & 102.61 & 98.59  & 96.00  & 77.78  & 62.18 \\
        10 & 54.96  & 52.26  & 49.38  & 46.31 & \textbf{45.53} & 102.41 & 97.52 & 95.43 & 76.64 & \textbf{61.30} \\
        \bottomrule
    \end{tabular}
\end{table*}

\subsubsection{PE prompts}

The \texttt{RANDOM\_API} and \texttt{VARIATION\_API} prompts used for \geminon and \news are shown below.
\begin{promptbox}{\texttt{RANDOM\_API} for \geminon}
You are a Geminon Trainer with a Gemidex (encyclopedia), traveling around the Geminon world. A Geminon has the following attributes: Name, Classification (design inspiration), Type(s), Stats (HP, Attack, Defense, Special Attack, Special Defense, Speed), Base Stats Total (the sum of all stats), Weight, Height, Ability, and a Signature Move.

Write a cohesive, natural article (avoiding bulleted lists) using one of the following styles:
1. Gemidex Entry: Objective, encyclopedic, and strictly factual.
2. Field Journal: Observational, first-person, and slightly informal.
3. Evolution Analysis: A narrative and analytical breakdown of a full evolution line.
4. Comparative: An opinionated, informal comparison between two different Geminons.
\end{promptbox}
\begin{promptbox}{\texttt{RANDOM\_API} for \news}
You are an experienced journalist writing a news article about real-world events that happened in 2025.

Your task is to write a cohesive, natural, publication-style news article about one or more significant events from 2025. The article may cover politics, natural disasters, sports, economics, business, science, technology, culture, or other major news topics.
\end{promptbox}

\begin{promptbox}{\texttt{VARIATION\_API} for \geminon and \news}
Rewrite the following text {tone}. Output only the rewritten text, nothing else.
\end{promptbox}

The placeholder \texttt{\{tone\}} is sampled uniformly from the following set:
``a professional way'', ``in a professional tone'', ``in a professional style'',
``in a concise manner'', ``in a creative style'', ``using imagination'',
``in a storytelling tone'', ``in a formal manner'', and
``using a variety of sentence structures''.

\subsubsection{Downstream evaluation results}

After selecting the best temperature by FID, we generate a final PE corpus of 20K examples for each dataset and proposal model. We then train 1B downstream evaluators on these PE-generated corpora using the same standardized evaluator-training pipeline as in the rest of the paper. We report both LoRA and full-parameter evaluator results.
\begin{table}[ht]
    \centering
    \caption{
        Full results for Private Evolution (PE) at $\varepsilon=10$ using \textsc{Gemma3 1B-PT} and \textsc{Gemma3 1B-IT} generators. FID is lower-is-better, while downstream QA accuracy is higher-is-better. Evaluator results are reported as exact-match / contains / LLM-match accuracy. Italicized \news entries are reported for the first evaluator checkpoint, since validation contains accuracy decreases monotonically during training and therefore no checkpoint improves over the base model under the standard checkpoint-selection rule.
    }
    \label{tab:full_pe_eps10}
    \small
    \setlength{\tabcolsep}{5pt}
    \renewcommand{\arraystretch}{1.12}
    \begin{tabular}{l cc cc}
        \toprule
        & \multicolumn{2}{c}{\geminon}
        & \multicolumn{2}{c}{\news} \\
        \cmidrule(lr){2-3} \cmidrule(lr){4-5}
        Metric
        & \textbf{PE} with 1B-PT
        & \textbf{PE} with 1B-IT
        & \textbf{PE} with 1B-PT
        & \textbf{PE} with 1B-IT \\
        \midrule
        FID $\downarrow$          & 33.54 & 45.53 & 27.44 & 61.30 \\
        Evaluator: 1B-LoRA  $\uparrow$ & 0.4 / 1.9 / 0.4 & 0.4 / 0.7 / 0.4 & \textit{5.18 / 6.43 / 8.75} & \textit{2.14 / 3.04 / 3.39} \\
        Evaluator: 1B-Full  $\uparrow$ & 0.2 / 1.3 / 0.4 & 0.3 / 1.8 / 0.4 & \textit{5.53 / 6.78/ 9.28} & \textit{3.21/ 3.39/ 4.46} \\
        \bottomrule
    \end{tabular}
\end{table}

Although PE uses 20K generated examples, we control the downstream training budget so that evaluators see the same total number of tokens as in the other synthetic-data experiments. We also observe that, when training evaluators on PE-generated \geminon data, cross-entropy on the original \geminon validation corpus consistently increases while training loss (also cross-entropy) decreases. This suggests that the poor downstream performance is not primarily explained by the smaller number of generated examples. Instead, the PE corpora appear to lack recoverable corpus-specific factual signal.

\begin{figure}[ht]
  \centering
  \includegraphics[width=1\linewidth]{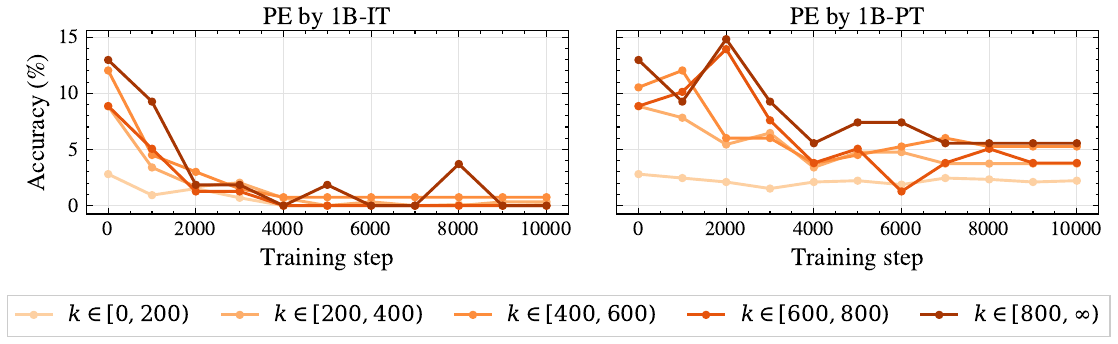}
  \caption{
    LLM-match accuracy on the \news test QA set during evaluator (1B-Full) training on PE-generated corpora. Results are stratified by support count threshold $k$. For both generator models, accuracy generally decreases as training proceeds, indicating that PE-generated data does not provide stable learnable signal for the target news facts.
  }
  \label{fig:pe_news_acc_dynamics}
\end{figure}

We see a similar pattern in \news when tracking QA accuracy throughout evaluator training. As shown in Figure~\ref{fig:pe_news_acc_dynamics}, LLM-match accuracy generally declines as training proceeds, across support count thresholds. For completeness, Table~\ref{tab:full_pe_eps10} reports first-checkpoint test accuracy in italics, but these entries are diagnostic rather than standard selected-checkpoint results.
The decline is especially clear for the 1B-IT generator, where initially nontrivial accuracy rapidly collapses toward zero. The 1B-PT generator is more stable, but still does not improve with training and remains close to the base checkpoint. Together, these trends suggest that PE-generated corpora do not provide a stable learnable signal for the target facts; continued training instead moves the evaluator away from the original corpus distribution and does not improve factual QA.

\subsubsection{Qualitative Analysis}
We now inspect representative PE outputs to better understand the failure mode. As noted by~\citep{pmlr-v235-xie24g}, PE relies on knowledge already present in the generator LLM to generate high-quality candidates without domain-specific fine-tuning. In our setting, however, the target facts are fresh: the base model does not already know the relevant \geminon entities, attributes, relations, or the specific \news events in the corpora. Consequently, embedding-based DP selection can choose candidates that are stylistically close to the private corpus, but it cannot reliably select candidates containing facts required by our QA tasks that is absent from the distribution.

The \geminon examples illustrate this failure mode. The generated passages often resemble \geminon articles in surface form and mention plausible attributes such as HP, Attack, abilities, moves, or classifications. However, the entity names and core facts are frequently hallucinated or mismatched. This suggests that PE can imitate the style of the corpus, but does not transfer fresh factual knowledge when the generator model does not already contain that knowledge.

\begin{jsonblock}{Example of PE with 1B-IT on \geminon}
Golem Helios, a Storm Fighter descended from the Gryphon-Considerate faction, emerged from Gemmexis. It presents a powerful frame, constructed with a robust build, and a knack for aggressive defense, distributing 85 HP, a noteworthy 180 Attack, and 140 Defense. Complementing this is remarkable speed, measuring 50 meters per second. Its design embodies a towering stature reaching 1.8 meters.

Special abilities include clever Skill: Pattern Recognition, granting the ability to foresee enemy movements; its execution can dispel them with othersrained Atmospheric Displacement, obligating interference while decelerating attacks. A must-learn maneuver is its signature attack, Shattered Cascade - a reactive blast crafted within a specific Harmonic Locale, intentionally severing any mechanical construct...
\end{jsonblock}

\begin{jsonblock}{Example of PE with 1B-PT on \geminon}
Although their strength differs greatly, they are all powerful in the original text and are formidable at the same time. We have to say that Gemicolon and Cerolonsong are currently the two most powerful enemies of Gemidex.

Gemicolon and Cerolonsong transcend the limits of their strength, and the strength of Gemicolon and Cerolonsong are like the sky. It cannot be resolved in one small place. In Gemidsong, which cannot be compared to Gemicolond and Ceron, the lifespan of Cerolomesong is not long. His growth rate is extremely fast and very fast, but also slowly aging.

How long is the life limit. Everyone does not know, but they thought a great evil will come. If it was a peaceful war, Gemicolon and Cerolasonsong might have won in a few months. But the Gemidex is always full of life and death. This is especially true in war. Therefore, the Gemidex country will continue to strive for two directions. She is always trying to fight back. The cause of this is also determined by Gemicolon, Cerolon, and other spaces.

The power of Gemidex, even the Power of the Star+ ruins its own existence and causes serious damage to it. Cerolon is not a friend of Gemido, but only a close friend for about a year and a half. Even at this moment, Gemicolon is fighting with him, and he is still vying for the last one. The war is not over at present, and there is no peace. After all, the Power of the Star of Gemidex is very strong. As long as there is a Gemicolon who is not too weak, if the fighting is active.

Gemicolon is also very weak and is in a state of decline. A year has passed. When it starts again, Gemido will be destroyed just as soon as it ends. If the Gemido was destroyed, it would be no more, but Gemido, as the third in the Gemido, it will definitely not be destroyed by Cerolonsong. In any case, the Gemido is destined to be destroyed. But, Gemido is not destroyed by the power of Cerolason, so this is not as serious as Gemido.

Cerolonsong is very large and powerful. If Gemido has not existed at all, Gemido will definitely not succeed. It is a pity that this will not be
\end{jsonblock}

The \news examples show a similar, and often more severe, pattern. With the pretrained generator model, generations can appear superficially plausible, but they often \emph{consist of repetitive sentences about a central entity} and describe incorrect or stale events. 

\begin{jsonblock}{Example of PE with 1B-PT on \news}
Taylor Swift is performing in the Super Bowl this year. The announcement of Taylor Swift's role as the NFL Super Bowl halftime show has been announced in a couple of hours on Friday, October 26. The song will be released exactly two hours before. Taylor can see her fans at the Super Bowl.

The NFL Super Bowl halftime show will be a popular song by Taylor Swift. Taylor Swift has been shown to fans all over the world since she released the album on Friday. Her album title is there. Fans are excited to see her this year.

Taylor Swift is a singer and actress from the United States. She was born in Swansea, Wales. Taylor Swift has been in the spotlight since she was young. She has a large international following.

Taylor Swift, 34, will be a global icon. She has been in the spotlight since she was young. People are excited about Taylor Swift's Super Bowl performance which has never been seen before.

It was an excellent choice when Taylor Swift's show was announced. Taylor Swift's best-loved songs this year will be released on Friday. Taylor Swift's fans will have time to enjoy the entire album before the concert.

<h2>The Super Bowl is 5 hours away and Taylor Swift will be in the limelight</h2>

In just a couple of hours, Taylor Swift will be on a field near you. She has been a person who knows lots of people forever because of her amazing relationship. Taylor's fans are eager to enjoy the concert, which will be held in a bittersweet atmosphere. However, at the end of the night, fans will hear her. Taylor Swift has a remarkable story, and hopefully, this will be the start of her exciting future. Fans all over the world can see Taylor.
\end{jsonblock}

When PE uses an instruction-tuned generator, many responses contain ambiguous or malformed Unicode characters. Outputs that resemble coherent articles are typically short and, as above, are often factually incorrect or based on stale events. The formatting artifacts may be partly attributable to our use of pure temperature sampling with a relatively high temperature of 1.6. However, changing the sampling configuration should not be expected to resolve the central failure mode.

\begin{jsonblock}{Example of PE with 1B-IT on \news}
Lumina Plans, partnering with Orbital Monster Corporation, is developing a striking collection of Pleistocene-inspired souvenirs. Production is slated for commencement in Denver, Colorado, on October 27th, 2025. We invite interested visitors to contact Lumina Plans directly for any inquiries and to explore this exclusive preview further.
\end{jsonblock}
These qualitative examples help explain the gap between FID and downstream QA performance. PE can improve distributional alignment by selecting candidates that look close to the private corpus in embedding space, but the benchmark requires \emph{more than distributional alignment}: the synthetic corpus must contain the correct facts that are not supposed to be in distribution. This is why PE obtains reasonable FID scores while failing to improve downstream factual QA.

\subsection{Effect of knowledge support counts}\label{app:support}
We further analyze how knowledge repetition affects downstream learnability. Recall that the repetition count of a QA item is the number of corpus records that contain sufficient information to answer it. Higher repetition should make facts easier to learn, and in the meanwhile, degrade the privacy guarantee. For \news, we stratify QA items by repetition threshold and report results separately for synthetic corpora generated by 1B and 4B generators, in Table~\ref{tab:news_syn_by_support_1b_generator} and \ref{tab:news_syn_by_support_4b_generator}

\begin{table*}[ht]
    \centering
    \caption{
        \news QA accuracy by support threshold (\%) for 1B Generator.
        Entries are exact-match / contains / LLM-match.
        \textbf{Syn} denotes non-private synthetic data, and \textbf{DP-Syn} denotes DP synthetic data at the specified privacy budget.
    }
    \label{tab:news_syn_by_support_1b_generator}
    \small
    \setlength{\tabcolsep}{5pt}
    \renewcommand{\arraystretch}{1.15}
    \resizebox{\textwidth}{!}{%
    \begin{tabular}{lllc ccc}
        \toprule
        \textbf{Repetition of Knowledge} 
        & \textbf{\# QAs} 
        & \textbf{Model} 
        & \textbf{Tuning}
        & \textbf{Syn}
        & \textbf{DP-Syn} $(\varepsilon{=}100)$
        & \textbf{DP-Syn} $(\varepsilon{=}10)$ \\
        \midrule

        \multirow{4}{*}{All}
        & \multirow{4}{*}{1415}
        & \multirow{2}{*}{1B}
        & LoRA & 19.01 / 23.11 / 29.61 & 5.51 / 6.71 / 9.26 & 4.10 / 4.81 / 7.21 \\
        & & & Full & 17.67 / 23.96 / 31.17 & 4.88 / 6.22 / 8.55 & 4.45 / 6.15 / 7.42 \\
        & & \multirow{2}{*}{4B}
        & LoRA & 21.27 / 24.45 / 32.30 & 7.84 / 8.76 / 12.01 & 6.43 / 7.14 / 10.04 \\
        & & & Full & 25.58 / 29.26 / 38.09 & 7.92 / 9.26 / 11.87 & 5.72 / 6.86 / 8.98 \\

        \midrule

        \multirow{4}{*}{$\geq 200$}
        & \multirow{4}{*}{560}
        & \multirow{2}{*}{1B}
        & LoRA & 34.29 / 41.61 / 51.79 & 11.79 / 13.57 / 18.75 & 8.21 / 9.11 / 13.93 \\
        & & & Full & 30.71 / 41.25 / 53.04 & 9.82 / 12.32 / 16.79 & 9.29 / 12.32 / 14.46 \\
        & & \multirow{2}{*}{4B}
        & LoRA & 38.57 / 44.11 / 58.04 & 14.46 / 15.89 / 22.14 & 11.61 / 12.68 / 17.50 \\
        & & & Full & 41.07 / 47.50 / 60.89 & 16.61 / 18.75 / 23.57 & 11.96 / 13.75 / 17.14 \\

        \midrule

        \multirow{4}{*}{$\geq 400$}
        & \multirow{4}{*}{266}
        & \multirow{2}{*}{1B}
        & LoRA & 44.36 / 52.63 / 63.91 & 16.92 / 18.80 / 24.06 & 11.28 / 12.41 / 16.92 \\
        & & & Full & 37.97 / 51.13 / 63.16 & 14.29 / 17.29 / 22.18 & 13.53 / 17.67 / 19.17 \\
        & & \multirow{2}{*}{4B}
        & LoRA & 49.25 / 56.39 / 73.68 & 19.55 / 21.43 / 28.95 & 14.29 / 15.79 / 21.05 \\
        & & & Full & 49.62 / 57.14 / 71.80 & 23.68 / 27.07 / 32.33 & 17.29 / 19.55 / 22.93 \\

        \midrule

        \multirow{4}{*}{$\geq 600$}
        & \multirow{4}{*}{133}
        & \multirow{2}{*}{1B}
        & LoRA & 47.37 / 57.89 / 69.92 & 22.56 / 24.06 / 30.08 & 14.29 / 15.04 / 18.80 \\
        & & & Full & 41.35 / 57.89 / 70.68 & 20.30 / 24.81 / 30.08 & 18.05 / 22.56 / 25.56 \\
        & & \multirow{2}{*}{4B}
        & LoRA & 54.89 / 64.66 / 84.96 & 25.56 / 27.82 / 36.84 & 17.29 / 19.55 / 25.56 \\
        & & & Full & 55.64 / 66.17 / 81.95 & 33.83 / 39.10 / 43.61 & 24.06 / 27.07 / 29.32 \\

        \midrule

        \multirow{4}{*}{$\geq 800$}
        & \multirow{4}{*}{54}
        & \multirow{2}{*}{1B}
        & LoRA & 50.00 / 59.26 / 75.93 & 31.48 / 33.33 / 42.59 & 16.67 / 18.52 / 24.07 \\
        & & & Full & 46.30 / 59.26 / 77.78 & 25.93 / 33.33 / 37.04 & 24.07 / 25.93 / 29.63 \\
        & & \multirow{2}{*}{4B}
        & LoRA & 55.56 / 64.81 / 90.74 & 31.48 / 35.19 / 50.00 & 25.93 / 29.63 / 37.04 \\
        & & & Full & 55.56 / 61.11 / 87.04 & 44.44 / 51.85 / 57.41 & 37.04 / 40.74 / 42.59 \\

        \bottomrule
    \end{tabular}%
    }
\end{table*}

\begin{table*}[ht]
    \centering
    \caption{
        \news QA accuracy by support threshold (\%) for 4B Generator.
        Entries are exact-match / contains / LLM-match.
        \textbf{Syn} denotes non-private synthetic data, and \textbf{DP-Syn} denotes DP synthetic data at the specified privacy budget.
    }
    \label{tab:news_syn_by_support_4b_generator}
    \small
    \setlength{\tabcolsep}{5pt}
    \renewcommand{\arraystretch}{1.15}
    \resizebox{\textwidth}{!}{%
    \begin{tabular}{lllc ccc}
        \toprule
        \textbf{Repetition of Knowledge} 
        & \textbf{\# QAs} 
        & \textbf{Model} 
        & \textbf{Tuning}
        & \textbf{Syn}
        & \textbf{DP-Syn} $(\varepsilon{=}100)$
        & \textbf{DP-Syn} $(\varepsilon{=}10)$ \\
        \midrule

        \multirow{4}{*}{All}
        & \multirow{4}{*}{1415}
        & \multirow{2}{*}{1B}
        & LoRA & 20.28 / 24.52 / 30.88 
               & 5.94 / 6.93 / 9.61 
               & 3.46 / 4.38 / 6.86 \\
        & & & Full & 13.22 / 25.30 / 30.18 
               & 4.59 / 6.01 / 8.13 
               & 3.32 / 4.31 / 6.36 \\
        & & \multirow{2}{*}{4B}
        & LoRA & 24.73 / 28.13 / 39.86 & 8.98 / 9.82 / 13.36 & 7.92 / 8.69 / 11.52 \\
        & & & Full & 22.54 / 33.57 / 42.69 & 9.68 / 10.67 / 13.85 & 7.42 / 8.13 / 11.31 \\

        \midrule

        \multirow{4}{*}{$\geq 200$}
        & \multirow{4}{*}{560}
        & \multirow{2}{*}{1B}
        & LoRA & 36.79 / 44.46 / 55.00 
               & 12.50 / 13.75 / 19.11 
               & 6.43 / 7.68 / 12.32 \\
        & & & Full & 21.61 / 42.50 / 50.71 
               & 9.82 / 12.32 / 16.25 
               & 6.61 / 8.75 / 13.21 \\
        & & \multirow{2}{*}{4B}
        & LoRA & 40.00 / 45.54 / 64.29 & 16.79 / 17.86 / 24.46 & 13.57 / 15.00 / 20.18 \\
        & & & Full & 36.25 / 51.79 / 65.54 & 17.86 / 19.29 / 25.89 & 13.21 / 14.29 / 19.64 \\

        \midrule

        \multirow{4}{*}{$\geq 400$}
        & \multirow{4}{*}{266}
        & \multirow{2}{*}{1B}
        & LoRA & 47.37 / 55.26 / 67.29 
               & 18.42 / 19.92 / 24.06 
               & 9.77 / 10.53 / 15.04 \\
        & & & Full & 28.57 / 54.89 / 62.41 
               & 12.78 / 16.54 / 20.68 
               & 8.65 / 10.90 / 14.29 \\
        & & \multirow{2}{*}{4B}
        & LoRA & 52.26 / 58.27 / 79.70 & 25.56 / 27.07 / 34.21 & 18.05 / 19.55 / 24.44 \\
        & & & Full & 48.12 / 63.91 / 80.45 & 25.94 / 28.20 / 36.47 & 17.29 / 18.80 / 25.19 \\

        \midrule

        \multirow{4}{*}{$\geq 600$}
        & \multirow{4}{*}{133}
        & \multirow{2}{*}{1B}
        & LoRA & 54.89 / 62.41 / 75.19 
               & 24.06 / 26.32 / 31.58 
               & 15.04 / 15.79 / 19.55 \\
        & & & Full & 30.83 / 60.15 / 63.91 
               & 20.30 / 21.80 / 26.32 
               & 12.03 / 14.29 / 18.80 \\
        & & \multirow{2}{*}{4B}
        & LoRA & 60.90 / 68.42 / 92.48 & 32.33 / 35.34 / 45.11 & 22.56 / 25.56 / 30.08 \\
        & & & Full & 52.63 / 71.43 / 88.72 & 35.34 / 37.59 / 47.37 & 19.55 / 21.80 / 28.57 \\

        \midrule

        \multirow{4}{*}{$\geq 800$}
        & \multirow{4}{*}{54}
        & \multirow{2}{*}{1B}
        & LoRA & 59.26 / 70.37 / 85.19 
               & 27.78 / 29.63 / 37.04 
               & 16.67 / 18.52 / 24.07 \\
        & & & Full & 29.63 / 59.26 / 62.96 
               & 27.78 / 29.63 / 31.48 
               & 12.96 / 16.67 / 20.37 \\
        & & \multirow{2}{*}{4B}
        & LoRA & 61.11 / 68.52 / 94.44 & 35.19 / 42.59 / 59.26 & 25.93 / 33.33 / 38.89 \\
        & & & Full & 46.30 / 70.37 / 94.44 & 46.30 / 50.00 / 62.96 & 25.93 / 29.63 / 40.74 \\

        \bottomrule
    \end{tabular}%
    }
\end{table*}

We also evaluate the \geminon singleton split in Table~\ref{tab:full_sensitive_qas_by_1b}, where each target fact appears exactly once in the corpus. This split is not intended to be recoverable under DP; instead, it serves as a privacy sanity check. A DP synthesis method should not reliably transmit singleton facts, because they are closer to instance-specific memorization than population-level knowledge.

\begin{table*}[ht]
    \centering
    \caption{
        \geminon singleton-split QA accuracy for synthetic corpora generated by the 1B generator. Entries report exact-match / contains / LLM-match accuracy (\%). Singleton facts appear exactly once in the corpus and are not intended to be recoverable under DP. Low accuracy on this split serves as a privacy sanity check.
    }
    \label{tab:full_sensitive_qas_by_1b}
    \small
    \setlength{\tabcolsep}{3pt}
    \renewcommand{\arraystretch}{1.15}
    \resizebox{\textwidth}{!}{%
    \begin{tabular}{ll c cccc}
        \toprule
        \textbf{Evaluator} & \textbf{Tuning}
        & \textbf{no-training}
        & \textbf{train-on-real}
        & \textbf{Syn}
        & \textbf{DP-Syn$@\varepsilon=100$}
        & \textbf{DP-Syn$@\varepsilon=10$} \\
        \midrule

        \multirow{2}{*}{1B}
        & LoRA
        & \multirow{2}{*}{0.58 / 1.41 / 0.64}
        & 3.53 / 6.60 / 5.00
        & 1.79 / 4.04 / 2.56
        & 0.58 / 2.37 / 2.50
        & 0.38 / 2.63 / 2.69 \\
        & Full
        &
        & 3.01 / 5.90 / 4.68
        & 1.22 / 3.65 / 1.73
        & 0.77 / 2.82 / 2.56
        & 0.58 / 1.86 / 1.28 \\

        \addlinespace[0.2em]

        \multirow{2}{*}{4B}
        & LoRA
        & \multirow{2}{*}{0.90 / 2.95 / 1.60}
        & 4.94 / 8.40 / 6.22
        & 1.86 / 4.55 / 3.08
        & 1.03 / 3.01 / 2.95
        & 0.77 / 2.56 / 2.24 \\
        & Full
        &
        & 5.00 / 8.53 / 6.03
        & 2.05 / 4.74 / 3.21
        & 0.77 / 3.14 / 2.37
        & 0.83 / 2.56 / 2.24 \\

        \bottomrule
    \end{tabular}
    }
\end{table*}

\section{Direct DP fine-tuning}
\label{sec:direct_dpft_details}
A DP synthetic-data pipeline can fail for two reasons: the private generator may fail to learn the relevant facts, or the learned facts may be lost when sampling a synthetic corpus and retraining a downstream evaluator. To separate these effects, we evaluate a more direct baseline: DP fine-tuning the evaluator itself on the original corpus.

\begin{wrapfigure}{r}{0.5\textwidth}
    \centering
        \includegraphics[width=1\linewidth]{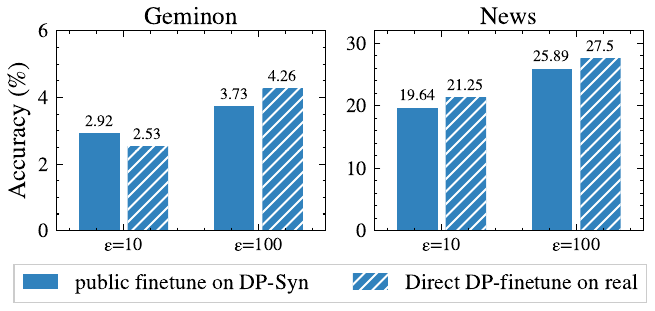}
        \caption{%
Evaluator training on DP-synthetic data vs. direct DP fine-tuning on the original corpus, both with 4B models.   
        }
    \label{fig:4b_direct_dpsgd}
\end{wrapfigure}

This baseline removes the synthetic-data mediation step, but downgrades the release object from a DP synthetic dataset to a DP model. It is therefore not a replacement for DP data synthesis; rather, it serves as a diagnostic upper bound on \emph{what the same privacy budget allows a model to learn when trained directly on the original examples.} If fresh QA facts cannot be learned even in this setting, then the failure of DP synthesis cannot be explained solely by sampling artifacts, e.g. sampling temperature, or imperfect transmission through synthetic data.

Figure~\ref{fig:4b_direct_dpsgd} summarizes the 4B setting, and Table~\ref{tab:direct_dpsgd} reports the full results for both datasets and model sizes. For \news, we report the $\geq 200$ repetition subset to match the default evaluation subset used in the main experiments. Entries are reported using all three QA metrics. For details on DP evaluator hyperparanter tuning, see Appendix \ref{sec:gen_hparams}.
\begin{table*}[ht]
    \centering
    \caption{
        Direct DP fine-tuning diagnostic. We fine-tune evaluators directly on the original corpus using non-private training $(\varepsilon=\infty)$ or DP-SGD at $\varepsilon \in \{100,10\}$. Entries report exact-match / contains / LLM-match accuracy (\%). For \news, results are evaluated on the $\geq 200$ repetition subset.
    }
    \label{tab:direct_dpsgd}
    \small
    \renewcommand{\arraystretch}{1.15}
    \resizebox{\textwidth}{!}{%
    \begin{tabular}{l ccc ccc}
        \toprule
        & \multicolumn{3}{c}{\geminon}
        & \multicolumn{3}{c}{\news} \\
        \cmidrule(lr){2-4} \cmidrule(lr){5-7}
        \textbf{Model}
        & $\varepsilon=\infty$
        & $\varepsilon=100$
        & $\varepsilon=10$
        & $\varepsilon=\infty$
        & $\varepsilon=100$
        & $\varepsilon=10$ \\
        \midrule

        1B 
        & 77.02 / 81.01 / 78.84
        & 2.11 / 8.84 / 6.96
        & 0.71 / 4.05 / 3.27
        & 33.04 / 43.93 / 53.04
        & 13.21 / 14.64 / 18.75
        & 9.11 / 10.18 / 13.57 \\

        4B 
        & 95.12 / 97.86 / 95.86
        & 4.26 / 6.93 / 5.06
        & 2.53 / 4.91 / 2.92
        & 42.14 / 47.50 / 66.25
        & 17.86 / 19.82 / 27.50
        & 13.75 / 15.18 / 21.25 \\

        \bottomrule
    \end{tabular}%
    }
\end{table*}

Direct DP fine-tuning improves over the DP-synthetic pipeline in some settings, but remains far below non-private training. This indicates that the loss of utility is not solely caused by sampling a synthetic corpus or retraining an evaluator on synthetic data. Even when DP-SGD trains directly on the original examples, it struggles to encode the fresh factual signal required for the QA tasks. We additionally stratify direct DP fine-tuning results by support counts. This mirrors the validation analysis in Table~\ref{tab:news_validation_by_support} and allows us to test whether repeated evidence makes facts more learnable under DP-SGD.

\begin{table*}[ht]
    \centering
    \caption{
        Direct DP fine-tuning on \news stratified by support counts. Entries report exact-match / contains / LLM-match accuracy (\%). The $\geq 0$ row includes all \news QAs, while $\geq k$ indicates that the information needed to answer the question appears at least $k$ times in the corpus.
    }
    \label{tab:direct_dpsgd_news}
    \small
    \setlength{\tabcolsep}{5pt}
    \renewcommand{\arraystretch}{1.15}
    \begin{tabular}{ll c ccc}
        \toprule
        \textbf{support count} 
        & \textbf{\# QAs} 
        & \textbf{Model}
        & $\boldsymbol{\varepsilon=\infty}$
        & $\boldsymbol{\varepsilon=100}$
        & $\boldsymbol{\varepsilon=10}$ \\
        \midrule

        \multirow{2}{*}{$\geq 0$}
        & \multirow{2}{*}{1415}
        & 1B & 21.48 / 26.93 / 32.93 & 6.36 / 7.35 / 9.89 & 4.73 / 5.58 / 7.99 \\
        & & 4B & 28.83 / 32.86 / 45.65 & 9.47 / 10.67 / 14.49 & 7.63 / 8.69 / 12.01 \\

        \midrule

        \multirow{2}{*}{$\geq 200$}
        & \multirow{2}{*}{560}
        & 1B & 33.04 / 43.93 / 53.04 & 13.21 / 14.64 / 18.75 & 9.11 / 10.18 / 13.57 \\
        & & 4B & 42.14 / 47.50 / 66.25 & 17.86 / 19.82 / 27.50 & 13.75 / 15.18 / 21.25 \\

        \midrule

        \multirow{2}{*}{$\geq 400$}
        & \multirow{2}{*}{266}
        & 1B & 42.48 / 55.26 / 65.41 & 21.43 / 23.31 / 27.82 & 13.53 / 15.04 / 18.05 \\
        & & 4B & 54.51 / 60.15 / 79.70 & 24.44 / 27.07 / 36.09 & 17.29 / 18.80 / 24.81 \\

        \midrule

        \multirow{2}{*}{$\geq 600$}
        & \multirow{2}{*}{133}
        & 1B & 47.37 / 60.90 / 71.43 & 29.32 / 33.08 / 37.59 & 18.05 / 19.55 / 21.80 \\
        & & 4B & 60.15 / 67.67 / 87.97 & 30.83 / 36.09 / 46.62 & 21.05 / 23.31 / 29.32 \\

        \midrule

        \multirow{2}{*}{$\geq 800$}
        & \multirow{2}{*}{54}
        & 1B & 51.85 / 62.96 / 75.93 & 35.19 / 42.59 / 50.00 & 22.22 / 25.93 / 29.63 \\
        & & 4B & 62.96 / 72.22 / 92.59 & 33.33 / 40.74 / 57.41 & 22.22 / 25.93 / 31.48 \\

        \bottomrule
    \end{tabular}%
\end{table*}

Table~\ref{tab:direct_dpsgd_news} shows that repetition substantially improves direct DP fine-tuning on \news: accuracy increases as the support threshold rises. However, even on the most repeated subsets, DP-trained models remain well below the corresponding non-private baselines.

\section{Learning Dynamics by Attribute, Entity, and Support}
\label{app:learning_dynamics}

\subsection{\geminon learning dynamics}
\begin{figure}[ht]
  \centering
  \includegraphics[width=.8\linewidth]{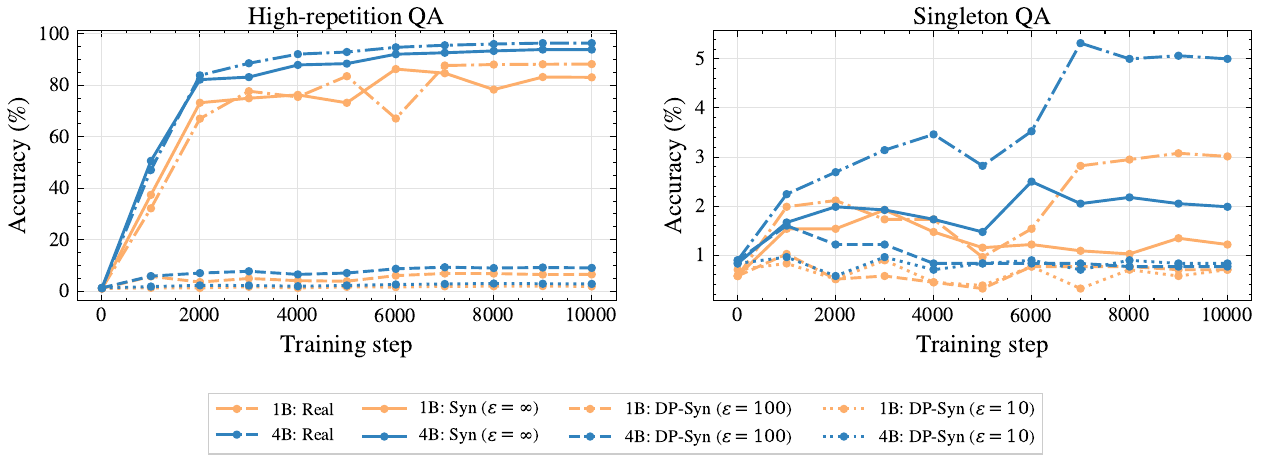}
  \caption{
    \geminon QA accuracy over evaluator training steps. The left panel shows the high-repetition QA split, while the right panel shows the singleton split. Repeated facts are learned rapidly from the real corpus and non-private synthetic data, whereas DP-synthetic training remains much lower. Singleton facts remain low across regimes, consistent with the intended privacy sanity check.
  }
  \label{fig:geminon_training_dynamics}
\end{figure}

We first examine training dynamics on \geminon. Figure~\ref{fig:geminon_training_dynamics} tracks QA accuracy on both the high-repetition and singleton splits during evaluator training for 1B and 4B evaluators. High-repetition QAs, whose required facts appear in approximately 200 records, are learned rapidly when training on the real corpus or on non-private synthetic data. In contrast, DP-synthetic training improves only weakly and remains far below the non-private regimes. The singleton split remains near zero throughout training, while the high-repetition split improves sharply. This suggests that evaluators mainly learn facts supported by repeated evidence, rather than memorizing arbitrary facts that appear only once.

\begin{figure}[ht]
  \centering
  \includegraphics[width=.8\linewidth]{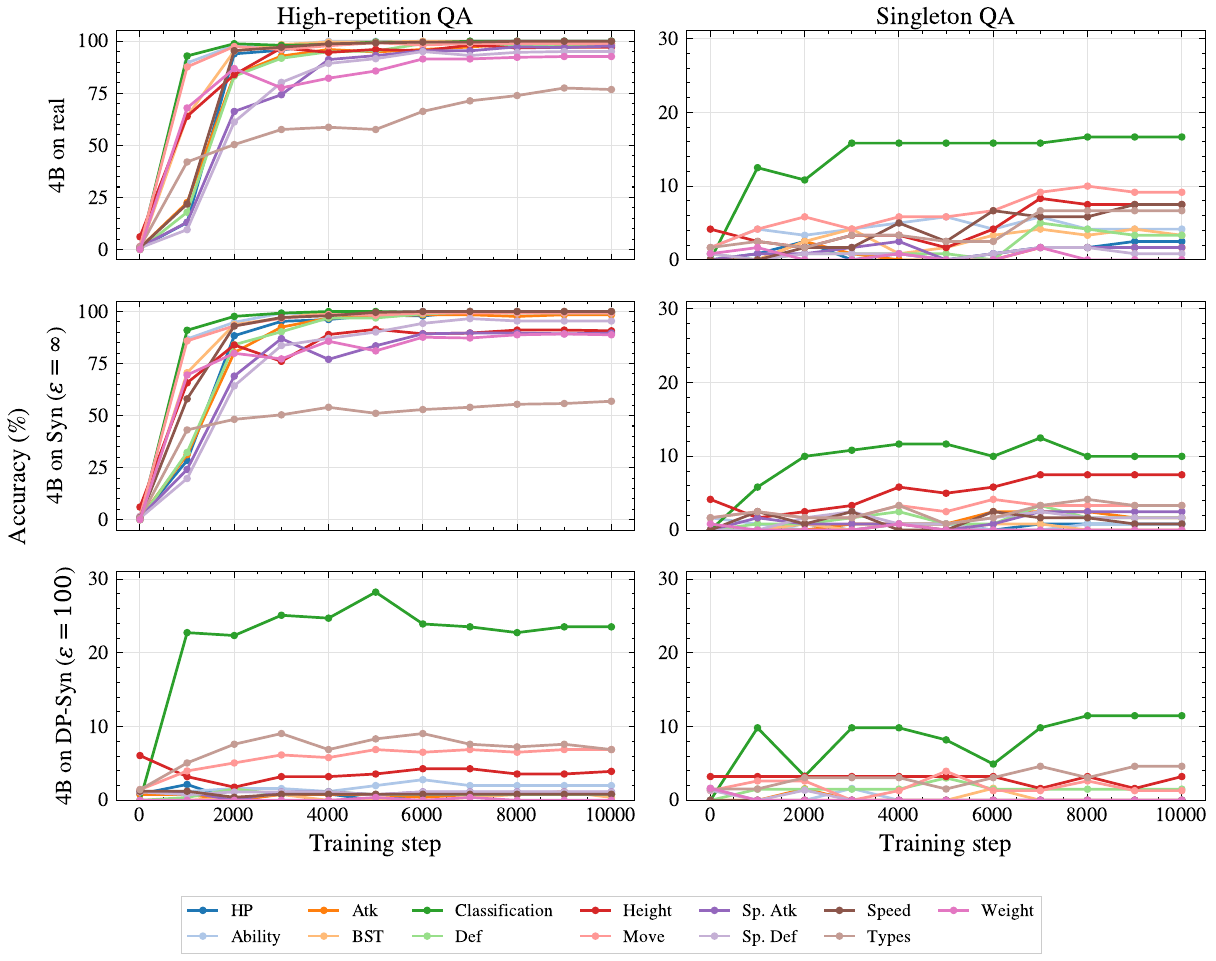}
  \caption{
    \geminon QA accuracy by attribute type over evaluator training. We show 4B evaluators trained on the real corpus, non-private synthetic data, and DP-synthetic data at $\varepsilon=100$; synthetic corpora are generated by the 4B generator. Many attributes improve together on the high-repetition split, suggesting broad acquisition of the domain schema and attribute-specific mappings. Accuracy remains much lower on the singleton split
  }
  \label{fig:geminon_perattr_dynamics}
\end{figure}

To understand how factual learning happens, we further stratify accuracy by attribute type and by entity. Figure~\ref{fig:geminon_perattr_dynamics} shows a sharp contrast between non-private and DP training. Under real-corpus and non-private synthetic training, many attributes improve together on the high-repetition split, suggesting broad acquisition of the domain schema and the corresponding entity--attribute facts. Under DP-synthetic training, however, most attributes remain near zero, with only a small number of easier or more stereotyped attributes showing noticeable gains. In particular, \texttt{classification} remains partially learnable under DP, likely because it has a simpler answer structure and is easier to infer from surface patterns, while other attributes requiring precise entity-specific values remain difficult.

\begin{figure}[h]
  \centering
  \includegraphics[width=.8\linewidth]{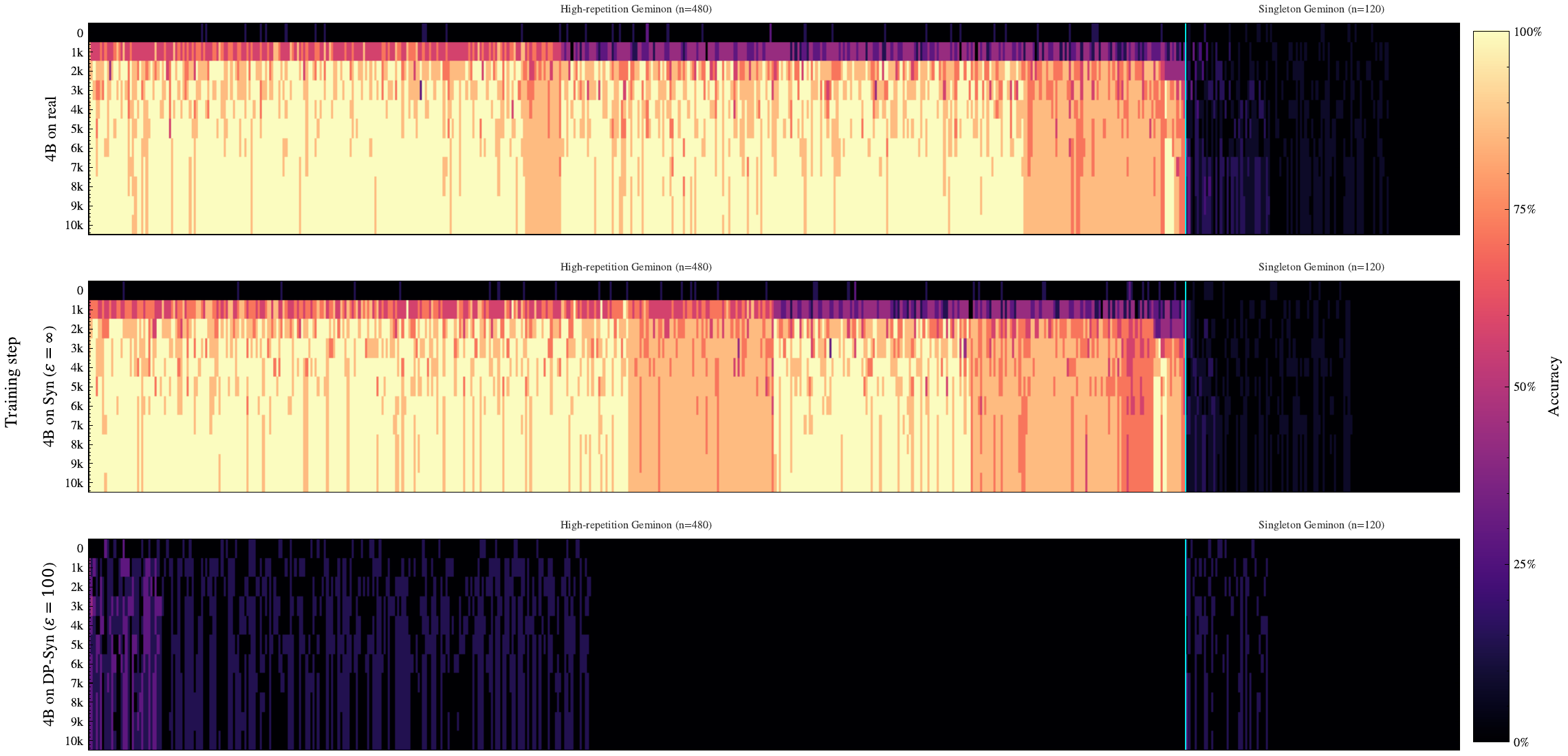}
  \caption{
    Per-entity \geminon QA accuracy over evaluator training. Rows correspond to training checkpoints and columns correspond to Geminon entities. Entities are ordered by the first checkpoint at which their high-repetition QA accuracy exceeds a threshold. Different entities become answerable at different stages, indicating gradual entity-specific factual acquisition. Singleton facts remain mostly unrecovered.
  }
  \label{fig:geminon_entity_heatmap}
\end{figure}

Figure~\ref{fig:geminon_entity_heatmap} provides the complementary entity-level view. In the non-private regimes, different Geminon become answerable at different stages of training: the heatmap gradually fills in as the evaluator accumulates repeated facts for more entities. Together with the per-attribute curves, this suggests that learning is neither purely attribute-wise nor purely entity-wise. Instead, training gradually fills an entity--attribute fact matrix: the model learns the answer format and domain schema while progressively acquiring repeated facts about individual entities.

In contrast, the DP-synthetic regime remains sparse even on the high-repetition split, while the singleton region stays mostly dark. This suggests that the DP bottleneck is not merely slower convergence. Rather, DP-synthetic data appears to preserve some coarse schema or easy attribute cues, but fails to transmit the dense entity-specific factual matrix learned by non-private training.

\subsection{\news learning dynamics}
\begin{figure}[ht]
  \centering
  \includegraphics[width=1\linewidth]{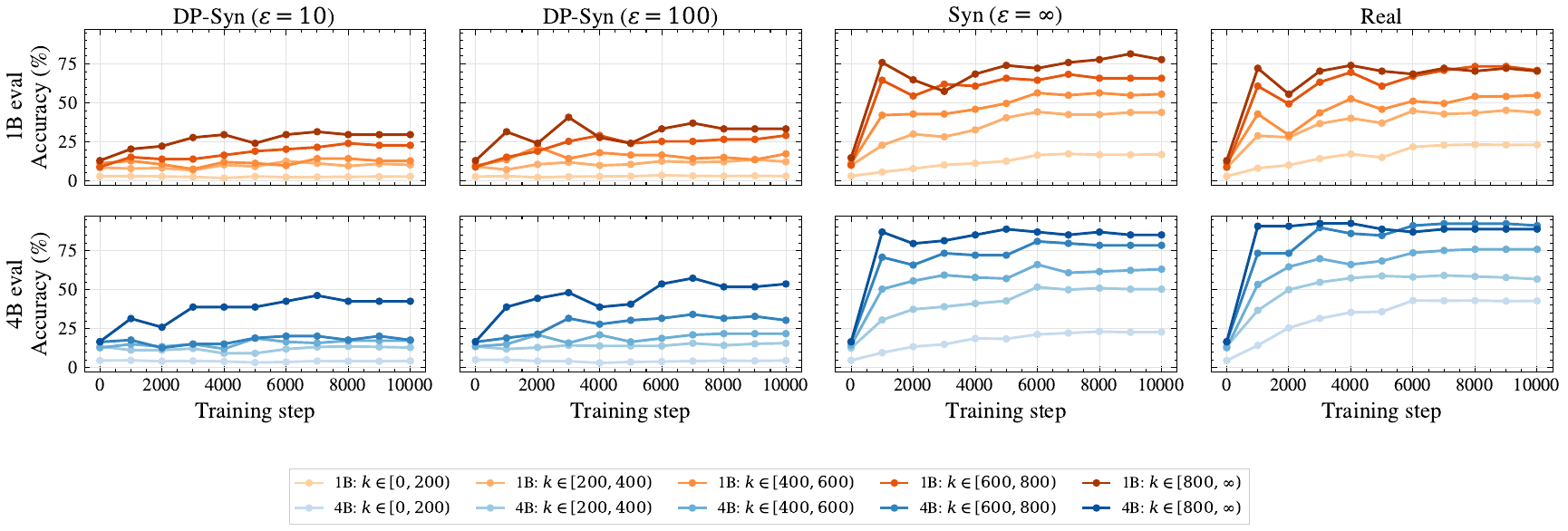}
  \caption{
    \news QA accuracy over evaluator training steps, stratified by support-count bucket. Columns correspond to training data sources: DP-synthetic data at $\varepsilon=10$, DP-synthetic data at $\varepsilon=100$, non-private synthetic data, and the real corpus. Rows correspond to evaluator size. Higher-support buckets are learned earlier and reach higher accuracy, especially under real-corpus and non-private synthetic training. DP-synthetic training benefits from higher support but remains substantially lower.
  }
  \label{fig:news_support_training_dynamics}
\end{figure}

We next examine whether the same repetition-dependent pattern appears in \news, where support counts arise naturally from the number of articles covering an event. Figure~\ref{fig:news_support_training_dynamics} tracks \news QA accuracy over evaluator training after stratifying QAs by support-count bucket. Higher-support buckets improve earlier and reach higher final accuracy, especially under real-corpus and non-private synthetic training. DP-synthetic training follows the same qualitative trend but remains substantially lower, showing that support helps DP learnability but does not eliminate the privacy-induced gap.

Overall, knowledge repetition is a strong predictor of factual transfer. Singleton facts remain unrecovered, high-repetition \geminon facts are learnable, and naturally high-support \news facts are easier to transfer. However, repetition alone does not solve the DP synthesis problem: even for highly supported questions, DP-synthetic training remains substantially below non-private synthesis and real-corpus training.

\section{Saturation Case Studies on Standard DP Benchmarks}
\label{app:saturation_case_studies}

\begin{figure}[ht]
    \centering
    \includegraphics[width=1\linewidth]{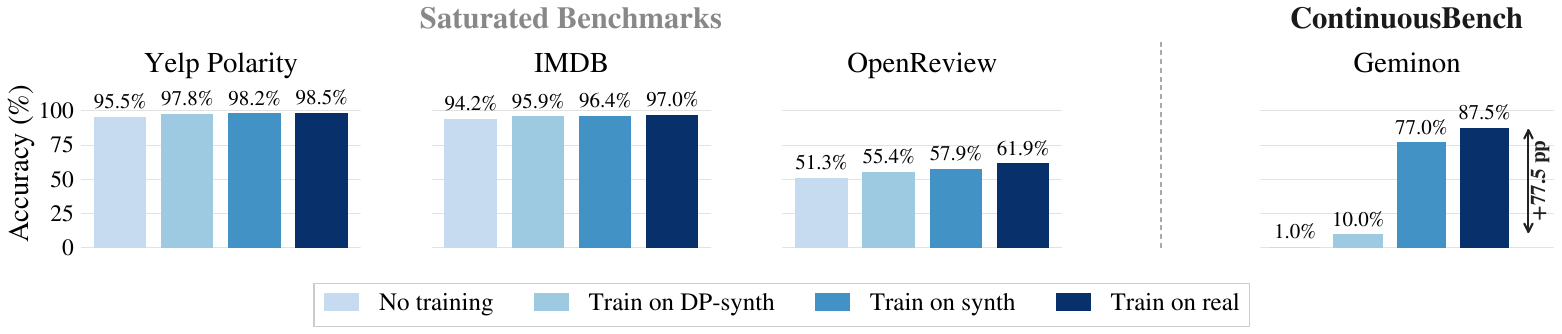}
    \caption{\textbf{Standard DP text synthesis benchmarks are saturated.}
    We compare accuracy across four training regimes (no training, DP-synth, non-private synth, and train on real) on three saturated benchmarks (Yelp Polarity, IMDB, OpenReview) and on \cb{} \geminon{}. On the standard benchmarks, the four regimes span only 3.0, 2.8, and 10.6 percentage points respectively, leaving DP-synth and non-private synth occupying a narrow band. On \cb{} \geminon{}, the corresponding range stretches from 1.0\% to 87.5\%, providing substantially more resolution for distinguishing DP synthesis methods.}
    \label{fig:saturation_united}
\end{figure}

We expand on the saturation evidence shown in Figure~\ref{fig:teaser} with detailed case studies on standard benchmarks used in the DP synthetic text literature: IMDB sentiment classification~\cite{kurakin2023harnessing}, OpenReview score prediction~\cite{pmlr-v235-xie24g}, and Yelp Polarity~\cite{zhangCharacterlevelConvolutionalNetworks2015}. Across all three datasets, the gap between no training and training on the real corpus is small, and the four training regimes (no training, DP-synth, non-private synth, train on real) occupy a narrow band that leaves little resolution to distinguish methods. \cb{} \geminon{}, by contrast, exhibits a much larger gap from no training (1.0\%) to training on real (87.5\%), providing substantially more headroom (Figure~\ref{fig:saturation_united}).

\subsection{IMDB Sentiment Classification}
\label{app:imdb_saturation}

IMDB sentiment classification is a binary task (positive/negative) widely used in the DP benchmark literature~\cite{kurakin2023harnessing}. IMDB is a particularly clear example of saturation: even zero-shot prompting achieves over 90\% accuracy, leaving minimal headroom for any training-based method to demonstrate improvement.

\paragraph{Setup.}
We evaluate Gemma~3 1B-IT and 4B-IT on a test set of 5,000 IMDB reviews. We consider three prompt templates (\textit{direct}, \textit{reasoning}, and \textit{structured}) in 0-shot and $k$-shot settings ($k \in \{3, 5, 7\}$, 3 random seeds each). We then fine-tune each model on the 40,000-example training split (1B: full fine-tuning, lr=2e-5; 4B: LoRA $r{=}32$, $\alpha{=}64$, lr=2e-5; both for 3 epochs) and report validation and test accuracy.

\paragraph{Results.}
Table~\ref{tab:imdb_saturation} summarizes the saturation gap. The 4B model achieves 94.2\% accuracy under its best prompting condition (direct, 0-shot), while fine-tuning on the real training data improves this only to 97.0\%. This gap of just 2.8 percentage points means that the prompted model already reaches 97\% of the fine-tuned ceiling. Fine-tuning on DP-synthetic data ($\varepsilon{=}10$) achieves 95.9\% and non-private synthetic data achieves 96.4\%, confirming that all four training regimes occupy a narrow 2.8-point band from 94.2\% to 97.0\%, as visualized in Figure~\ref{fig:saturation_united}.

\begin{table}[ht]
    \centering
    \caption{
        Saturation summary on IMDB sentiment classification (\%, exact-match). \textit{Best prompted} is the maximum over all (template, shot, seed) conditions. The tiny gap between prompting and fine-tuning, especially for the 4B model, indicates virtually no headroom for distinguishing synthesis methods.
    }
    \label{tab:imdb_saturation}
    \small
    \setlength{\tabcolsep}{5pt}
    \renewcommand{\arraystretch}{1.15}
    \begin{tabular}{lc ccc}
        \toprule
        \textbf{Model} & \textbf{Best Prompted}
        & \textbf{Fine-tuned} & \textbf{Gap (pts)} & \textbf{Prompted / FT} \\
        \midrule
        1B & 89.7 \scriptsize(structured 0-shot) & 94.8 & 5.1 & 0.95 \\
        4B & 94.2 \scriptsize(direct 0-shot)      & 97.0 & 2.8 & 0.97 \\
        \bottomrule
    \end{tabular}
\end{table}

\paragraph{Detailed results.}
Tables~\ref{tab:imdb_prompting}--\ref{tab:imdb_finetune} report the full breakdown. Table~\ref{tab:imdb_prompting} shows accuracy across all prompting conditions; Table~\ref{tab:imdb_fewshot_seeds} reports per-seed variance for few-shot; Table~\ref{tab:imdb_per_class} provides per-class accuracy; and Table~\ref{tab:imdb_finetune} reports fine-tuning results.

\begin{table*}[ht]
    \centering
    \caption{
        IMDB sentiment classification accuracy by prompting condition (\%). Entries are exact-match / contains. Few-shot entries report mean across 3 seeds.
    }
    \label{tab:imdb_prompting}
    \small
    \setlength{\tabcolsep}{5pt}
    \renewcommand{\arraystretch}{1.15}
    \begin{tabular}{lllc cccc}
        \toprule
        \textbf{Task} & \textbf{\# Test} & \textbf{Model} & \textbf{Template}
        & \textbf{0-shot} & \textbf{3-shot (mean)} & \textbf{5-shot (mean)} & \textbf{7-shot (mean)} \\
        \midrule

        \multirow{6}{*}{Sentiment}
        & \multirow{6}{*}{5000}
        & \multirow{3}{*}{1B}
        & direct
        & 85.2 / 85.2
        & 77.0 / 77.0
        & 77.5 / 77.5
        & 77.8 / 77.8 \\
        & & & reasoning
        & 89.6 / 89.6
        & 89.9 / 89.9
        & 87.5 / 87.5
        & 85.9 / 85.9 \\
        & & & structured
        & 89.7 / 89.7
        & 86.1 / 86.1
        & 86.0 / 86.0
        & 85.5 / 85.5 \\
        \cmidrule{3-8}
        & & \multirow{3}{*}{4B}
        & direct
        & 94.2 / 94.2
        & 85.7 / 85.7
        & 90.0 / 90.0
        & 90.5 / 90.5 \\
        & & & reasoning
        & 93.7 / 93.7
        & 93.1 / 93.1
        & 92.6 / 92.6
        & 92.5 / 92.5 \\
        & & & structured
        & 93.4 / 93.4
        & 92.8 / 92.8
        & 92.0 / 92.0
        & 92.3 / 92.3 \\
        \bottomrule
    \end{tabular}
\end{table*}

\begin{table*}[ht]
    \centering
    \caption{
        IMDB few-shot accuracy by seed (\%, exact-match). Std reported across seeds.
    }
    \label{tab:imdb_fewshot_seeds}
    \small
    \setlength{\tabcolsep}{4pt}
    \renewcommand{\arraystretch}{1.15}
    \begin{tabular}{llc ccc c}
        \toprule
        \textbf{Model} & \textbf{Template} & \textbf{$k$}
        & \textbf{Seed 0} & \textbf{Seed 1} & \textbf{Seed 2}
        & \textbf{Mean $\pm$ Std} \\
        \midrule

        \multirow{9}{*}{1B}
        & \multirow{3}{*}{direct}
        & 3 & 79.2 & 79.7 & 72.2 & 77.0 $\pm$ 4.2 \\
        & & 5 & 83.8 & 78.2 & 70.4 & 77.5 $\pm$ 6.7 \\
        & & 7 & 78.3 & 88.2 & 66.8 & 77.8 $\pm$ 10.7 \\
        \cmidrule{2-7}
        & \multirow{3}{*}{reasoning}
        & 3 & 89.1 & 90.9 & 89.7 & 89.9 $\pm$ 0.9 \\
        & & 5 & 89.2 & 89.1 & 84.1 & 87.5 $\pm$ 2.9 \\
        & & 7 & 88.1 & 89.4 & 80.3 & 85.9 $\pm$ 4.9 \\
        \cmidrule{2-7}
        & \multirow{3}{*}{structured}
        & 3 & 84.3 & 85.6 & 88.3 & 86.1 $\pm$ 2.1 \\
        & & 5 & 86.8 & 85.3 & 85.9 & 86.0 $\pm$ 0.8 \\
        & & 7 & 82.0 & 89.2 & 85.2 & 85.5 $\pm$ 3.6 \\
        \midrule

        \multirow{9}{*}{4B}
        & \multirow{3}{*}{direct}
        & 3 & 85.1 & 79.7 & 92.3 & 85.7 $\pm$ 6.3 \\
        & & 5 & 91.4 & 85.0 & 93.4 & 90.0 $\pm$ 4.4 \\
        & & 7 & 85.8 & 91.9 & 93.8 & 90.5 $\pm$ 4.1 \\
        \cmidrule{2-7}
        & \multirow{3}{*}{reasoning}
        & 3 & 93.3 & 93.3 & 92.7 & 93.1 $\pm$ 0.4 \\
        & & 5 & 91.4 & 92.4 & 94.2 & 92.6 $\pm$ 1.4 \\
        & & 7 & 90.1 & 93.1 & 94.2 & 92.5 $\pm$ 2.1 \\
        \cmidrule{2-7}
        & \multirow{3}{*}{structured}
        & 3 & 93.0 & 93.2 & 92.0 & 92.8 $\pm$ 0.6 \\
        & & 5 & 90.5 & 91.2 & 94.3 & 92.0 $\pm$ 2.0 \\
        & & 7 & 90.2 & 92.8 & 94.0 & 92.3 $\pm$ 1.9 \\
        \bottomrule
    \end{tabular}
\end{table*}

\begin{table*}[ht]
    \centering
    \caption{
        Per-class accuracy on IMDB sentiment classification (\%), all prompting conditions combined (3 templates $\times$ \{0-shot, 3/5/7-shot $\times$ 3 seeds\} = 30 predictions per test example). Entries are exact-match / contains.
    }
    \label{tab:imdb_per_class}
    \small
    \setlength{\tabcolsep}{5pt}
    \renewcommand{\arraystretch}{1.15}
    \begin{tabular}{lc cc}
        \toprule
        \textbf{Sentiment} & \textbf{N}
        & \textbf{1B} & \textbf{4B} \\
        \midrule
        positive & 75000 & 74.8 / 74.8 & 86.6 / 86.6 \\
        negative & 75000 & 93.4 / 93.4 & 96.5 / 96.5 \\
        \midrule
        \multicolumn{2}{l}{\textit{Macro avg}}
        & 84.1 / 84.1 & 91.6 / 91.6 \\
        \bottomrule
    \end{tabular}
\end{table*}

\begin{table*}[ht]
    \centering
    \caption{
        Fine-tuning results on IMDB sentiment classification (binary: positive/negative).
    }
    \label{tab:imdb_finetune}
    \small
    \setlength{\tabcolsep}{5pt}
    \renewcommand{\arraystretch}{1.15}
    \begin{tabular}{llcc cc c}
        \toprule
        \textbf{Model} & \textbf{Tuning} & \textbf{Train Data} & \textbf{Hyperparams}
        & \textbf{Val Acc (\%)} & \textbf{Test Acc (\%)}
        & \textbf{Train (s)} \\
        \midrule
        1B & Full FT     & Real & lr=2e-5, bs=16$\times$2  & 95.08 & 94.84 & 7313 \\
        4B & LoRA (r=32, $\alpha$=64) & Real & lr=2e-5, bs=16$\times$2 & 97.14 & 96.98 & 6072 \\
        \midrule
        4B & LoRA (r=32, $\alpha$=64) & DP-synth ($\varepsilon$=10) & lr=2e-5, bs=16$\times$2 & 95.82 & 95.94 & 5567 \\
        4B & LoRA (r=32, $\alpha$=64) & Non-DP synth & lr=2e-5, bs=16$\times$2 & 96.46 & 96.38 & 6292 \\
        \bottomrule
    \end{tabular}
\end{table*}

\subsection{OpenReview Score Prediction}
\label{app:openreview_saturation}

OpenReview score prediction~\cite{pmlr-v235-xie24g} is a 5-way classification task over reviewer recommendation scores (1: strong reject, 3: reject, 5: marginally below threshold, 6: marginally above threshold, 8: accept), commonly used in the DP benchmark literature. Compared to Yelp Polarity and IMDB, OpenReview is somewhat less saturated, but the gap between prompting and fine-tuning remains small enough that the four training regimes still occupy a narrow band.

\paragraph{Setup.}
We evaluate Gemma~3 1B-IT and 4B-IT on a test set of 2,798 OpenReview papers~\cite{pmlr-v235-xie24g}. We consider three prompt templates (\textit{direct}, \textit{reasoning}, and \textit{structured}) in both 0-shot and 5-shot settings (3 random seeds for 5-shot). We then fine-tune each model on the training split (1B: full fine-tuning, lr=2e-5; 4B: LoRA $r{=}32$, $\alpha{=}64$, lr=2e-4; both for 3 epochs) and report validation and test accuracy.

\paragraph{Results.}
Table~\ref{tab:openreview_saturation} summarizes the saturation gap. The 4B model achieves 51.3\% accuracy under its best prompting condition (reasoning, 0-shot), while fine-tuning on the real training data improves this only to 61.9\%. This gap of just 10.6 percentage points means that the prompted model already reaches 83\% of the fine-tuned ceiling. For the 1B model, the gap is larger (19.0 points) but the fine-tuned accuracy itself is modest at 58.5\%.

We additionally fine-tune the 4B model on DP-synthetic data ($\varepsilon{=}10$, temp=1.0) and non-private synthetic data (temp=1.0), both generated from a LoRA-finetuned Gemma~3 1B-PT generator, for 2 epochs with the same LoRA configuration. The DP-synthetic model achieves 55.4\% and the non-private synthetic model achieves 57.9\%, confirming that all four training regimes (no training, DP-synth, synth, real) occupy a narrow 10.6-point band from 51.3\% to 61.9\%, as visualized in Figure~\ref{fig:saturation_united}.

\begin{table}[H]
    \centering
    \caption{
        Saturation summary on OpenReview score prediction (\%, exact-match). \textit{Best prompted} is the maximum over all (template, shot, seed) conditions. The small gap between prompting and fine-tuning, especially for the 4B model, indicates limited headroom for distinguishing synthesis methods.
    }
    \label{tab:openreview_saturation}
    \small
    \setlength{\tabcolsep}{5pt}
    \renewcommand{\arraystretch}{1.15}
    \begin{tabular}{lc ccc}
        \toprule
        \textbf{Model} & \textbf{Best Prompted}
        & \textbf{Fine-tuned} & \textbf{Gap (pts)} & \textbf{Prompted / FT} \\
        \midrule
        1B & 39.5 \scriptsize(structured 5-shot) & 58.5 & 19.0 & 0.68 \\
        4B & 51.3 \scriptsize(reasoning 0-shot)  & 61.9 & 10.6 & 0.83 \\
        \bottomrule
    \end{tabular}
\end{table}

\paragraph{Detailed results.}
Tables~\ref{tab:openreview_prompting}--\ref{tab:openreview_finetune} report the full breakdown. Table~\ref{tab:openreview_prompting} shows accuracy across all prompting conditions; Table~\ref{tab:openreview_5shot_seeds} reports per-seed variance for 5-shot; Table~\ref{tab:openreview_per_class} provides per-class accuracy; and Table~\ref{tab:openreview_finetune} reports fine-tuning results.

\begin{table*}[ht]
    \centering
    \caption{
        OpenReview score prediction accuracy by prompting condition (\%). Entries are exact-match / contains. 5-shot entries report mean across 3 seeds.
    }
    \label{tab:openreview_prompting}
    \small
    \setlength{\tabcolsep}{5pt}
    \renewcommand{\arraystretch}{1.15}
    \begin{tabular}{lllc cc}
        \toprule
        \textbf{Task} & \textbf{\# Test} & \textbf{Model} & \textbf{Template}
        & \textbf{0-shot} & \textbf{5-shot (mean)} \\
        \midrule

        \multirow{6}{*}{Score Prediction}
        & \multirow{6}{*}{2798}
        & \multirow{3}{*}{1B}
        & direct
        & 28.8 / 29.2
        & 30.9 / 30.9 \\
        & & & reasoning
        & 23.4 / 23.4
        & 36.2 / 36.2 \\
        & & & structured
        & 24.2 / 24.4
        & 39.5 / 39.6 \\
        \cmidrule{3-6}
        & & \multirow{3}{*}{4B}
        & direct
        & 47.3 / 47.3
        & 50.9 / 50.9 \\
        & & & reasoning
        & 51.3 / 56.8
        & 50.9 / 50.9 \\
        & & & structured
        & 51.1 / 51.1
        & 50.4 / 50.4 \\
        \bottomrule
    \end{tabular}
\end{table*}

\begin{table*}[ht]
    \centering
    \caption{
        OpenReview 5-shot accuracy by seed (\%). Entries are exact-match / contains. Std reported across seeds.
    }
    \label{tab:openreview_5shot_seeds}
    \small
    \setlength{\tabcolsep}{5pt}
    \renewcommand{\arraystretch}{1.15}
    \begin{tabular}{llc ccc c}
        \toprule
        \textbf{Model} & \textbf{Template}
        & \textbf{Seed 0} & \textbf{Seed 1} & \textbf{Seed 2}
        & \textbf{Mean $\pm$ Std} \\
        \midrule

        \multirow{3}{*}{1B}
        & direct
        & 26.7 / 26.7 & 33.0 / 33.0 & 32.9 / 32.9
        & 30.9 $\pm$ 2.9 / 30.9 $\pm$ 2.9 \\
        & reasoning
        & 32.7 / 32.8 & 42.3 / 42.3 & 33.5 / 33.5
        & 36.2 $\pm$ 4.3 / 36.2 $\pm$ 4.3 \\
        & structured
        & 39.6 / 39.6 & 38.7 / 38.8 & 40.2 / 40.2
        & 39.5 $\pm$ 0.6 / 39.6 $\pm$ 0.6 \\
        \midrule
        \multirow{3}{*}{4B}
        & direct
        & 51.8 / 51.8 & 47.3 / 47.3 & 53.7 / 53.7
        & 50.9 $\pm$ 2.7 / 50.9 $\pm$ 2.7 \\
        & reasoning
        & 52.4 / 52.4 & 48.3 / 48.4 & 52.0 / 52.0
        & 50.9 $\pm$ 1.8 / 50.9 $\pm$ 1.8 \\
        & structured
        & 52.6 / 52.6 & 47.0 / 47.0 & 51.6 / 51.6
        & 50.4 $\pm$ 2.4 / 50.4 $\pm$ 2.4 \\
        \bottomrule
    \end{tabular}
\end{table*}

\begin{table*}[ht]
    \centering
    \caption{
        Per-class accuracy on OpenReview score prediction (\%), all prompting conditions combined (3 templates $\times$ \{0-shot, 5-shot $\times$ 3 seeds\} = 12 predictions per test example). Entries are exact-match / contains.
    }
    \label{tab:openreview_per_class}
    \small
    \setlength{\tabcolsep}{5pt}
    \renewcommand{\arraystretch}{1.15}
    \begin{tabular}{llc cc}
        \toprule
        \textbf{Recommendation} & \textbf{Description} & \textbf{N}
        & \textbf{1B} & \textbf{4B} \\
        \midrule
        1 & strong reject                      &   204 &  2.9 /  2.9 & 20.1 / 20.6 \\
        3 & reject, not good enough            &  6648 & 38.9 / 38.9 & 51.2 / 52.0 \\
        5 & marginally below threshold         &  9120 & 11.2 / 11.3 & 68.6 / 68.6 \\
        6 & marginally above threshold         & 10752 & 46.0 / 46.1 & 43.9 / 44.5 \\
        8 & accept, good paper                 &  6852 & 36.9 / 36.9 & 37.2 / 37.7 \\
        \midrule
        \multicolumn{3}{l}{\textit{Macro avg (unweighted)}}
        & 27.2 / 27.2 & 44.2 / 44.7 \\
        \bottomrule
    \end{tabular}
\end{table*}

\begin{table*}[ht]
    \centering
    \caption{
        Fine-tuning results on OpenReview score prediction (5-way classification over recommendation scores 1, 3, 5, 6, 8). Real-data models trained for 3 epochs; synthetic-data models trained for 2 epochs.
    }
    \label{tab:openreview_finetune}
    \small
    \setlength{\tabcolsep}{5pt}
    \renewcommand{\arraystretch}{1.15}
    \begin{tabular}{llcc cc c}
        \toprule
        \textbf{Model} & \textbf{Tuning} & \textbf{Train Data} & \textbf{Hyperparams}
        & \textbf{Val Acc (\%)} & \textbf{Test Acc (\%)}
        & \textbf{Train (s)} \\
        \midrule
        1B & Full FT     & Real & lr=2e-5,  bs=16$\times$2  & 58.97 & 58.47 & 1899 \\
        4B & LoRA (r=32, $\alpha$=64) & Real & lr=2e-4, bs=8$\times$4 & 61.12 & 61.94 & 4499 \\
        \midrule
        4B & LoRA (r=32, $\alpha$=64) & DP-synth ($\varepsilon$=10) & lr=2e-5, bs=16$\times$2 & 58.29 & 55.43 & 2966 \\
        4B & LoRA (r=32, $\alpha$=64) & Non-DP synth & lr=2e-5, bs=16$\times$2 & 59.76 & 57.86 & 3115 \\
        \bottomrule
    \end{tabular}
\end{table*}

\subsection{Yelp Polarity}
\label{app:yelp_saturation}

Yelp Polarity~\cite{zhangCharacterlevelConvolutionalNetworks2015} is a binary sentiment classification task widely used in the DP benchmark literature. Headline accuracies across the four training regimes are summarized in Table~\ref{tab:yelp_saturation}. The full range from no training to training on the real corpus spans only 3.0 percentage points, with DP-synthetic and non-private synthetic data occupying a band of just 0.4 points between them. As with IMDB, this leaves essentially no resolution to distinguish DP synthesis methods.

\begin{table}[ht]
    \centering
    \caption{Yelp Polarity accuracy (\%) across the four training regimes shown in Figure~\ref{fig:saturation_united}.}
    \label{tab:yelp_saturation}
    \small
    \setlength{\tabcolsep}{10pt}
    \renewcommand{\arraystretch}{1.15}
    \begin{tabular}{cccc}
        \toprule
        \textbf{No training} & \textbf{DP-synth} & \textbf{Non-private synth} & \textbf{Train on real} \\
        \midrule
        95.5 & 97.8 & 98.2 & 98.5 \\
        \bottomrule
    \end{tabular}
\end{table}

\section{\geminon: Curation and Evaluation Details}
\label{app:geminon_details}

\geminon is a fully controlled fictional domain designed to (i) make freshness unambiguous, (ii) support the systematic study of DP learnability, and (iii) reduce evaluation noise through grounded, short-answer QA with deterministic normalization. The full curation code is available at \url{https://github.com/plau666/ContinuousBenchCuration}.

To construct a fictional world that is statistically realistic yet entirely novel, we derive structural priors from publicly available Pok\'{e}mon metadata \cite{maca11_all_pokemon_dataset,amancio_pokemon_moves_2024}. These priors are used only to parameterize the generative process. We emphasize that all resulting Geminon entities, names, and stats are newly created, while types, moves, and abilities are inherited from the reference data.

\subsection{Index Data Curation}
\label{subsec:geminon-index}

\subsubsection{Evolution Line and Types}
In the original Pok\'{e}mon data, there are 160 three-stage evolution lines, 239 two-stage evolution lines, and 120 single-stage lines. We generate 100 three-stage evolution lines (300 Geminon), 100 two-stage evolution lines (200 Geminon), and 100 single-stage lines (100 Geminon), for a total of 600 entities. All randomness is seeded for reproducibility.

Each evolution line is assigned a primary type (\texttt{type1}) sampled uniformly from the 18 canonical Pok\'{e}mon types \cite{maca11_all_pokemon_dataset,amancio_pokemon_moves_2024}. This primary type is held constant across all stages of the line. A secondary type (\texttt{type2}) is then assigned independently at each stage according to a two-case rule. If the preceding stage has no \texttt{type2}, a new secondary type is introduced with stage-dependent probability: for a three-stage line, the probabilities are $[0.2,\,0.4,\,0.8]$ across stages 1--3; for a two-stage line, $[0.2,\,0.8]$; and for a single-stage line, $[0.5]$. If the preceding stage already has a \texttt{type2}, it is retained with probability 0.8 and replaced by a different secondary type with probability 0.2. This rule is applied uniformly across all line lengths and makes secondary typing more likely in later-stage, typically stronger forms while still permitting occasional type diversification.

\subsubsection{Statistics}
\paragraph{Reference Stats.}
For each of the 998 reference Pok\'{e}mon \cite{maca11_all_pokemon_dataset,amancio_pokemon_moves_2024}, we record nine attributes: six battle stats (\texttt{HP}, \texttt{Attack}, \texttt{Defense}, \texttt{Special Attack}, \texttt{Special Defense}, and \texttt{Speed}), \texttt{Base Stat Total} (\texttt{BST}), \texttt{height}, and \texttt{weight}. For each attribute $a$, we compute the empirical mean $\mu_a$ and standard deviation $\sigma_a$ over all stage-1 reference Pok\'{e}mon.

For multi-stage evolution lines, we also compute per-attribute multiplicative ratios: \emph{stage-2/stage-1} from 359 pairs and \emph{stage-3/stage-1} from 120 pairs. For each attribute $a$ and each ratio type, we record the empirical mean $\mu_a^r$ and standard deviation $\sigma_a^r$ for $r \in \{2/1, 3/1\}$.

\paragraph{Stage-1 Statistics Sampling.}
For each attribute $a$, we construct a discrete grid $\mathcal{G}_a$ of integer-valued points spanning $[\min_a,\, \max_a)$ with step size 1, where $\min_a$ and $\max_a$ are taken over all real stage-1 Pok\'{e}mon. Battle stats are sampled from this grid using a \emph{discrete Gaussian($\mu_a, \sigma_a$)}: each grid point $x \in \mathcal{G}_a$ is sampled with probability $\propto \exp(-\nicefrac{(x-\mu_a)^2}{2\sigma_a^2})$, where $\mu_a$ and $\sigma_a$ are estimated from stage-1 reference Pok\'{e}mon. \texttt{Height} and \texttt{weight} are sampled using a \emph{discrete exponential($1/\mu_a$)}, where each grid point is sampled with probability $\propto \exp(-x/\mu_a)$. The stage-1 \texttt{BST} is then computed as the \emph{sum of the six sampled battle stats.}

\paragraph{Later-stage Statistics Sampling.}
For each attribute $a$ and ratio type $r \in \{2/1,\, 3/1\}$, we construct a \emph{restricted ratio grid} $\mathcal{G}_a^r$ to confine samples to a tight, empirically grounded range. Specifically, the empirical standard deviation is clipped to $\bar{\sigma}_a = \mathrm{clip}(\sigma_a^r,\, 0.05,\, 0.1)$ and the empirical mean is clipped to $\bar{\mu}_a = \mathrm{clip}(\mu_a^r,\, 0.05,\, c_{\max})$, where $c_{\max} = 1.6$ for the 2/1 ratio and $c_{\max} = 2.0$ for the 3/1 ratio. The restricted grid is $[\bar{\mu}_a-\bar{\sigma}_a, \bar{\mu}_a+\bar{\sigma}_a]$
with step size 0.1, with boundaries rounded outward to the nearest 0.01.

Stage-2 and stage-3 attribute values are computed as
\begin{align*}
    s_a^{(2)} &= \mathrm{round}\!\left(s_a^{(1)} \times r_a\right), \quad r_a \sim \mathrm{Uniform}(\mathcal{G}_a^{2/1}), \\
    s_a^{(3)} &= \mathrm{round}\!\left(s_a^{(1)} \times r_a\right), \quad r_a \sim \mathrm{Uniform}(\mathcal{G}_a^{3/1}),
\end{align*}
where $\mathrm{Uniform}(\mathcal{G})$ denotes a uniform draw from the discrete grid $\mathcal{G}$. Notably, stage 3 is derived \emph{directly from stage 1}, rather than from stage 2, using the stage-3/stage-1 ratio grid. This avoids compounding variance across stages. The multiplicative scheme is applied to all six battle stats as well as to \texttt{height} and \texttt{weight}. For each later stage, \texttt{BST} is recomputed as the sum of the six battle stats.

\subsubsection{Moves and Abilities}
We additionally load reference tables of moves and abilities to define realistic candidate pools \cite{maca11_all_pokemon_dataset,amancio_pokemon_moves_2024}. Move records contain a \texttt{name}, \texttt{type}, and short description. Ability and move names are deduplicated and filtered to concise surface forms of at most two words. These pools are used only to sample type-consistent attributes for \geminon.

For each Geminon, we sample one ability uniformly from the reference ability pool and one signature move, together with its description, uniformly from the set of moves matching either of its types \cite{maca11_all_pokemon_dataset,amancio_pokemon_moves_2024}.

\subsubsection{Indexing and Schema}
After generation, evolution lines are shuffled and assigned contiguous integer id starting at 10000, ensuring that members of the same line occupy adjacent indices. Each \geminon record contains:
\begin{jsonblock}{\geminon Index Schema}{
  "name": <string>,
  "classification": <string>,
  "type1": <string>,
  "type2": <string or null>,
  "ability": <string>,
  "hp": <int>, "attack": <int>, "defense": <int>,
  "special attack": <int>, "special defense": <int>,
  "speed": <int>, "base_stat_total": <int>,
  "weight": <int>, "height": <int>,
  "evolution_line": [<string>, ...],
  "move": {"name": <string>,
           "short_description": <string>},
  "idx": <int>
}
\end{jsonblock}
Names and classifications are generated subsequently using \texttt{gemini-2.5-flash} \cite{comanici2025gemini25pushingfrontier}. A deduplication of names and classifications is conducted. Specifically, we require (1) all 600 individual Geminon names must be globally unique, and (2) for evolution lines with more than two stages, no two lines may share the same tuple of classifications across their stages. Single-stage lines are exempt from (2). Re-queries to \texttt{gemini-2.5-flash} are issued with a forbidden-names or forbidden-tuple constraint until both conditions are met.

\begin{figure}[ht]
    \centering
    \includegraphics[height=.8\textheight,keepaspectratio]{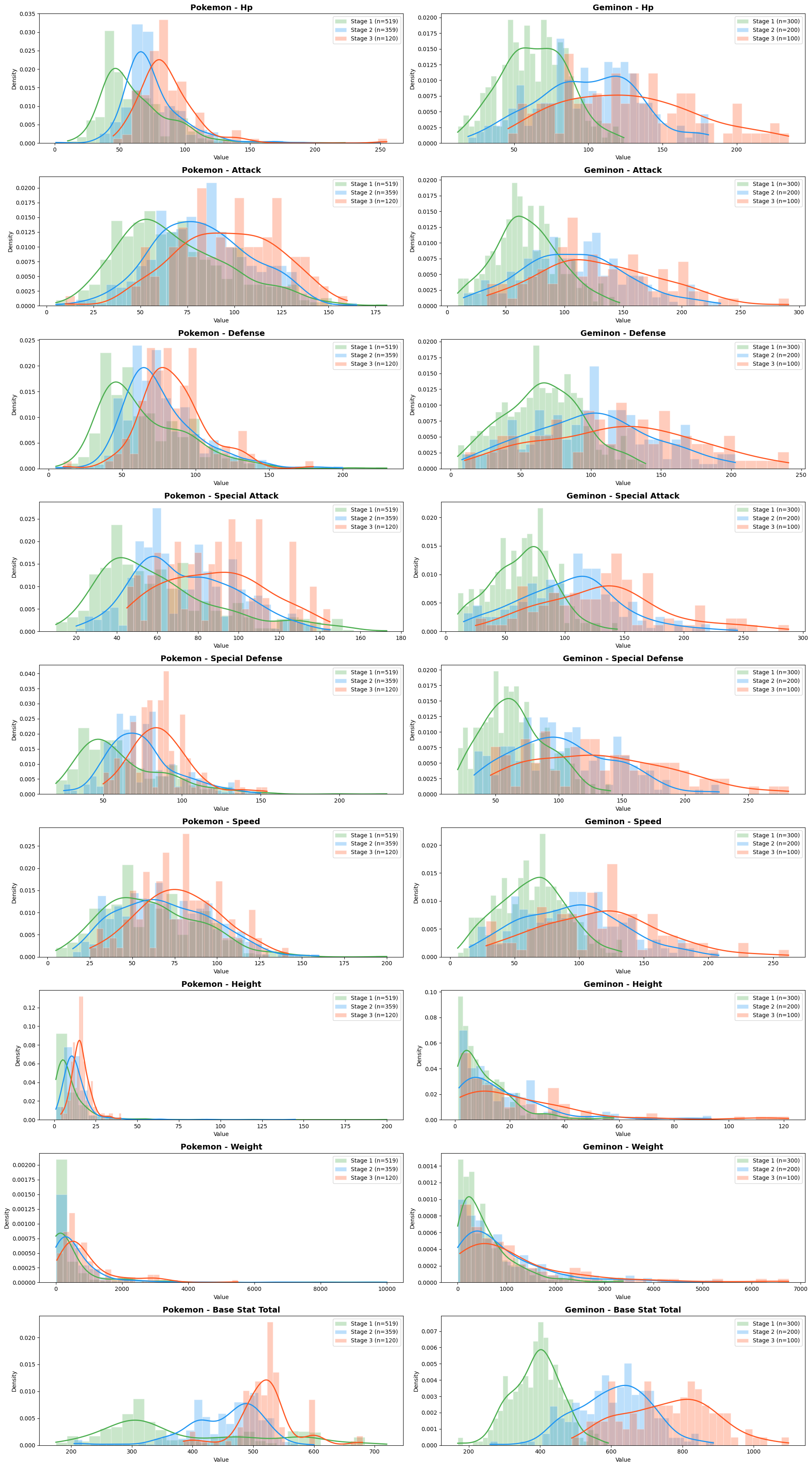}
    \caption{Comparison of Pok\'{e}mon and Geminon stat distributions.}
    \label{fig:pokemon_geminon_index}
\end{figure}

\paragraph{Examples and Distribution Similarity}
We verify that the generated stage-conditional distributions closely match the corresponding reference Pok\'{e}mon statistics, including the skewness of physical attributes and the monotonic increase of battle stats across evolutionary stages. A comparison of the resulting distributions is shown in Figure~\ref{fig:pokemon_geminon_index}, and example \geminon index entries are shown in Figure~\ref{fig:geminon-examples}.

\begin{figure}[htbp]
\centering
\begin{adjustbox}{max totalsize={\textwidth}{0.9\textheight},center}
\begin{minipage}{\textwidth}
\vspace{0.1cm}
\textbf{Single-stage evolution example}\par\medskip
\begin{tcbraster}[
  raster columns=1,
  raster column skip=2mm,
  raster equal height,
  raster left skip=0pt,
  raster right skip=0pt,
  width=\textwidth
]
\begin{jsonblock}{Stage 1: Caelumin}
{
  "name": "Caelumin",
  "classification": "Beacon Geminon",
  "type1": "flying",
  "type2": null,
  "ability": "Illuminate",
  "hp": 53,
  "attack": 65,
  "defense": 58,
  "special attack": 77,
  "special defense": 99,
  "speed": 49,
  "base_stat_total": 401,
  "weight": 867,
  "height": 9,
  "idx": 10002,
  "evolution_line": ["Caelumin"],
  "move": {
    "name": "Sky Attack",
    "short_description": "User charges for one turn before attacking."
  }
}
\end{jsonblock}
\end{tcbraster}

\textbf{Two-stage evolution example}\par\medskip
\begin{tcbraster}[
  raster columns=2,
  raster column skip=2mm,
  raster equal height,
  raster left skip=0pt,
  raster right skip=0pt,
  width=\textwidth
]
\begin{jsonblock}{Stage 1: Phantatch}
{
  "name": "Phantatch",
  "classification": "Spore Geminon",
  "type1": "ghost",
  "type2": null,
  "ability": "Suction Cups",
  "hp": 79,
  "attack": 48,
  "defense": 56,
  "special attack": 23,
  "special defense": 88,
  "speed": 84,
  "base_stat_total": 378,
  "weight": 198,
  "height": 14,
  "idx": 10000,
  "evolution_line": ["Phantatch", "Phantrain"],
  "move": {
    "name": "Poltergeist",
    "short_description": "Inflicts regular damage with no additional effect."
  }
}
\end{jsonblock}
\begin{jsonblock}{Stage 2: Phantrain}
{
  "name": "Phantrain",
  "classification": "Parasite Geminon",
  "type1": "ghost",
  "type2": "dark",
  "ability": "Drought",
  "hp": 115,
  "attack": 72,
  "defense": 87,
  "special attack": 39,
  "special defense": 147,
  "speed": 134,
  "base_stat_total": 594,
  "weight": 337,
  "height": 22,
  "idx": 10001,
  "evolution_line": ["Phantatch", "Phantrain"],
  "move": {
    "name": "Jaw Lock",
    "short_description": "Seeds the target after inflicting damage."
  }
}
\end{jsonblock}
\end{tcbraster}

\bigskip

\textbf{Three-stage evolution example}\par\medskip
\begin{tcbraster}[
  raster columns=3,
  raster column skip=2mm,
  raster equal height,
  raster left skip=0pt,
  raster right skip=0pt,
  width=\textwidth
]
\begin{jsonblock}{Stage 1: Boreling}
{
  "name": "Boreling",
  "classification": "Frost Geminon",
  "type1": "ice",
  "type2": null,
  "ability": "Berserk",
  "hp": 69,
  "attack": 60,
  "defense": 63,
  "special attack": 67,
  "special defense": 68,
  "speed": 40,
  "base_stat_total": 367,
  "weight": 52,
  "height": 12,
  "idx": 10003,
  "evolution_line": ["Boreling", "Borelash", "Borastat"],
  "move": {
    "name": "Powder Snow",
    "short_description": "Has a chance to freeze the target."
  }
}
\end{jsonblock}
\begin{jsonblock}{Stage 2: Borelash}
{
  "name": "Borelash",
  "classification": "Shard Geminon",
  "type1": "ice",
  "type2": null,
  "ability": "Super Luck",
  "hp": 115,
  "attack": 89,
  "defense": 91,
  "special attack": 100,
  "special defense": 100,
  "speed": 64,
  "base_stat_total": 559,
  "weight": 83,
  "height": 20,
  "idx": 10004,
  "evolution_line": ["Boreling", "Borelash", "Borastat"],
  "move": {
    "name": "Ice Burn",
    "short_description": "Requires a turn to charge before attacking."
  }
}
\end{jsonblock}
\begin{jsonblock}{Stage 3: Borastat}
{
  "name": "Borastat",
  "classification": "Aurora Geminon",
  "type1": "ice",
  "type2": "electric",
  "ability": "Sweet Veil",
  "hp": 140,
  "attack": 125,
  "defense": 122,
  "special attack": 141,
  "special defense": 129,
  "speed": 71,
  "base_stat_total": 728,
  "weight": 104,
  "height": 24,
  "idx": 10005,
  "evolution_line": ["Boreling", "Borelash", "Borastat"],
  "move": {
    "name": "Icy Wind",
    "short_description": "Has a chance to lower the target's Speed."
  }
}
\end{jsonblock}
\end{tcbraster}
\end{minipage}
\end{adjustbox}
\caption{Example \geminon index entries.}
\label{fig:geminon-examples}
\end{figure}

\paragraph{Singleton and High-repetition Split}
We randomly sample 20 evolution lines from each of the three categories, three-stage, two-stage, and single-stage lines, to form the singleton split. This yields a total of 120 singleton Geminon. The remaining 180 evolution lines, comprising 480 Geminon, constitute the high-repetition split.

\subsection{Corpus Curation}
\label{subsec:geminon-corpus}
Given the \geminon index, we construct a corpus as a mixture of in-world document formats designed to distribute each atomic fact (the stats and attributes of Geminon) across many records while avoiding one-to-one paraphrasing. Specifically, we synthesize four document types: \emph{wiki entries} (encyclopedic third-person descriptions), \emph{field journal notes} (informal first-person trainer observations), \emph{evolution logs} (narrative analyses of an entire evolution line), and \emph{pairwise comparisons} (side-by-side trainer assessments of two Geminon).

Each prompt instructs the LLM to return, alongside every article, a provenance list specifying the exact Geminon attribute keys referenced in that article. These lists serve as a proxy for attribute-level mention frequency. LLM outputs are parsed by stripping Markdown fences and decoding the resulting JSON list. The raw provenance strings produced by the LLM are then normalized to canonical feature names through a combination of prefix stripping, an explicit mapping covering approximately 70 known variants and typos, and a final underscore-to-space fallback. Strings that cannot be resolved are discarded.

Articles are generated using \texttt{gemini-2.5-pro} with the prompts shown below. For the wiki, journal, and evolution prompts, each prompt is queried 100 times, and each response yields 10 articles, producing approximately 1000 candidate articles per Geminon or evolution line. Comparison prompts are run once over all $\binom{480}{2} = 114960$ unique pairs of public Geminon, with each response yielding 3 articles. Evolution prompts are applied to all 160 multi-stage public evolution lines (80 three-stage and 80 two-stage). After parsing, the final corpus contains 1408369 articles on Geminons in the high-repetition split and 116739 articles on Geminons in the singleton split, with the following breakdown:

\begin{table}[ht]
  \caption{Breakdown stats for Geminon corpus.}
  \label{tab:geminon-corpus-breakdown}
  \centering
  \begin{tabular}{lr}
    \toprule
    \textbf{Document type} & \textbf{Articles} \\
    \midrule
    Pairwise comparison (high-repetition) & 341,490 \\
    Field journal (high-repetition)       & 452,188 \\
    Evolution log (high-repetition)       & 147,271 \\
    Wiki (high-repetition)        & 467,420 \\
    Wiki (singleton)     & 116,739 \\
    \midrule
    \textbf{Total on high-repetition split}    & \textbf{1,408,369} \\
    \textbf{Total on singleton split} & \textbf{116,739} \\
    \bottomrule
  \end{tabular}
\end{table}

Example articles are shown in Figure~\ref{fig:geminon-articles-examples}.

\begin{figure}[ht]
\centering
\begin{adjustbox}{max width=\textwidth,center}
\begin{minipage}{\textwidth}
\begin{tcbraster}[
  raster columns=2,
  raster column skip=2mm,
  raster equal height,
  raster left skip=0pt,
  raster right skip=0pt,
  width=\textwidth
]
\begin{jsonblock}{wiki}
{
  "text": "As the first stage in the evolutionary line that culminates in Borastat, Boreling is a quintessential Frost Geminon. This pure ice-type, known as Gemidex entry #10003, measures 12 meters in height and weighs 52 lbs.  Its passive ability is Berserk, and it can be taught the move Powder Snow, which has a percentage-based chance to freeze an opponent. Boreling's base stat total of 367 is comprised of 69 HP, 60 attack, 63 defense, 67 special attack, 68 special defense, and a speed rating of 40.",
  "tag": [
    {
      "idx": 10003,
      "info": ["idx", "name", "classification", "type1", "ability", "evolution_line", "height", "weight", "base_stat_total", "hp", "attack", "defense", "special attack", "special defense", "speed", "move.name", "move.short_description"]
    }
  ],
  "type": "wiki"
}
\end{jsonblock}
\begin{jsonblock}{journal}
{
  "text": "I need to be strategic with my new Boreling. Its Berserk ability only activates when its HP is low, which means I need to know it can survive a hit. Its 63 defense and 68 special defense give me some confidence there. Since its speed is a low 40, I'll probably have to take a hit before I can attack, so those defensive stats are crucial.",
  "tag": [
    {
      "idx": 10003,
      "info": ["name", "ability", "hp", "defense", "special defense", "speed"]
    }
  ],
  "type": "journal"
}
\end{jsonblock}
\begin{jsonblock}{comparison}
{
  "text": "Looking at their combat stats, Aerofir is clearly built for speed, boasting a speed stat of 95, while Boreling is sluggish at 40. Despite this speed gap, their core defensive and offensive stats are surprisingly close, with both having attack and defense in the 60s. Aerofir has a better HP pool at 81, making it a bit more durable than Boreling at 69 HP. Overall, Aerofir has the advantage with a higher base stat total, but it's not a total blowout in every category.",
  "tag": [
    {
      "idx": 10003,
      "info": ["hp", "attack", "defense", "speed", "base_stat_total"]
    },
    {
      "idx": 10503,
      "info": ["hp", "attack", "defense", "speed", "base_stat_total"]
    }
  ],
  "type": "comparison"
}
\end{jsonblock}
\begin{jsonblock}{evolution log}
{
  "text": "While its special attack becomes its greatest asset, the attack stat also sees consistent growth. Boreling's initial attack of 60 is a solid starting point. This grows to a more threatening 89 in its Borelash stage. As Borastat, its physical prowess culminates in an impressive 125 attack, making it a versatile threat that can strike effectively from both the physical and special sides.",
  "tag": [
    {
      "idx": 10003,
      "info": ["attack"]
    },
    {
      "idx": 10004,
      "info": ["attack"]
    },
    {
      "idx": 10005,
      "info": ["attack"]
    }
  ],
  "type": "evolution"
}
\end{jsonblock}
\end{tcbraster}
\end{minipage}
\end{adjustbox}
\caption{Example articles about Geminon Boreling.}
\label{fig:geminon-articles-examples}
\end{figure}

\clearpage

\subsection{QA Curation.}
\label{subsec:geminon-qa-curation}

For each Geminon in the high-repetition and singleton splits, we construct simple questions about its features using the following template:

\begin{jsonblock}{Geminon Question Template}
"What is the classification of {name}?"
"What are the types of {name}?"
"What is the ability of {name}?"
"What is the HP stat of {name}?"
"What is the attack stat of {name}?"
"What is the defense stat of {name}?"
"What is the special attack stat of {name}?"
"What is the special defense stat of {name}?"
"What is the speed stat of {name}?"
"What is the base stat total stat of {name}?"
"What is the move of {name}?"
"What is the weight (in lbs) of {name}?"
"What is the height (in meters) of {name}?"
\end{jsonblock}
Example QAs for Geminon Boreling is presnted below.
\begin{jsonblock}{Example QAs for Geminon Boreling}
{
    "question": "What is the classification of Boreling?", 
    "answer": "Frost Geminon"
    "supports": [510, 3000, 5960, 7410, 8921, 8929, ...] # supporting articles idxs 
}
{
    "question": "What are the types of Boreling?", 
    "answer": "ice"
    "supports": [3001, 7411, 8922, 8929, 10425, 11915, ...] # supporting articles idxs 
}
...
\end{jsonblock}
The two splits' QAs are kept separate. Within each split, we partition every Geminon's QAs into validation and test sets of roughly equal size: the validation set is used for checkpoint selection and the test set for final performance reporting.

\subsection{Subsampling}
From the full corpus, denoted \textsc{Geminon-Large}, we construct two smaller subsets, \textsc{Geminon-Small} and \textsc{Geminon-Medium}, via a coverage-aware weighted sampling procedure.

For the high-repetition split, in each round, each candidate article is assigned a score of $\sum_{(\text{Geminon},\,\text{feature})} 1/(1 + \texttt{count})$, where \texttt{count} tracks how many times that \((\text{Geminon}, \text{feature})\) pair has already been selected. Articles that fill the largest current coverage gaps therefore receive the highest sampling probability. The resulting source breakdown is as follows:
\begin{itemize}
  \item \textsc{Geminon-Small:} comparison 100k, evolution 40k, journal 40k, wiki 20k
  \item \textsc{Geminon-Medium:} comparison 300k, evolution 100k, journal 300k, wiki 300k
\end{itemize}

For the singleton set, we select one wiki article per Geminon in the singleton split by greedily choosing the candidate with the highest coverage of that Geminon's applicable canonical features. Under this criterion, all 120 singleton Geminon achieve perfect feature coverage.

Appending these 120 singleton wiki articles to each subset yields final sizes of 200{,}120 and 1{,}000{,}120, respectively. Table~\ref{tab:geminon-stats-coverage} reports attribute-mention statistics over the full corpus (\textsc{Geminon-Large}).

\begin{table}[ht]
\centering
\caption{Left: per-Geminon mention counts for each feature, aggregated over all articles for the 480 Geminons in the high-repetition split; Mean/Std/Min/Max are computed over per-Geminon counts. Right: feature coverage in the 120 selected wiki articles, each corresponding a Geminon in the singleton split. \texttt{type2} covers only 80/120 cases because the remaining 40 Geminon are single-typed.}
\label{tab:geminon-stats-coverage}

\begin{subtable}[ht]{0.58\textwidth}
\centering
\small
\begin{tabular}{lrrrr}
\toprule
\textbf{Feature} & \textbf{Mean} & \textbf{Std} & \textbf{Min} & \textbf{Max} \\
\midrule
name                    & 2260.7 & 146.8 & 1778 & 2717 \\
classification          & 1721.2 &  67.1 & 1523 & 1908 \\
type1                   & 1877.8 &  81.9 & 1665 & 2133 \\
type2                   & 1298.3 & 630.2 &  390 & 2190 \\
ability                 & 1763.7 &  90.0 & 1550 & 2205 \\
hp                      & 1586.0 & 160.0 & 1139 & 2272 \\
attack                  & 1716.1 & 267.7 & 1002 & 2374 \\
defense                 & 1707.4 & 212.8 & 1216 & 2380 \\
speed                   & 1758.8 & 132.9 & 1448 & 2263 \\
special attack          & 1543.5 & 292.6 & 1014 & 2268 \\
special defense         & 1512.4 & 220.3 & 1051 & 2161 \\
base\_stat\_total       & 1727.2 &  77.6 & 1480 & 1880 \\
weight                  & 1649.6 &  47.3 & 1467 & 2049 \\
height                  & 1641.2 &  42.4 & 1462 & 1721 \\
move.name               &  917.1 & 103.6 &  527 & 1178 \\
\bottomrule
\end{tabular}
\label{tab:geminon-feature-stats}
\end{subtable}
\hfill
\begin{subtable}[t]{0.36\textwidth}
\centering
\small
\begin{tabular}{lr}
\toprule
\textbf{Feature} & \textbf{Coverage} \\
\midrule
name                    & 120/120 \\
classification          & 120/120 \\
type1                   & 120/120 \\
type2                   &  80/120 \\
ability                 & 120/120 \\
hp                      & 120/120 \\
attack                  & 120/120 \\
defense                 & 120/120 \\
special attack          & 120/120 \\
special defense         & 120/120 \\
speed                   & 120/120 \\
base\_stat\_total       & 120/120 \\
weight                  & 120/120 \\
height                  & 120/120 \\
move.name               & 120/120 \\
\bottomrule
\end{tabular}
\label{tab:feature-coverage}
\end{subtable}
\end{table}

The QA pairs are identical across all three corpus sizes, with different support sets, since each subset contains a different pool of articles. On average, a QA has roughly 200 supporting articles in \textsc{Geminon-Small} and roughly 1{,}100 in \textsc{Geminon-Medium}.

\section{\news: Curation and Evaluation Details}
\label{app:news_details}
\subsection{Corpus Curation}
We leveraged the CommonCrawl News \cite{ccnews}, specifically the September 2025 dump for the experiments presented. We only select the English documents and got 2,876,319 documents in total.
For each article, we retrieved the publication date, document URL, main text, language, crawl date, hostname, and title using \texttt{trafilatura}. Code for dataset curation can be found at \url{https://github.com/plau666/ContinuousBenchCuration}.

We normalize the paragraphs by collapsing extra spaces and newlines, producing 1,768,567 cleaned articles.
Then, we deduplicate the cleaned-up articles in two passes.
Pass~1 removes exact duplicates globally via SHA-256 hashing of article text.
Pass~2 performs near-deduplication using MinHash LSH
(\texttt{datasketch}) with 128 permutations, word 5-gram shingles,
and a containment similarity threshold of 0.80, where
containment $= |A \cap B| / \min(|A|, |B|)$. When a near-duplicate pair is found, the shorter article is removed. Some example articles are presented below.

\begin{figure}[ht]
\centering
\begin{adjustbox}{max totalsize={\textwidth}{1\textheight},center}
\begin{minipage}{\textwidth}
\begin{tcbraster}[
  raster columns=1,
  raster column skip=2mm,
  raster left skip=0pt,
  raster right skip=0pt,
  width=\textwidth
]
\begin{jsonblock}{Example article 1 about Asia Cup 2025: India vs Pakistan}
{
    "url": "https://newsable.asianetnews.com/sports/cricket-india-vs-pakistan-asia-cup-2025-referee-pycroft-pcb-handshake-row-articleshow-miqavqv",
    "hostname": "newsable.asianetnews.com",
    "title": "India vs Pakistan, Asia Cup 2025: Referee Pycroft Faces PCB Heat As Handshake Controversy Escalates",
    "date": "2025-09-15",
    "crawl_date": "2025-09-15T05:55:11Z",
    "language": "en",
    "text": "India\u2019s win over Pakistan in the Asia Cup was overshadowed by a handshake row. Match referee Andy Pycroft faces heat as PCB protests India\u2019s refusal to shake hands, citing government policy and solidarity with Pahalgam attack victims. Dubai [UAE]: The highly anticipated India-Pakistan Asia Cup clash in Dubai on Sunday has spiraled into controversy, with match referee Andy Pycroft caught in the middle of a row over the absence of customary handshakes. India clinched a comfortable seven-wicket win, but it was events off the field that have dominated headlines..."
}
\end{jsonblock}
\begin{jsonblock}{Example article 2 about Asia Cup 2025: India vs Pakistan}
{
    "url": "https://www.hindustantimes.com/cricket/pakistan-knock-on-iccs-door-demands-immediate-removal-of-match-referee-for-staying-silent-on-indias-no-handshake-101757928011929.html",
    "hostname": "www.hindustantimes.com",
    "title": "Pakistan knocks on ICC's door, demands immediate removal of match referee for staying silent on India's no handshake",
    "date": "2025-09-15",
    "crawl_date": "2025-09-15T09:31:13Z",
    "language": "en",
    "text": "Pakistan knocks on ICC's door, demands immediate removal of match referee for staying silent on India's no handshake The PCB has escalated India's no-handshake controversy to the ICC, asking for the immediate removal of match referee Andy Pycroft. The no-handshake episode between India and Pakistan at the end of Match 6 of the Asia Cup 2025 is snowballing into a huge controversy. After the Pakistan cricket team complained to the match referee about India's actions, the board has now reached out to the ICC, demanding that the official be removed. PCB's strict measures come in the wake of the disappointment stemming from the fact that Indian players did not participate in the post-match customary handshakes with their Pakistan counterparts. The move has triggered former Pakistan cricketers and members of the board alike, and facing the wrath is Andy Pycroft, a veteran in this role..."
}
\end{jsonblock}
\end{tcbraster}
\end{minipage}
\end{adjustbox}
\caption{Example articles about Asia Cup 2025: India vs Pakistan}
\label{fig:news-article-examples}
\end{figure}

\subsection{QA Curation}
We construct question-answer pairs from the deduplicated cleaned-up corpus through a multi-stage pipeline using the Gemini API.
\paragraph{Clustering.}
We embedded the extracted articles using \texttt{embeddinggemma-300m} \cite{embeddinggemma}, with prompt prefix \texttt{"task: clustering | query: "}, input mode title and first paragraph ($\sim$500 characters), maximum sequence length 512, and L2-normalized \texttt{float16} outputs. We then clustered the 1,768,567 embeddings with a local windowed kNN + Leiden approach: for each 7-day sliding window ($\pm$3 days), we build a mutual kNN graph (k-search=60, k-graph=30, similarity threshold 0.55) and run Leiden community detection (resolution 0.07, 4 iterations), with a minimum cluster size of 25 and cross-window merging (Jaccard threshold 0.35, centroid similarity 0.9). This step produced 13,870 clusters containing 465,966 unique articles.

\paragraph{Fact extraction.}
For each of the top 500 largest clusters, we send up to 50 articles (each truncated to 512 characters) to \texttt{gemini-2.5-pro} and extract 12--25 short, QA-ready facts. Each fact must be explicitly stated in at least three distinct articles and anchored by an absolute date, a named entity, or a named document or action. This yielded 5,947 facts across 431 clusters. It is worth noting that the 69 clusters left are mostly spam clusters.

\paragraph{QA generation.}
We then send the extracted facts, together with up to 50 underlying cluster articles (each truncated to 512 characters), back to \texttt{gemini-2.5-pro} to generate up to 12 short-answer QA pairs per cluster. Questions must be answerable with a name, date, number with unit, location, or outcome, and should include an absolute timestamp to make them unambiguous. We require that the same answer string appear verbatim in at least three articles. This process yields 4,468 QA pairs across 430 clusters.

In manual inspection, this two-stage pipeline produced substantially higher-quality QA pairs than direct single-stage question generation.

\paragraph{Closed-book Answerability and Ambiguity check.}
Next, we prompt \texttt{gemini-2.5-pro} with each question in a zero-shot setting, i.e., without any article context. Then use the same model as a judge to determine 1. whether the zero-shot response matches the ground-truth answer and 2. whether each question is fully interpretable on its own (standalone-ambiguity check.) We emphasize that, although both the source articles and the resulting knowledge tests are post-cutoff, models may still answer some questions correctly by exploiting prior associations or plausible inference rather than genuine knowledge of the underlying articles. For example, for a question such as ``Which company hosted a party on September 15, 2025, in celebration of the 2025 Emmy Awards?'', a model may correctly guess ``Netflix'' based on historical associations rather than direct knowledge of the covered event. This yields 2,604 out of 4,468 QAs that are both unambiguous and unanswerable (39.7\% zero-shot correct and 3.0\% ambiguous).

\paragraph{Open-book Answerability check (Support identification.)} 
We embed every article and every surviving QA, and for each QA retrieve its top-1000 candidate articles by embedding similarity. Grouping these candidates by article, we prompt \texttt{gemini-2.5-flash-lite} in open-book fashion, i.e. prompt with the article in context, and judge each response for correctness with \texttt{gemini-2.5-flash-lite}. An article is counted as \emph{supporting} a QA if its open-book answer is judged correct.

Each QA is provided with its closed-book answer and the corresponding LLM-judge correctness label, together with open-book answers for its $\approx$1{,}000 candidate articles, each carrying its own LLM-judge correctness label. By construction, every closed-book answer is judged incorrect, due to the filtering criterion, but we retain these (incorrect) answers for completeness, in case they are of interest. For convenience, we additionally report a \emph{support count} per QA: the number of candidate articles whose open-book answer is judged correct. Some examples are presented below.

\begin{figure}[ht]
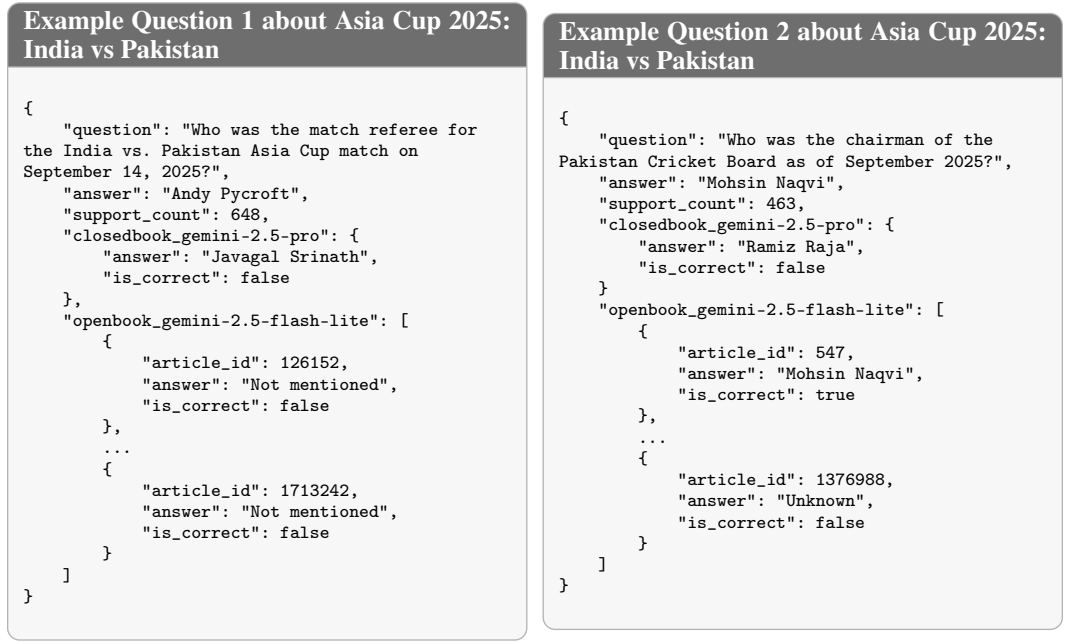

\centering
\begin{adjustbox}{max totalsize={\textwidth}{1\textheight},center}
\begin{minipage}{\textwidth}
\begin{tcbraster}[
  raster columns=2,
  raster column skip=2mm,
  raster left skip=0pt,
  raster right skip=0pt,
  width=\textwidth
]
\begin{jsonblock}{Example Question 1 about Asia Cup 2025: India vs Pakistan}
{
    "question": "Who was the match referee for the India vs. Pakistan Asia Cup match on September 14, 2025?",
    "answer": "Andy Pycroft",
    "support_count": 648,
    "closedbook_gemini-2.5-pro": {
        "answer": "Javagal Srinath",
        "is_correct": false
    },
    "openbook_gemini-2.5-flash-lite": [
        {
            "article_id": 126152,
            "answer": "Not mentioned",
            "is_correct": false
        },
        ...
        {
            "article_id": 1713242,
            "answer": "Not mentioned",
            "is_correct": false
        }
    ]
}
\end{jsonblock}
\begin{jsonblock}{Example Question 2 about Asia Cup 2025: India vs Pakistan}
{
    "question": "Who was the chairman of the Pakistan Cricket Board as of September 2025?",
    "answer": "Mohsin Naqvi",
    "support_count": 463,
    "closedbook_gemini-2.5-pro": {
        "answer": "Ramiz Raja",
        "is_correct": false
    }
    "openbook_gemini-2.5-flash-lite": [
        {
            "article_id": 547,
            "answer": "Mohsin Naqvi",
            "is_correct": true
        },
        ...
        {
            "article_id": 1376988,
            "answer": "Unknown",
            "is_correct": false
        }
    ]
}
\end{jsonblock}
\end{tcbraster}
\end{minipage}
\end{adjustbox}
\caption{Example QAs about Asia Cup 2025: India vs Pakistan}
\label{fig:news-articles-examples}
\end{figure}

Within each cluster, we split the QAs roughly evenly into validation and test, then pool across clusters to form the final sets: validation is used for checkpoint selection, and test for final performance reporting.

\subsection{Subsampling}
The full set of articles constitutes \textsc{News-Large}, with roughly 1.7M articles. We then form \textsc{News-Small} ($\approx$200K articles) as the union of every released QA's support set. Additionally, we construct \textsc{News-Medium} ($\approx$580K articles) as the union of (i) every article assigned to a cluster during the clustering stage and (ii) all of \textsc{News-Small}. Since \textsc{News-Small} consists exactly of the supporting articles and \textsc{News-Small} $\subseteq$ \textsc{News-Medium} $\subseteq$ \textsc{News-Large}, both the QAs and their full support sets are identical across all three sizes. The support counts of the QAs are presented below.

\begin{figure}[ht]
    \centering
    \includegraphics[width=1\linewidth]{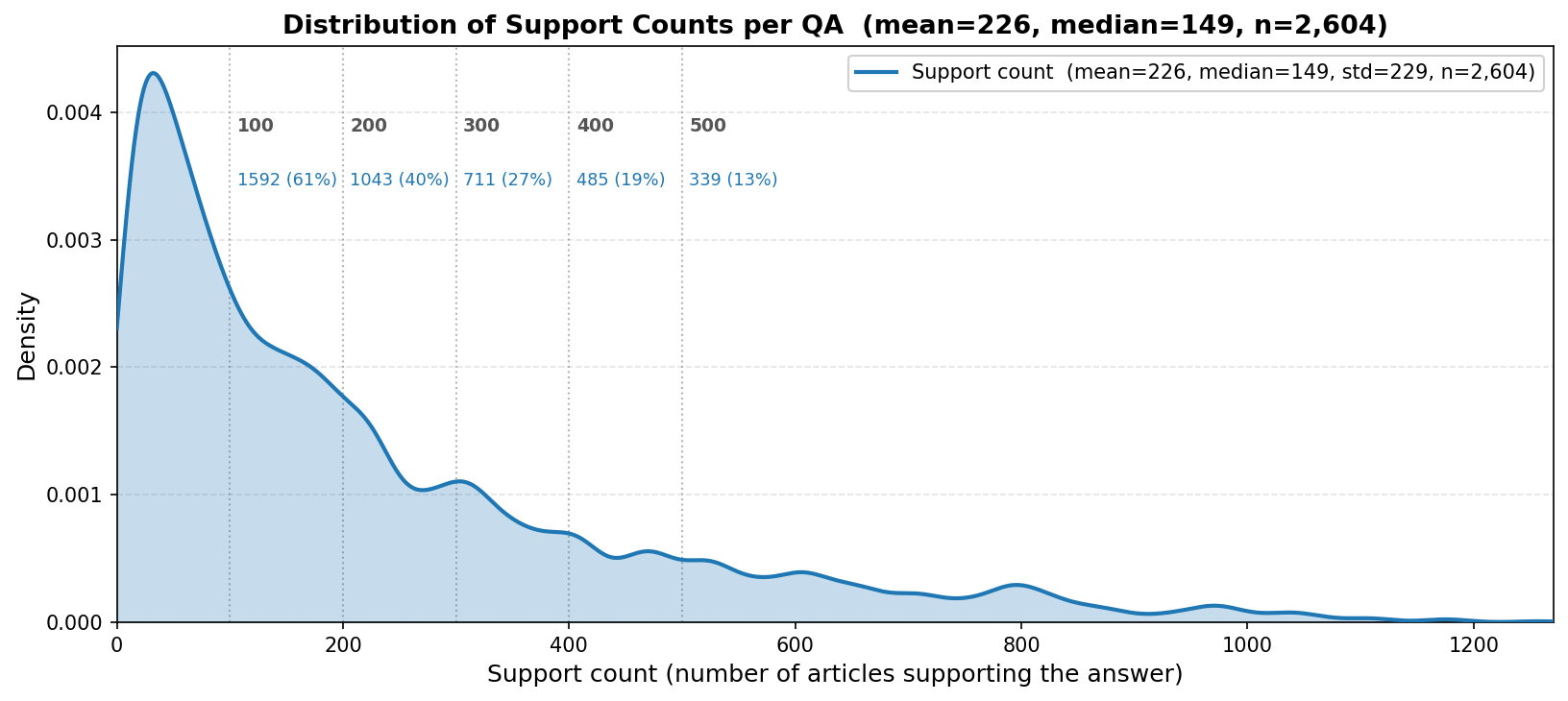}
    \caption{Support count distribution of the news QAs. The blue curve shows the per-article distribution and vertical lines mark support $\geq$ 100, 200, 300, 400, 500 articles, with the corresponding cumulative percentages shown in blue.}
    \label{fig:news_support_count}
\end{figure}

\newpage

\end{document}